\documentclass{article} %

\usepackage[utf8]{inputenc} %
\usepackage[T1]{fontenc}    %
\usepackage[hyperfootnotes=false]{hyperref}       %
\usepackage{url}            %
\usepackage{booktabs}       %
\usepackage{amsmath,amsfonts,amsthm}       %
\usepackage{mathtools}      %
\usepackage{nicefrac}       %
\usepackage{microtype}      %
\usepackage{subcaption}
\usepackage{bbm}
\usepackage{multirow}
\usepackage[inline]{enumitem}
\usepackage{diagbox}
\usepackage{pifont}
\usepackage[capitalise]{cleveref}  %
\usepackage{comment}
\usepackage{etoolbox}
\usepackage{xcolor}
\usepackage{graphicx}
\usepackage[font=small]{caption}

\graphicspath{{figures/}{../figures/}}
\usepackage{import}
\usepackage{subfiles}

\newtoggle{arxiv}
\toggletrue{arxiv}

\newtheorem{theorem}{Theorem}

\newtheorem{proposition}[theorem]{Proposition}
\newtheorem{definition}{Definition}

\newcommand{\R}{\mathbb{R}}
\newcommand{\N}{\mathbb{N}}
\newcommand{\Rn}[1]{(-\infty, #1]}
\renewcommand{\d}{\mathop{}\!\mathrm{d}}
\newcommand{\dd}{\mathop{}\!\mathrm{d}} %
\newcommand{\ppt}{\frac{\partial}{\partial t}}
\newcommand{\ddt}{\frac{d}{d t}}
\newcommand{\defeq}{\coloneqq}
\newcommand{\norm}[1]{\left\|{#1}\right\|} %
\providecommand{\abs}[1]{\left\lvert#1\right\rvert}
\DeclareMathOperator*{\diag}{diag}
\DeclareMathOperator{\proj}{proj}

\DeclareMathOperator{\coef}{coef}
\DeclareMathOperator{\hippo}{hippo}

\newcommand*\samethanks[1][\value{footnote}]{\footnotemark[#1]}

\usepackage[numbers,sort]{natbib}
\newlength{\defbaselineskip}
\title{HiPPO: Recurrent Memory with Optimal Polynomial Projections}
\usepackage{authblk}
\author[$\dagger$]{Albert Gu\thanks{Equal contribution. Order determined by coin flip.}}
\author[$\dagger$]{Tri Dao\samethanks}
\author[$\dagger$]{Stefano Ermon}
\author[$\ddagger$]{Atri Rudra}
\author[$\dagger$]{Christopher R{\'e}}
\affil[$\dagger$]{Department of Computer Science, Stanford University}
\affil[$\ddagger$]{Department of Computer Science and Engineering, University at Buffalo, SUNY\vspace{4pt}}
\affil[ ]{{\texttt{\{albertgu,trid\}@stanford.edu}, \texttt{ermon@cs.stanford.edu}, \texttt{atri@buffalo.edu}, \texttt{chrismre@cs.stanford.edu}}}

\begin{document}

\maketitle

\begin{abstract}
  A central problem in learning from sequential data is representing cumulative
  history in an incremental fashion as more data is processed.
  We introduce a general framework (HiPPO) for the online compression of continuous signals and discrete time series
  by projection onto polynomial bases.
  Given a measure that specifies the importance of each time step in the past,
  HiPPO produces an optimal solution to a natural \emph{online function approximation} problem.
  As special cases, our framework yields a short derivation of the recent Legendre Memory Unit (LMU) from first principles,
  and generalizes the ubiquitous gating mechanism of recurrent neural networks such as GRUs.
  This formal framework yields a new memory update mechanism (HiPPO-LegS) that scales
  through time to remember all history, avoiding priors on the timescale.
  HiPPO-LegS enjoys the theoretical benefits of timescale robustness, fast updates,
  and bounded gradients.
  By incorporating the memory dynamics into recurrent neural networks, HiPPO RNNs can
  empirically capture complex temporal dependencies.
  On the benchmark permuted MNIST dataset, HiPPO-LegS sets a new
  state-of-the-art accuracy of 98.3\%.
  Finally, on a novel trajectory classification task testing robustness to out-of-distribution timescales and missing data, HiPPO-LegS outperforms RNN and neural ODE baselines by 25-40\% accuracy.
\end{abstract}

\section{Introduction}
\label{sec:intro}

  Modeling and learning from sequential data is a fundamental problem in
  modern machine learning, underlying tasks such as language modeling, speech
  recognition, video processing, and reinforcement learning.
  A core aspect of modeling long-term and complex temporal dependencies is \emph{memory}, or
  storing and incorporating information from previous time steps.
  The challenge is learning a representation of the entire cumulative history using bounded storage,
  which must be updated online as more data is received.

  One established approach is to model
  a state that evolves over time as it incorporates more information.
  The deep learning instantiation of this approach is the recurrent neural network (RNN),
  which is known to suffer from a limited memory horizon~\citep{lstm,jaeger2004harnessing,pascanu2013difficulty} (e.g., the ``vanishing gradients'' problem).
  Although various heuristics have been proposed to overcome this,
  such as gates in the successful LSTM and GRU~\citep{lstm, cho2014learning},
  or higher-order frequencies in the recent Fourier Recurrent
  Unit~\citep{zhang2018learning} and Legendre Memory Unit
  (LMU)~\citep{voelker2019legendre},
  a unified understanding of memory remains a challenge.
  Furthermore, existing methods generally require priors on the sequence length or timescale and are ineffective outside this range~\citep{tallec2018can, voelker2019legendre}; this can be problematic in settings with distribution shift (e.g.\ arising from different instrument sampling rates in medical data~\cite{saab2020weak,shah2018temple}).
  Finally, many of them lack theoretical guarantees on how well they capture long-term
  dependencies, such as gradient bounds.
  To design a better memory representation, we would ideally
  (i) have a unified view of these existing methods,
  (ii) be able to address dependencies of any length without priors on the timescale,
  and
  (iii) have a rigorous theoretical understanding of their memory mechanism.

  Our insight is to phrase \emph{memory} as a technical problem of \emph{online
    function approximation} where a function $f(t) : \R_+ \to \R$ is summarized by storing its
  optimal coefficients in terms of some basis functions.
  This approximation is evaluated with respect to a measure that
  specifies the importance of each time in the past.
  Given this function approximation formulation, orthogonal polynomials (OPs) emerge as a natural basis since
  their optimal coefficients can be expressed in closed form~\citep{chihara}.
  With their rich and well-studied history~\citep{szego}, along with their
  widespread use in approximation theory~\citep{trefethen2019approximation} and
  signal processing~\citep{proakis2001digital}, OPs bring a library of
  techniques to this memory representation problem.
  We formalize a framework, \textbf{HiPPO} (high-order polynomial projection
  operators), which produces operators that project arbitrary functions onto the space of
  orthogonal polynomials with respect to a given measure.
  This general framework allows us to analyze several families of measures,
  where this operator, as a closed-form ODE or linear recurrence, allows fast incremental updating of the optimal polynomial approximation as the input function is revealed through time.

  By posing a formal optimization problem underlying recurrent sequence models,
  the HiPPO framework (Section~\ref{sec:framework})
  generalizes and explains previous methods, unlocks new methods
  appropriate for sequential data at different timescales, and comes with several theoretical guarantees.
  (i) For example, with a short derivation we exactly recover as a special case
  the LMU~\citep{voelker2019legendre} (Section~\ref{subsec:high_order_projection}), which proposes an update rule that projects
  onto fixed-length sliding windows through time.%
  \footnote{The LMU was originally motivated by spiking neural networks in
    modeling biological nervous systems; its derivation is not self-contained but a
    sketch can be pieced together from~\citep{voelker2019legendre,voelker2018improving,voelker2019dynamical}.}
  HiPPO also sheds new light on classic techniques such as the gating mechanism
  of LSTMs and GRUs, which arise in one extreme using only low-order degrees in
  the approximation (Section~\ref{subsec:low_order_projection}).
  (ii)
    By choosing more suitable measures, HiPPO yields a novel mechanism (Scaled Legendre, or LegS) that
    always takes into account the function's full history instead of a sliding window.
    This flexibility removes the need for hyperparameters or priors on the sequence length,
    allowing LegS to generalize to different input timescales.
    (iii)
    The connections to dynamical systems and approximation theory
    allows us to show several theoretical benefits of HiPPO-LegS:
    invariance to input timescale, asymptotically more efficient updates, and
    bounds on gradient flow and approximation error (Section~\ref{sec:theory_legs}).

  We integrate the HiPPO memory mechanisms into RNNs, and empirically show that they outperform baselines on standard tasks used to benchmark long-term dependencies.
  On the permuted MNIST dataset, %
  our hyperparameter-free HiPPO-LegS method achieves a new state-of-the-art
  accuracy of 98.3\%,
  beating the previous RNN SoTA by over 1 point and even outperforming models
  with global context such as transformers (Section~\ref{subsec:membenchmark}).
  Next, we demonstrate the timescale robustness of HiPPO-LegS on a novel
  trajectory classification task, where it is able to generalize to unseen
  timescales and handle missing data whereas RNN and neural ODE baselines fail
  (Section~\ref{subsec:exp-timescale}).
  Finally, we validate HiPPO's theory, including computational efficiency and
  scalability, allowing fast and accurate online function reconstruction over
  millions of time steps (Section~\ref{subsec:exp-scalability}).
  Code for reproducing our experiments is available at \url{https://github.com/HazyResearch/hippo-code}.

\section{The HiPPO Framework: High-order Polynomial Projection Operators}
\label{sec:framework}

We motivate the problem of online function approximation with projections
as an approach to learning memory representations
(Section~\ref{subsec:hippo-setup}).
Section~\ref{subsec:hippo-framework} describes the general
HiPPO framework to derive memory updates, including a precise definition of the technical problem 
we introduce, and an overview of our approach to solving it.
Section~\ref{subsec:high_order_projection} instantiates the framework to recover
the LMU and yield new memory updates (e.g.\ HiPPO-LagT), demonstrating the generality
of the HiPPO framework.
Section~\ref{sec:discretization} discusses how to convert the main continuous-time results into practical discrete versions.
Finally in Section~\ref{subsec:low_order_projection} we show how gating
in RNNs is an instance of HiPPO memory.

\subsection{HiPPO Problem Setup}
\label{subsec:hippo-setup}

Given an input function $f(t) \in \mathbb{R}$ on $t \ge 0$, many problems require operating on the cumulative \emph{history} $f_{\le t} := f(x) \mid_{x \le t}$ at every time $t \ge 0$,
in order to understand the inputs seen so far and make future predictions.
Since the space of functions is intractably large, the history cannot be perfectly memorized and must be compressed; we propose the general approach of projecting it onto a subspace of bounded dimension.
Thus, our goal is to maintain (online) this compressed representation of the history.
In order to specify this problem fully, we require two ingredients: a way to quantify the approximation, and a suitable subspace.

\paragraph{Function Approximation with respect to a Measure.}
Assessing the quality of an approximation requires defining a distance in function space.
Any probability measure $\mu$ on $[0, \infty)$ equips the space of square integrable functions with inner product
$
  \langle f, g \rangle_\mu = \int_0^\infty f(x) g(x) \d \mu(x),
$
inducing a Hilbert space structure $\mathcal{H}_\mu$ and corresponding norm $\| f \|_{L_2(\mu)} = \langle f, f \rangle_\mu^{1/2}$.

\paragraph{Polynomial Basis Expansion.}
Any $N$-dimensional subspace $\mathcal{G}$ of this function space is a suitable candidate for the approximation.
The parameter $N$ corresponds to the order of the approximation, or the size of the compression;
the projected history can be represented by the $N$ coefficients of its expansion in any basis of $\mathcal{G}$.
For the remainder of this paper, we use the polynomials as a natural basis, so that $\mathcal{G}$ is the set of polynomials of degree less than $N$.
We note that the polynomial basis is very general; for example, the Fourier basis $\sin(nx), \cos(nx)$ can be seen as polynomials on the unit circle $(e^{2\pi i x})^n$ (cf.\ \cref{sec:derivation-fourier}).
In \cref{sec:hippo-framework-details}, we additionally formalize a more
general framework that allows different bases other than polynomials by tilting the measure with another function.

\paragraph{Online Approximation.}
Since we care about approximating $f_{\le t}$ for every time $t$, we also let the measure vary through time.
For every $t$, let $\mu^{(t)}$ be a measure supported on $(-\infty, t]$ (since $f_{\le t}$ is only defined up to time $t$).
Overall, we seek some $g^{(t)}\in\mathcal{G}$ that minimizes $\|f_{\le t} - g^{(t)}\|_{L_2(\mu^{(t)})}$.
Intuitively, the measure $\mu$ controls the importance of various parts of the input domain,
and the basis defines the allowable approximations.
The challenge is how to solve the optimization problem in closed form given $\mu^{(t)}$,
and how these coefficients can be maintained online as $t \to \infty$.

\subsection{General HiPPO framework}
\label{subsec:hippo-framework}

We provide a brief overview of the main ideas behind solving this problem, which provides a surprisingly simple and general strategy for many measure families $\mu^{(t)}$.
This framework builds upon a rich history of the well-studied \emph{orthogonal polynomials} and related transforms in the signal processing literature.
Our formal abstraction (\cref{def:hippo}) departs from prior work on sliding transforms in several ways, which we discuss in detail in \cref{sec:rw-ops}.
For example, our concept of the time-varying measure allows choosing $\mu^{(t)}$ more appropriately, which will lead to solutions with qualitatively different behavior.
\cref{sec:hippo-framework-details} contains the full details and formalisms of our framework.

\paragraph{Calculating the projection through continuous dynamics.}
As mentioned, the approximated function can be represented by the $N$ coefficients of its expansion in any basis;
the first key step is to choose a suitable basis $\{g_n\}_{n < N}$ of $\mathcal{G}$.
Leveraging classic techniques from approximation theory, a natural basis is the set of orthogonal polynomials for the measure $\mu^{(t)}$,
which forms an orthogonal basis of the subspace.
Then the coefficients of the optimal basis expansion are simply $c^{(t)}_n := \langle f_{\le t}, g_n \rangle_{\mu^{(t)}}$.

The second key idea is to differentiate this projection in $t$, where
differentiating through the integral (from the inner product
$\langle f_{\leq t},  g_n \rangle_{\mu^{(t)}}$) will often lead to a self-similar relation
allowing $\frac{d}{dt} c_n(t)$ to be expressed in terms of $(c_k(t))_{k\in [N]}$ and $f(t)$.
Thus the coefficients $c(t) \in \R^N$ should evolve as an ODE, with dynamics determined by $f(t)$.

\paragraph{The HiPPO abstraction: online function approximation.}
\begin{definition}%
    \label{def:hippo}
    Given a time-varying measure family $\mu^{(t)}$ supported on $(-\infty, t]$, an
    $N$-dimensional subspace $\mathcal{G}$ of polynomials, and a continuous
    function $f \colon \mathbb{R}_{\geq 0} \to \mathbb{R}$,
    HiPPO defines a \emph{projection} operator $\proj_t$ and a \emph{coefficient extraction} operator $\coef_t$ at every time $t$, with the following properties:
    \begin{enumerate}[label=(\arabic*), topsep=0em, partopsep=0.0em, itemsep=0em, parsep=0.1em, leftmargin=2em]
      \item $\proj_t$ takes the function $f$ restricted up to time $t$,
      $f_{\leq t} \defeq f(x) \mid_{x \leq t}$, and maps it to a
      polynomial $g^{(t)} \in \mathcal{G}$, that minimizes the approximation error $\|f_{\leq t} - g^{(t)}\|_{L_2(\mu^{(t)})}$.
        \item $\coef_t: \mathcal{G} \to \mathbb{R}^N$ maps the polynomial $g^{(t)}$
        to the coefficients $c(t) \in \mathbb{R}^N$ of the basis of orthogonal
        polynomials defined with respect to the measure $\mu^{(t)}$.
    \end{enumerate}
    The composition $\coef \circ \proj$ is called $\hippo$, which is an operator
    mapping a function $f: \R_{\geq 0} \to \R$ to the optimal projection coefficients
    $c: \R_{\geq 0} \to \R^N$, i.e.\ $(\hippo(f))(t) = \coef_t(\proj_t(f))$.
\end{definition}

For each $t$, the problem of optimal projection
$\proj_t(f)$ is well-defined by the above inner products,
but this is intractable to compute naively.
Our derivations (\cref{sec:derivations}) will show that
the coefficient function $c(t) = \coef_t(\proj_t(f))$ has the form of an ODE
satisfying $\frac{d}{dt} c(t) = A(t) c(t) + B(t) f(t)$ for some $A(t) \in \R^{N \times N}$, $B(t) \in \R^{N \times 1}$.
Thus our results show how to tractably obtain $c^{(t)}$ \emph{online} by solving an
ODE, or more concretely by running a discrete recurrence.
When discretized, HiPPO takes in a sequence of real values and produces a
sequence of $N$-dimensional vectors.

\begin{figure}
  \centering
  \includegraphics[width=\linewidth]{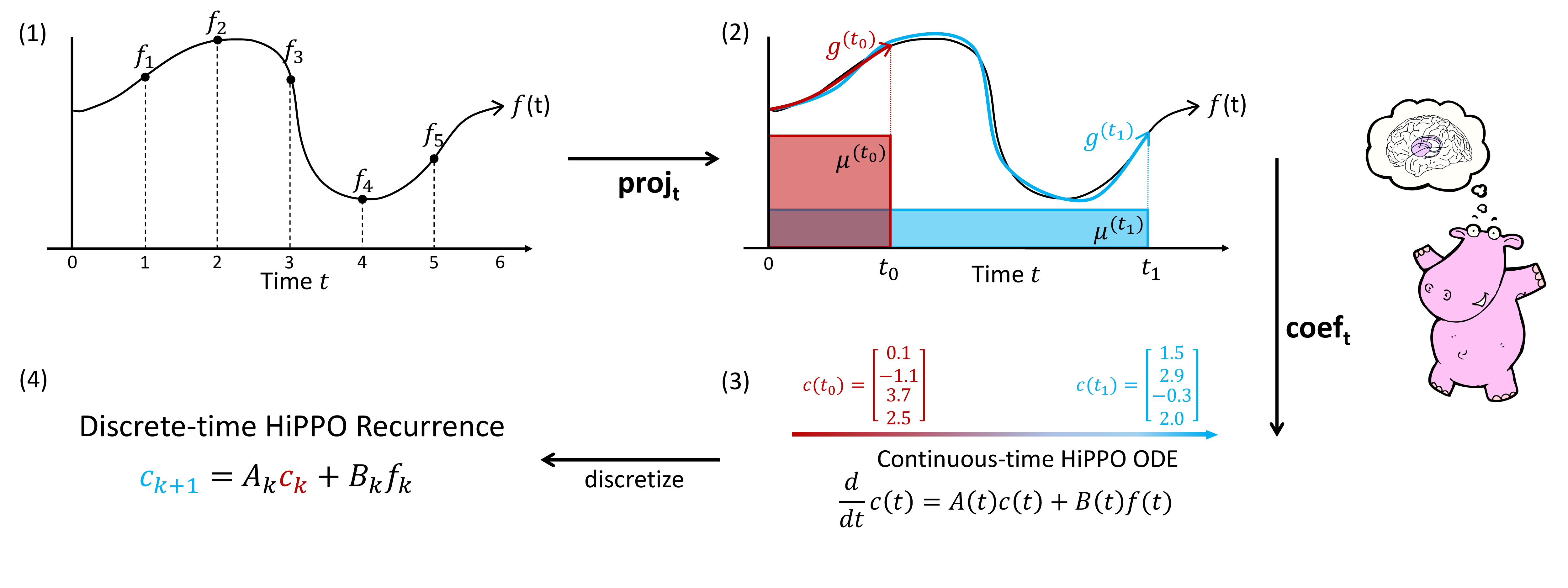}
  \caption{
    \textbf{Illustration of the HiPPO framework.}
    (1) For any function $f$, (2) at every time $t$ there is an optimal projection $g^{(t)}$ of $f$ onto the space of polynomials, with respect to a measure $\mu^{(t)}$ weighing the past.
    (3) For an appropriately chosen basis, the corresponding coefficients $c(t)\in\R^N$ representing a compression of the history of $f$ satisfy linear dynamics.
    (4) Discretizing the dynamics yields an efficient closed-form recurrence for online compression of time series $(f_k)_{k\in\N}$.
  }
  \label{fig:framework}
\end{figure}

Figure~\ref{fig:framework} illustrates the overall framework when we use uniform measures.
Next, we give our main results showing $\hippo$ for several concrete instantiations of the framework.

\subsection{High Order Projection: Measure Families and HiPPO ODEs}
\label{subsec:high_order_projection}

Our main theoretical results are instantiations of HiPPO for various measure families $\mu^{(t)}$.
We provide two examples of natural sliding window measures and the corresponding projection operators.
The unified perspective on memory mechanisms allows us to derive these closed-form solutions with the same strategy, provided in
Appendices \ref{sec:derivation-legt},\ref{sec:derivation-lagt}.
The first explains the core Legendre Memory Unit (LMU)~\citep{voelker2019legendre} update in a principled way and characterizes its limitations,
while the other is novel, demonstrating the generality of the HiPPO framework.
\cref{sec:derivations} contrasts the tradeoffs of these measures
(\cref{fig:measures}), contains proofs of their derivations, and derives
additional HiPPO formulas for other bases such as Fourier (recovering the
Fourier Recurrent Unit~\citep{zhang2018learning}) and Chebyshev.

The \textbf{translated Legendre (LegT)} measures assign uniform weight to the most recent history $[t-\theta, t]$.
There is a hyperparameter $\theta$ representing the length of the sliding window,
or the length of history that is being summarized.
The \textbf{translated Laguerre (LagT)} measures instead use the exponentially decaying measure, assigning more importance to recent history.
\begin{equation*}
    \textbf{LegT}: \mu^{(t)}(x) = \frac{1}{\theta} \mathbb{I}_{[t-\theta, t]}(x)
    \qquad
    \textbf{LagT}: \mu^{(t)}(x)
    = e^{-(t-x)} \mathbb{I}_{(-\infty, t]}(x)
    =
    \begin{cases}
      e^{x-t} & \mbox{if } x \le t \\
      0 & \mbox{if } x > t
    \end{cases}
\end{equation*}

\begin{theorem}
  \label{thm:legt-lagt}
  For LegT and LagT, the $\hippo$ operators satisfying \cref{def:hippo} are given by linear time-invariant (LTI) ODEs $\ddt c(t) = - A c(t) + B f(t)$, where $A \in \R^{N \times N}, B \in \R^{N \times 1}$:

  \small
  \begin{minipage}{.60\linewidth}
    \textnormal{\textbf{LegT}:}
    \vspace{-1em}
    \begin{equation}
      \label{eq:translated-legendre-dynamics}
      A_{nk} =
      \frac{1}{\theta}
      \begin{cases}
        (-1)^{n-k} (2n+1) & \mbox{if } n \ge k \\
        2n+1 & \mbox{if } n \le k
      \end{cases},
      \quad
      B_n = \frac{1}{\theta} (2n+1) (-1)^n
    \end{equation}
  \end{minipage}%
  \hfill%
  \begin{minipage}{.35\linewidth}
    \textnormal{\textbf{LagT}:}
    \vspace{-1em}
    \begin{equation}
      \label{eq:laguerre-dynamics}
      A_{nk} =
      \begin{cases}%
        1 & \mbox{if } n \ge k \\
        0 & \mbox{if } n < k \\
      \end{cases},
      \quad
      B_n = 1
    \end{equation}
  \end{minipage}

\end{theorem}

Equation~\eqref{eq:translated-legendre-dynamics} proves the LMU update \citep[equation (1)]{voelker2019legendre}.
Additionally, our derivation (Appendix~\ref{sec:derivation-legt}) shows that outside of the projections,
there is another source of approximation.
This sliding window update rule requires access to $f(t-\theta)$, which is no longer available;
it instead assumes that the current coefficients $c(t)$ are an accurate enough model of the function $f(x)_{x \le t}$ that $f(t-\theta)$ can be recovered.

\subsection{HiPPO recurrences: from Continuous to Discrete Time with ODE Discretization}
\label{sec:discretization}

Since actual data is inherently discrete (e.g.\ sequences and time series),
we discuss how the HiPPO projection operators can be discretized using standard techniques,
so that the continuous-time HiPPO ODEs become discrete-time linear recurrences.

In the continuous case, these operators consume an input function $f(t)$ and produce an output function $c(t)$.
The discrete time case (i) consumes an input sequence $(f_k)_{k \in \N}$, %
(ii) implicitly defines a function $f(t)$ where $f(k \cdot \Delta t) = f_k$ for some step
size $\Delta t$, (iii) produces a function $c(t)$ through the ODE dynamics,
and (iv) discretizes back to an output sequence $c_k := c(k \cdot \Delta t)$.

The basic method of discretizating an ODE $\frac{d}{dt} c(t) = u(t, c(t), f(t))$ chooses a step size $\Delta t$ and performs the discrete updates
$c(t+\Delta t) = c(t) + \Delta t \cdot u(t, c(t), f(t))$.\footnote{This is known as the Euler method, used for illustration here; our experiments use the more numerically stable Bilinear and ZOH methods.
\cref{sec:discretization-full} provides a self-contained overview of our full discretization framework.}
In general, this process is sensitive to the \emph{discretization step size} hyperparameter $\Delta t$.

Finally, we note that this provides a way to seamlessly handle timestamped data, even with missing values:
the difference between timestamps indicates the (adaptive) $\Delta t$ to use in discretization~\citep{chen2018neural}.
\cref{sec:discretization-full} contains a full discussion of discretization.

\subsection{Low Order Projection: Memory Mechanisms of Gated RNNs}
\label{subsec:low_order_projection}

As a special case, we consider what happens if we do not incorporate higher-order polynomials in the projection problem.
Specifically, if $N=1$, then the discretized version of HiPPO-LagT \eqref{eq:laguerre-dynamics} becomes
$c(t + \Delta t) = c(t) + \Delta t(-Ac(t) + Bf(t)) = (1 - \Delta t)c(t) + \Delta t f(t)$, since $A = B = 1$.
If the inputs $f(t)$ can depend on the hidden state $c(t)$ and the discretization step size $\Delta t$ is chosen adaptively (as a function of input $f(t)$ and state $c(t)$),
as in RNNs, then this becomes exactly a \emph{gated} RNN.
For instance, by stacking multiple units in parallel and choosing a specific update function,
we obtain the GRU update cell as a special case.\footnote{The LSTM cell update is similar, with a parameterization known as ``tied'' gates \cite{greff2016lstm}.}
In contrast to HiPPO which uses one hidden feature and projects it onto high order polynomials,
these models use many hidden features but only project them with degree 1.
This view sheds light on these classic techniques by showing how they can be derived from first principles.

\section{HiPPO-LegS: Scaled Measures for Timescale Robustness}
\label{sec:theory_legs}

Exposing the tight connection between online function approximation and memory allows us to produce memory mechanisms with better theoretical properties, simply by choosing the measure appropriately.
Although sliding windows are common in signal processing (\cref{sec:rw-ops}), a more intuitive approach for memory should \emph{scale} the window over time to avoid forgetting.

Our novel \textbf{scaled Legendre measure (LegS)} assigns uniform weight to all history $[0, t]$:
$\mu^{(t)} = \frac{1}{t} \mathbb{I}_{[0, t]}$. %
App~\ref{sec:derivations}, Fig.~\ref{fig:measures} compares LegS, LegT, and LagT visually, showing the advantages of the scaled measure.

Simply by specifying the desired measure, specializing the HiPPO framework (Sections~\ref{subsec:hippo-framework}, \ref{sec:discretization})
yields a new memory mechanism (proof in \cref{sec:derivation-legs}).
\begin{theorem}
  \label{thm:legs}
  The continuous- \eqref{eq:scaled-legendre-dynamics}
  and discrete- \eqref{eq:legs-discrete} time dynamics for \textbf{HiPPO-LegS} are:
  \small

  \begin{minipage}{.35\linewidth}
    \begin{align}
      \ddt c(t) &= -\frac{1}{t} A c(t) + \frac{1}{t} B f(t)
      \label{eq:scaled-legendre-dynamics}
      \\
      c_{k+1} &= \left( 1-\frac{A}{k} \right) c_k + \frac{1}{k} B f_k
     \label{eq:legs-discrete}
    \end{align}
  \end{minipage}%
  \hfill
  \begin{minipage}{.52\linewidth}
    \begin{align*}
      A_{nk}
      =
      \begin{cases}
        (2n+1)^{1/2}(2k+1)^{1/2} & \mbox{if } n > k \\
        n+1 & \mbox{if } n = k \\
        0 & \mbox{if } n < k
      \end{cases},
      \qquad
      B_n = (2n+1)^{\frac{1}{2}}
    \end{align*}
  \end{minipage}
\end{theorem}

We show that HiPPO-LegS enjoys favorable
theoretical properties: it is invariant to input timescale, is fast to compute,
and has bounded gradients and approximation error.
All proofs are in \cref{sec:hippo-theory}.

\paragraph{Timescale robustness.}
As the window size of LegS is adaptive, %
projection onto this measure is intuitively robust to timescales.
Formally, the HiPPO-LegS operator is \emph{timescale-equivariant}:
dilating the input $f$ does not change the approximation coefficients.
\begin{proposition}
    \label{prop:timescale}
    For any scalar $\alpha > 0$, if $h(t) = f(\alpha t)$, then
    $\hippo(h)(t) = \hippo(f)(\alpha t)$.
    \\ In other words, if $\gamma : t \mapsto \alpha t$ is any dilation function, then
    $\hippo(f \circ \gamma) = \hippo(f) \circ \gamma$.
\end{proposition}

Informally, this is reflected by HiPPO-LegS having \emph{no timescale hyperparameters};
in particular, the discrete recurrence \eqref{eq:legs-discrete} is invariant to the discretization step size.%
\footnote{\eqref{eq:legs-discrete} uses the Euler method for illustration; HiPPO-LegS is invariant to other discretizations (\cref{sec:discretization-full}).}
By contrast, LegT has a hyperparameter $\theta$ for the window size, and both LegT and LagT have a step size hyperparameter $\Delta t$ in the discrete time case.
This hyperparameter is important in practice; \cref{subsec:low_order_projection} showed that $\Delta t$ relates to the gates of RNNs, which are known to be sensitive to their parameterization~\cite{jozefowicz2015empirical,tallec2018can,gu2020improving}.
We empirically demonstrate the benefits of timescale robustness in \cref{subsec:exp-timescale}.

\paragraph{Computational efficiency.}
In order to compute a single step of the discrete HiPPO update, the main operation is multiplication by the (discretized) square matrix $A$.
More general discretization specifically requires fast multiplication for any matrix of the form $I + \Delta t \cdot A$ and $(I - \Delta t \cdot A)^{-1}$ for arbitrary step sizes $\Delta t$.
Although this is generically a $O(N^2)$ operation, LegS operators use a fixed $A$ matrix with special structure
that turns out to have fast multiplication algorithms for any discretization.%
\footnote{It is known that large families of structured matrices related to orthogonal polynomials are efficient~\cite{de2018two}.}
\begin{proposition}%
    \label{prop:efficiency}
    Under any generalized bilinear transform discretization (cf.\
    \cref{sec:discretization-full}), each step of the HiPPO-LegS recurrence in equation~\eqref{eq:legs-discrete} can be computed in $O(N)$ operations.
\end{proposition}

\cref{subsec:exp-scalability} validates the efficiency of HiPPO layers in practice, where unrolling the discretized versions of \cref{thm:legs} is 10x faster than standard matrix multiplication as done in standard RNNs.

\paragraph{Gradient flow.}
Much effort has been spent to alleviate the \emph{vanishing gradient problem} in RNNs~\cite{pascanu2013difficulty},
where backpropagation-based learning is hindered by gradient magnitudes decaying exponentially in time.
As LegS is designed for memory, it avoids the vanishing gradient issue.
\begin{proposition}%
  \label{prop:gradient-bound}
    For any times $t_0 < t_1$, the gradient norm of HiPPO-LegS operator
    for the output at time $t_1$ with respect to input at time $t_0$ is
    $\left\| \frac{\partial c(t_1)}{\partial f(t_0)} \right\| = \Theta\left( 1/t_1 \right)$.
\end{proposition}

\paragraph{Approximation error bounds.}
The error rate of LegS decreases with the smoothness of the input.

\begin{proposition}%
  \label{prop:approximation_error}
  Let $f \colon \mathbb{R}_+ \to \mathbb{R}$ be a differentiable function, and let
  $g^{(t)} = \proj_t(f)$ be its projection at time $t$ by
  HiPPO-LegS with maximum polynomial degree $N-1$.
  If $f$ is $L$-Lipschitz then $\norm{f_{\leq t} - g^{(t)}} = O(tL/\sqrt{N})$.
  If $f$ has order-$k$ bounded derivatives then
  $\norm{f_{\leq t} - g^{(t)}} = O(t^k N^{-k+1/2})$.
\end{proposition}

\section{Empirical Validation}
\label{sec:experiments}

The HiPPO dynamics are simple recurrences that can be easily incorporated into various models.
We validate three claims that suggest that when incorporated into a simple RNN, these methods--especially HiPPO-LegS--yield a recurrent architecture with improved memory capability.
In Section~\ref{subsec:membenchmark}, the HiPPO-LegS RNN outperforms
other RNN approaches in benchmark long-term dependency tasks for RNNs.
Section~\ref{subsec:exp-timescale} shows that HiPPO-LegS RNN is much more robust to
timescale shifts compared to other RNN and neural ODE models.
Section~\ref{subsec:exp-scalability} validates the distinct
theoretical advantages of the HiPPO-LegS memory mechanism, allowing fast and
accurate online function reconstruction over millions of time steps.
Experiment details and additional results are described in \cref{sec:experiment-details}.

\paragraph{Model Architectures.}
We first describe briefly how HiPPO memory updates can be incorporated into a
simple neural network architecture, yielding a simple RNN model reminiscent of
the classic LSTM.
Given inputs $x_t$ or features thereof $f_t = u(x_t)$ in any model, the HiPPO
framework can be used to memorize the history of features $f_t$.
Thus, given any RNN update function $h_{t} = \tau(h_{t-1}, x_t)$, we simply replace
$h_{t-1}$ with a projected version of the entire history of $h$, as described in
Figure~\ref{fig:cell}.
The output of each cell is $h_t$, which can be passed through any downstream module (e.g.\ a classification head trained with cross-entropy) to produce predictions.

We map the vector $h_{t-1}$ to 1D with a learned encoding before passing to
$\hippo$ (full architecture in App.~\ref{subsec:model_architectures}).

\subsection{Long-range Memory Benchmark Tasks}
\label{subsec:membenchmark}

\paragraph{Models and Baselines.}
We consider all of the HiPPO methods (\textbf{LegT}, \textbf{LagT}, and \textbf{LegS}).
As we show that many different update dynamics seem to lead to LTI systems that give sensible results (\cref{subsec:high_order_projection}), we additionally consider the \textbf{Rand} baseline that uses random $A$ and $B$ matrices (normalized appropriately) in its updates,
to confirm that the precise derived dynamics are important.
LegT additionally considers an additional hyperparameter $\theta$, which should be set to the timescale of the data if known a priori; to show the effect of the timescale, we set it to the ideal value as well as values that are too large and small.
The \textbf{MGU} is a minimal gated architecture, equivalent to a GRU without the reset gate. The HiPPO architecture we use is simply the MGU with an additional $\hippo$ intermediate layer.

We also compare to several RNN baselines designed for long-term dependencies, including the
\textbf{LSTM}~\cite{lstm}, \textbf{GRU}~\cite{chung2014empirical}, \textbf{expRNN}~\citep{lezcano2019cheap}, and \textbf{LMU}~\citep{voelker2019legendre}.%
\footnote{In our experiments, LMU refers to the architecture in~\citep{voelker2019legendre} while LegT uses the one described in \cref{fig:cell}.}

All methods have the same hidden size in our experiments.
In particular, for simplicity and to reduce hyperparameters, HiPPO variants tie the memory size $N$ to the hidden state dimension $d$, so that all methods and baselines have a comparable number of hidden units and parameters.
A more detailed comparison of model architectures is in \cref{subsec:model_architectures}.

\paragraph{Sequential Image Classification on Permuted MNIST.}
The permuted MNIST (pMNIST) task feeds inputs to a model pixel-by-pixel in the order of a fixed permutation.
The model must process the entire image sequentially -- with non-local structure -- before
outputting a classification label, requiring learning long-term dependencies.

\cref{tab:pmnist} shows the validation accuracy on the pMNIST task for
the instantiations of our framework and baselines.
We highlight that LegS has the best performance of all models.
While LegT is close at the optimal hyperparameter $\theta$, its performance can fall off drastically for a mis-specified window length.
LagT also performs well at its best hyperparameter $\Delta t$.

\cref{tab:pmnist} also compares test accuracy of our methods against reported results from the literature, where the LMU was the state-of-the-art for recurrent models.
In addition to RNN-based baselines, other sequence models have been evaluated on this dataset, despite being against the spirit of the task
because they have global receptive field instead of being strictly sequential.
With a test accuracy of 98.3\%, HiPPO-LegS sets a true state-of-the-art accuracy on the permuted MNIST dataset.

\begin{minipage}{.38\linewidth}%
  \iftoggle{arxiv}{\vspace*{-0.3em}{\vspace*{0.6em}}}
  \centering
  \includegraphics[width=\linewidth]{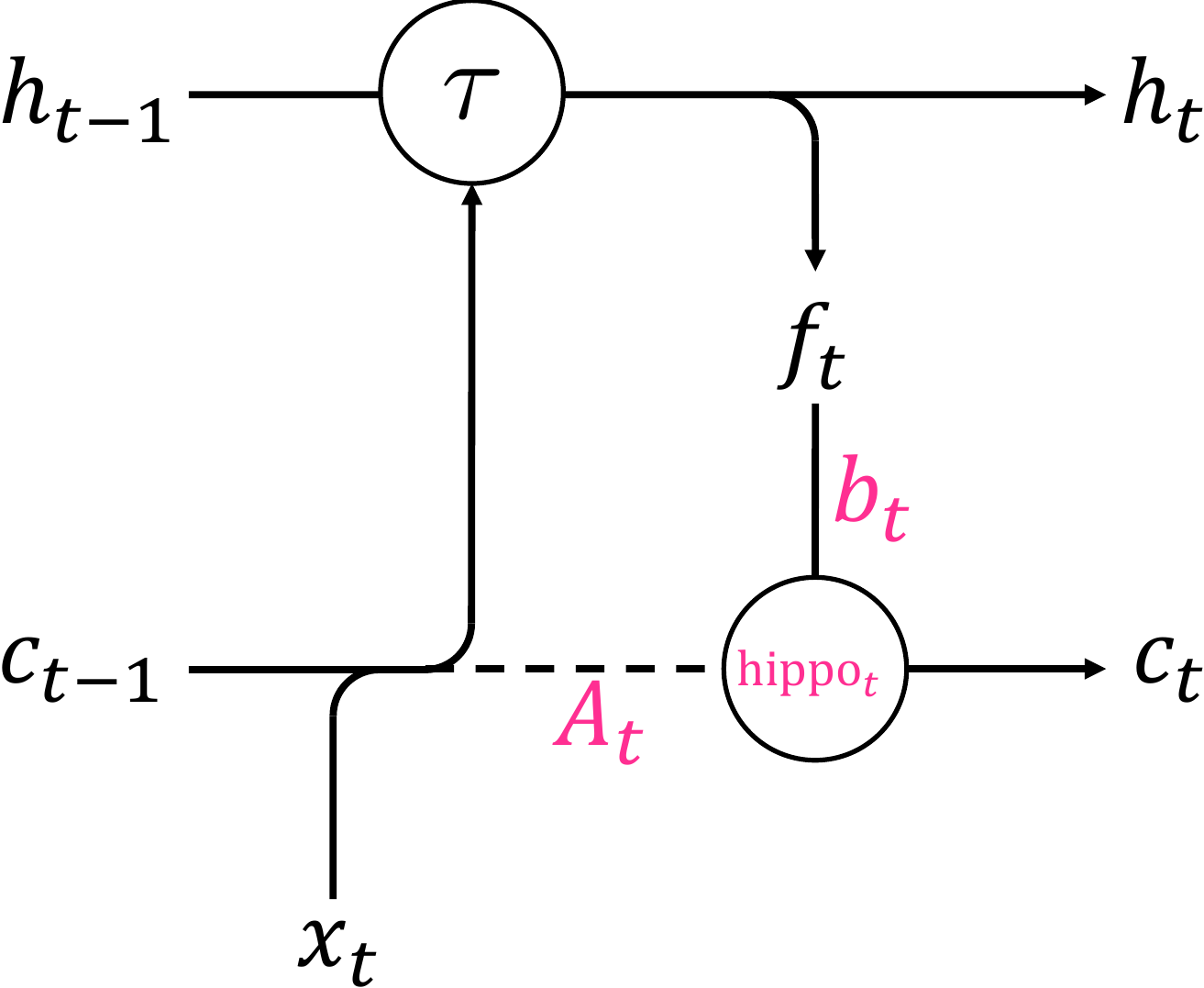}
  \captionof{figure}{HiPPO incorporated into a simple RNN model.
  $\hippo$ is the HiPPO memory operator which projects the history of the $f_t$ features depending on the chosen measure.
  }
  \label{fig:cell}
\end{minipage}%
\hfill
\begin{minipage}{.58\linewidth}
    \small
        \centering
        \begin{tabular}{@{}ll@{}}
            \toprule
            Method                     & Val. acc. (\%) \\
            \midrule
            \textbf{-LegS}        & \textbf{98.34}           \\
            -LagT            & 98.15                    \\
            -LegT $\theta = 200$  & 98.0                     \\
            -LegT $\theta = 20$   & 91.75                    \\
            -Rand                 & 69.93 \\
            \midrule
            LMU                        & 97.08                    \\
            ExpRNN                     & 94.67                    \\
            GRU                        & 93.04                    \\
            MGU                        & 89.37                    \\
            RNN                        & 52.98                    \\
            \bottomrule
        \end{tabular}%
        \hfill
        \begin{tabular}{@{}lll@{}}
            \toprule
            Model                                    & Test acc.        \\
            \midrule
            \textbf{HiPPO-LegS}                                & \textbf{98.3} \\
            \midrule
            LSTM~\citep{gu2020improving}              & 95.11         \\
            r-LSTM~\citep{trinh2018learning}         & 95.2          \\
            Dilated RNN~\citep{chang2017dilated}      & 96.1          \\
            IndRNN~\citep{indrnn}                     & 96.0          \\
            URLSTM~\citep{gu2020improving}            & 96.96         \\
            LMU~\citep{voelker2019legendre}           & 97.15         \\
            \midrule
            Transformer~\citep{trinh2018learning}     & 97.9          \\
            TCN~\citep{bai2018empirical} & 97.2          \\
            TrellisNet~\citep{trellisnet}             & 98.13         \\
            \bottomrule
        \end{tabular}
    \captionof{table}{\textbf{(Left)} pMNIST validation, average over 3 seeds. Top: Our methods. Bottom: RNN baselines.
      \textbf{(Right)} Reported test accuracies from previous works. Top: Our
      methods. Middle: Recurrent models. Bottom: Non-recurrent models requiring global receptive field.
    }
    \label{tab:pmnist}
\end{minipage}

\paragraph{Copying task.}
This standard RNN task \cite{arjovsky2016unitary} directly tests memorization, where models must regurgitate a sequence of tokens seen at the beginning of the sequence.
It is well-known that standard models such as LSTMs struggle to solve this task.
\cref{sec:experiment-details}  shows the loss for the Copying task with length $L=200$.
Our proposed update LegS solves the task almost perfectly, while LegT
is very sensitive to the window length hyperparameter.
As expected, most baselines make little progress.

\iftoggle{arxiv}{}{\vspace*{-1em}}

\subsection{Timescale Robustness of HiPPO-LegS}
\label{subsec:exp-timescale}

\paragraph{Timescale priors.}
Sequence models generally benefit from priors on the timescale,
which take the form of additional hyperparameters in standard models.
Examples include the ``forget bias'' of LSTMs which needs to be modified to address long-term dependencies~\cite{jozefowicz2015empirical,tallec2018can},
or the discretization step size $\Delta t$ of HiPPO-Lag and HiPPO-LegT (\cref{sec:discretization}).
The experiments in \cref{subsec:membenchmark} confirm their importance.
\cref{fig:copy200} (Appendix) and \cref{tab:pmnist} ablate these hyperparameters,
showing that for example the sliding window length $\theta$ must be set correctly for LegT.
Additional ablations for other hyperparameters are in \cref{sec:experiment-details}.

\paragraph{Distribution shift in trajectory classification.}
Recent trends in ML have stressed the importance of understanding robustness under distribution shift, when training and testing distributions are not i.i.d.
For time series data, for example, models may be trained on EEG data from one hospital, but deployed at another using instruments with different sampling rates~\citep{shah2018temple, saab2020weak}; or a time series may involve the same trajectory evolving at different speeds.
Following~\citet{kidger2020neural}, we consider the Character Trajectories dataset~\citep{bagnall2018uea},
where the goal is to classify a character from a sequence of pen stroke measurements, collected from one user at a fixed sampling rate.
To emulate timescale shift (e.g.\ testing on another user with slower handwriting),
we consider two standard time series generation processes:
(1) In the setting of sampling an underlying sequence at a fixed rate, we change the test sampling rate; crucially, the sequences are variable length so the models are unable to detect the sampling rate of the data.
(2) In the setting of irregular-sampled (or missing) data with timestamps, we scale the test timestamps.

Recall that the HiPPO framework models the underlying data as a continuous function and interacts with discrete input only through the discretization.
Thus, it seamlessly handles missing or irregularly-sampled data by simply evolving according to the given discretization step sizes (details in \cref{sec:discretization-full}).
Combined with LegS timescale invariance (Prop.~\ref{prop:timescale}), we expect HiPPO-LegS to work automatically in all these settings.
We note that the setting of missing data is a topic of independent interest and we compare against SOTA methods, including the GRU-D~\cite{che2018recurrent} which learns a decay between observations,
and neural ODE methods which models segments between observations with an ODE.

\cref{tab:charactertrajectories} validates that standard models can go catastrophically wrong when tested on sequences at different timescales than expected.
Though all methods achieve near-perfect accuracy ($\geq$ 95\%) without distribution shift,
aside from HiPPO-LegS, no method is able to generalize to unseen timescales.

\begin{table}[ht]
    \small
    \centering
    \caption{Test set accuracy on Character Trajectory classification on out-of-distribution timescales.   }
    \begin{tabular}{lccccccc}
        \toprule
        Model                     & \textbf{LSTM} & \textbf{GRU} & \textbf{GRU-D} & \textbf{ODE-RNN} & \textbf{NCDE} & \textbf{LMU} & \textbf{HiPPO-LegS} \\
        \midrule
        100Hz $\to$ 200Hz         & 31.9          & 25.4         & 23.1           & 41.8             & 44.7          & 6.0          & \textbf{88.8}       \\
        200Hz $\to$ 100Hz         & 28.2          & 64.6         & 25.5           & 31.5             & 11.3          & 13.1         & \textbf{90.1}       \\
        \midrule
        Missing values upsample   & 24.4          & 28.2         & 5.5            & 4.3              & 63.9          & 39.3         & \textbf{94.5}       \\
        Missing values downsample & 34.9          & 27.3         & 7.7            & 7.7              & 69.7          & 67.8         & \textbf{94.9}       \\
        \bottomrule
    \end{tabular}
    \label{tab:charactertrajectories}
\end{table}

\subsection{Theoretical Validation and Scalability}
\label{subsec:exp-scalability}

We empirically show that HiPPO-LegS can scale to capture dependencies
across millions of time steps, and its memory updates are computationally
efficient (processing up to 470,000 time steps/s).

\paragraph{Long-range function approximation.}
We test the ability of
different memory mechanisms in approximating an input function, as described in
the problem setup in Section~\ref{subsec:hippo-setup}.
The model only consists of the memory update (Section~\ref{sec:theory_legs}) and
not the additional RNN architecture.
We choose random samples from a continuous-time band-limited white noise
process, with length $10^6$.
The model is to traverse the input sequence, and then asked to reconstruct the
input, while maintaining no more than 256 units in memory (\cref{fig:function_approx}).
This is a difficult task; the LSTM fails with even sequences of length 1000 (MSE
$\approx$ 0.25).
As shown in Table~\ref{tab:mse_speed}, both the LMU and HiPPO-LegS are able to
accurately reconstruct the input function, validating that HiPPO can solve the
function approximation problem even for very long sequences.
\cref{fig:function_approx} illustrates the function and its approximations, with
HiPPO-LegS almost matching the input function while LSTM unable to do so.

\paragraph{Speed.}
HiPPO-LegS operator is computationally efficient both in theory
(Section~\ref{sec:theory_legs}) and in practice.
We implement the fast update in C++ with Pytorch binding and show in
Table~\ref{tab:mse_speed} that it can perform 470,000 time step updates per second on a single CPU
core, 10x faster than the LSTM and LMU.\footnote{The LMU is only known to be fast with the simple forward Euler
discretization~\citep{voelker2019legendre}, but not with more sophisticated methods such as bilinear and
ZOH that are required to reduce numerical errors for this task.}

\begin{minipage}{.4\linewidth}
  \small
  \centering
  \begin{tabular}{lll}
      \toprule
      Method     & Error  & Speed  \\
      & (MSE) & (elements / sec) \\
      \midrule
      LSTM       & 0.25 & 35,000             \\
      LMU        & 0.05 & 41,000             \\
      HiPPO-LegS & 0.02 & 470,000            \\
      \bottomrule
  \end{tabular}
  \captionof{table}{Function approximation error after 1 million time steps, with 256 hidden units.}
  \label{tab:mse_speed}
\end{minipage}%
\hfill
\begin{minipage}{.6\linewidth}%
  \centering
  \includegraphics[width=\linewidth]{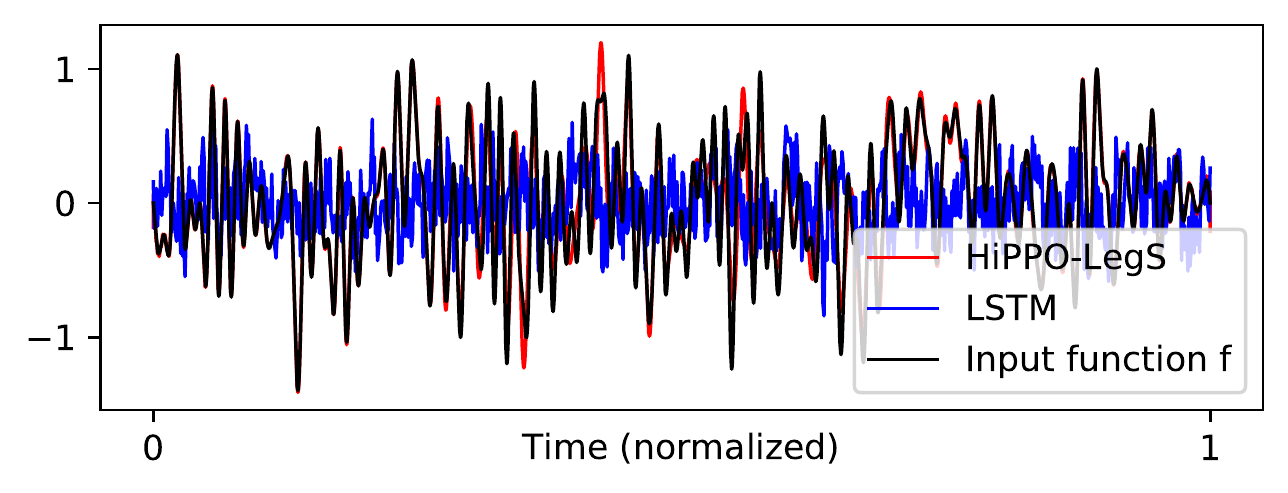}
  \captionof{figure}{Input function and its reconstructions.}
  \label{fig:function_approx}
\end{minipage}%

\subsection{Additional Experiments}
\label{subsec:sequencetasks}

We validate that the HiPPO memory updates also perform well on more generic
sequence prediction tasks not exclusively focused on memory.
Full results and details for these tasks are in \cref{sec:experiment-details}.

\paragraph{Sentiment classification task on the IMDB movie review dataset.}
Our RNNs with HiPPO memory updates
perform on par with the LSTM, while other 
long-range memory approaches such as expRNN perform poorly on this more generic task
(\cref{sec:imdb}).

\paragraph{Mackey spin glass prediction.} This physical simulation task tests the
ability to model chaotic dynamical systems. HiPPO-LegS outperforms the LSTM,
LMU, and the best hybrid LSTM+LMU model from~\cite{voelker2019legendre},
reducing normalized MSE by $30\%$ (\cref{sec:mackey}).

\section{Conclusion}
\label{sec:conclusion}

We address the fundamental problem of memory in
sequential data by proposing a framework (HiPPO) that poses the abstraction
of optimal function approximation with respect to time-varying measures.
In addition to unifying and explaining existing memory approaches,
HiPPO unlocks a new method (HiPPO-LegS) that takes a first step toward timescale robustness
and can efficiently handle dependencies across millions of time steps.
We anticipate that the study of this core problem
will be useful in improving a variety of sequence models,
and are excited about future work on integrating our memory mechanisms with other models in addition to RNNs.
We hope to realize the benefits of long-range memory on large-scale tasks such
as speech recognition, video processing, and reinforcement learning.

\subsubsection*{Acknowledgments}

We thank
Avner May, Mayee Chen, Dan Fu,
Aditya Grover,
and Daniel L\'{e}vy
for their helpful feedback.
We gratefully acknowledge the support of DARPA under Nos.\ FA87501720095 (D3M),
FA86501827865 (SDH), and FA86501827882 (ASED); NIH under No.\ U54EB020405
(Mobilize), NSF under Nos.\ CCF1763315 (Beyond Sparsity), CCF1563078 (Volume to
Velocity), and 1937301 (RTML); ONR under No.\ N000141712266 (Unifying Weak
Supervision); the Moore Foundation, NXP, Xilinx, LETI-CEA, Intel, IBM,
Microsoft, NEC, Toshiba, TSMC, ARM, Hitachi, BASF, Accenture, Ericsson,
Qualcomm, Analog Devices, the Okawa Foundation, American Family Insurance,
Google Cloud, Stanford HAI AWS cloud credit, Swiss Re, and members of the Stanford DAWN project: Teradata,
Facebook, Google, Ant Financial, NEC, VMWare, and Infosys.
The U.S.\ Government is authorized to reproduce and distribute reprints for
Governmental purposes notwithstanding any copyright notation thereon.
Any opinions, findings, and conclusions or recommendations expressed in this
material are those of the authors and do not necessarily reflect the views,
policies, or endorsements, either expressed or implied, of DARPA, NIH, ONR, or
the U.S.\ Government.
Atri Rudra’s research is supported by NSF grant CCF-1763481.

\bibliography{hippo}
\bibliographystyle{plainnat}

\appendix

\newpage

\section{Related Work}
\label{sec:related-work}

Our work touches on a variety of topics and related work, which we explore in detail.

\subsection{Signal Processing and Orthogonal Polynomials}
\label{sec:rw-ops}

\subsubsection{Sliding transforms}
The technical contributions in this work build on a rich history of approximation theory in signal processing.
Our main framework -- orthogonalizing functions with respect to time-varying measures (Section~\ref{sec:framework}) -- are related to ``online'' versions of classical signal processing transforms.
In short, these methods compute \emph{specific transforms} on \emph{sliding windows} of \emph{discrete sequences}.
Concretely, they calculate $c_{n,k} = \sum_{i=0}^{N-1} f_{k+i} \psi(i, n)$ given signal $(f_k)$, where $\{\psi(i, n)\}$ is a discrete orthogonal transform.
Our technical problem differs in several key aspects:
\begin{description}%
  \item[Specific discrete transforms]
    Examples of sliding transforms considered in the literature include the sliding DFT~\cite{farhang1994generalized,jacobsen2003sliding,jacobsen2004update,duda2010accurate}, sliding DCT~\cite{kober2004fast}, sliding discrete (Walsh-)Hadamard transform~\cite{mozafari2007efficient,ouyang2009fast,wu2012sliding}, Haar~\cite{macias2005efficient}, sliding discrete Hartley transform~\cite{kober2007fast}, and sliding discrete Chebyshev moments~\cite{chen2015fast}.
    While each of these address a specific transform, we present a general approach (\cref{sec:framework}) that addresses several transforms at once.
    Furthermore, we are unaware of sliding transform algorithms for the OPs we consider here, in particular the Legendre and Laguerre polynomials.
    Our derivations in \cref{sec:derivations} cover Legendre, (generalized) Laguerre, Fourier, and Chebyshev continuous sliding transforms.
  \item[Fixed-length sliding windows]
    All mentioned works operate in the sliding window setting, where a fixed-size context window on the discrete signal is taken into account.
    Our measure-based abstraction for approximation 
    allows considering a new type of \emph{scaled} measure where the window size increases over time, leading to methods with qualitatively different theoretical (\cref{sec:theory_legs}) and empirical properties (\cref{subsec:exp-timescale}).
    We are not aware of any previous works addressing this scaled setting.
  \item[Discrete vs. continuous time]
    Even in the fixed-length sliding window case,
    our solutions to the ``translated measure'' problems (e.g., HiPPO-LegT \cref{sec:derivation-legt})
    solve a continuous-time sliding window problem on an underlying continuous signal, then discretize.

    On the other hand, the sliding transform problems calculate transforms directly on a discrete stream.
    Discrete transforms are equivalent to calculating projection coefficients on a measure (equation \eqref{eq:coefficient}) by Gaussian quadrature,
    which assumes the discrete input is subsampled from a signal at the quadrature nodes~\citep{chihara}.
    However, since these nodes are non-uniformly spaced in general, the sliding discrete transform is not consistent with a discretization of an underlying continuous signal.

    Thus, our main abstraction (\cref{def:hippo}) has a fundamentally different interpretation than standard transforms,
    and our approach of first calculating the dynamics of the underlying continuous-time problem (e.g.\ equation~\eqref{eq:coefficient-dynamics}) is correspondingly new.

    We remark that our novel \emph{scaled measures} are fundamentally difficult to address with a standard discrete-time based approach.
    These discrete sliding methods require a fixed-size context in order to have consistent transform sizes,
    while the scaled measure would require solving transforms with an increasing number of input points over time.

\end{description}

\subsubsection{OPs in ML}
More broadly, orthogonal polynomials and orthogonal polynomial transforms have recently found applications in various facets of machine learning.
For example, \citet{dao2017gaussian} leverage the connection between orthogonal polynomials and quadrature to derive rules for computing kernel features in machine learning.
More directly, \cite{thomas2018learning} apply parametrized families of structured matrices directly inspired by orthogonal polynomial transforms (\citep{de2018two}) as layers in neural networks.
Some particular families of orthogonal polynomials such as the Chebyshev polynomials have desirable approximation properties that find many well-known classical uses in numerical analysis and optimization.
More recently, they have been applied to ML models such as graph convolutional neural networks\citep{defferrard2016convolutional}, and generalizations such as Gegenbauer and Jacobi polynomials have been used to analyze optimization dynamics\cite{yang2019mean,berthier2020accelerated}.
Generalization of orthogonal polynomials and Fourier transform, expressed as
products of butterfly matrices, have found applications in automatic algorithm
design~\citep{dao2019learning}, model compression~\citep{alizadeh2019butterfly},
and replacing hand-crafted preprocessing in speech
recognition~\citep{dao2020kaleidoscope}.
Orthogonal polynomials are known to have various efficiency results~\citep{de2018two}, and we conjecture that \cref{prop:efficiency} on the efficiency of HiPPO methods can be extended to arbitrary measures besides the ones considered in this work.

\subsection{Memory in Machine Learning}

\paragraph{Memory in sequence models}

Sequential or temporal data in areas such as language, reinforcement learning, and continual learning can involve increasingly long dependencies.
However, direct parametric modeling cannot handle inputs of unknown and potentially unbounded lengths.
Many modern solutions
such as attention~\citep{vaswani2017attention} and dilated convolutions~\citep{bai2018empirical},
are functions on finite windows, thus
sidestepping the need for an explicit memory representation.
While this suffices for certain tasks,
these approaches can only process a finite context window instead of an entire sequence.
Naively increasing the window length poses significant compute and memory challenges.
This has spurred various approaches to extend this fixed context window
subjected to compute and storage constraints~\citep{trellisnet, wu2019pay,
  dai2019transformer, child2019generating, sukhbaatar2019adaptive,
  rae2019compressive, kitaev2020reformer, roy2020efficient}.

We instead focus on the core problem of online processing and memorization of
continuous and discrete signals,
and anticipate that the study of this
foundational problem will be useful in improving a variety of models.

\paragraph{Recurrent memory}
Recurrent neural networks are a natural tool for modeling sequential data online, with the appealing property of having unbounded context; in other words they can summarize history indefinitely.
However, due to difficulties in the optimization process (vanishing/exploding gradients~\citep{pascanu2013difficulty}), particular care must be paid to endow them with longer memory.
The ubiquitous LSTM~\citep{lstm} and simplifications such as the
GRU~\citep{chung2014empirical} control the update with gates to smooth the
optimization process.
With more careful parametrization, the addition of gates alone make RNNs significantly more robust and able to address long-term dependencies~\citep{gu2020improving}.
\citet{tallec2018can} show that gates are in fact fundamental for recurrent dynamics by allowing time dilations.
Many other approaches to endowing RNNs with better memory exist, such as noise injection~\citep{gulcehre2016noisy} or non-saturating gates~\citep{chandar2019towards}, which can suffer from instability issues.
A long line of work controls the spectrum of the recurrent updates with (nearly-) orthogonal matrices to control gradients~\citep{arjovsky2016unitary},
but have been found to be less robust across different tasks~\citep{henaff2016recurrent}.

\subsection{Directly related methods}

\paragraph{LMU}
The main result of the Legendre Memory Unit~\citep{voelker2018improving,voelker2019dynamical,voelker2019legendre} is a direct instantiation of our framework using the LegT measure (\cref{subsec:high_order_projection}).
The original LMU is motivated by neurobiological advances and approaches the problem from the opposite direction as us:
it considers approximating spiking neurons in the frequency domain, while we directly solve an interpretable optimization problem in the time domain.
More specifically, they consider time-lagged linear time invariant (LTI) dynamical systems and approximate the dynamics with Pad{\'e} approximants;
\citet{voelker2019legendre} observes that the result also has an interpretation in terms of Legendre polynomials,
but not that it is the optimal solution to a natural projection problem.
This approach involves heavier machinery, and we were not able to find a complete proof of the update mechanism~\citep{voelker2018improving,voelker2019dynamical,voelker2019legendre}.

In contrast, our approach directly poses the relevant online signal approximation problem, which ties to orthogonal polynomial families and leads to simple derivations of several related memory mechanisms (\cref{sec:derivations}). %
Our interpretation in time rather than frequency space, and associated derivation (\cref{sec:derivation-legt}) for the LegT measure, reveals a different set of approximations stemming from the sliding window,
which is confirmed empirically (\cref{sec:exp-hippo-ablations}).

As the motivations of our work are substantially different from \citet{voelker2019legendre}, yet finds the same memory mechanism in a special case,
we highlight the potential connection between these sequence models and biological nervous systems as an area of exploration for future work, such as alternative interpretations of our methods in the frequency domain.

We remark that the term LMU in fact refers to a specific recurrent neural network architecture, which interleaves the projection operator with other specific neural network components.
By contrast, we use HiPPO to refer to the projection operator in isolation (\cref{thm:legt-lagt}), which is a function-to-function or sequence-to-sequence operator independent of model.
HiPPO is integrated into an RNN architecture in \cref{sec:experiments}, with slight improvements to the LMU architecture, as ablated in \cref{sec:pmnist-details,sec:copying-details}.
As a standalone module, HiPPO can be used as a layer in other types of models.

\paragraph{Fourier Recurrent Unit}

The Fourier Recurrent Unit (FRU)~\citep{zhang2018learning} uses Fourier basis
(cosine and sine) to express the input signal, motivated by the discrete Fourier
transform.
In particular, each recurrent unit computes the discrete Fourier transform of
the input signal for a randomly chosen frequency.
It is not clear how discrete transform with respect to other bases (e.g.,
Legendre, Laguerre, Chebyshev) can in turn yield similar memory mechanisms.
We show that FRU is also an instantiation of the HiPPO framework
(\cref{sec:derivation-fourier}), where the Fourier basis can be viewed as
orthogonal polynomials $z^n$ on the unit circle $\{ z \colon \abs{z} = 1 \}$.

\citet{zhang2018learning} prove that if a timescale hyperparameter is chosen appropriately,
FRU has bounded gradients, thus avoiding vanishing and exploding gradients.
This essentially follows from the fact that $(1 - \Delta t)^T = \Theta(1)$ if the discretization
step size $\Delta t = \Theta(\frac{1}{T})$ is chosen, if the time horizon $T$ is known (cf.\ \cref{sec:discretization-full,sec:hippo-theory}).
It is easily shown that this property is not intrinsic to the FRU but to sliding window methods,
and is shared by all of our translated measure HiPPO methods (all but HiPPO-LegS in \cref{sec:derivations}).
We show the stronger property that HiPPO-LegS, which uses scaling rather than sliding windows, also enjoys bounded gradient guarantees,
without needing a well-specified timescale hyperparameter (\cref{prop:gradient-bound}).

\paragraph{Neural ODEs}

HiPPO produces linear ODEs that describe the dynamics of the coefficients.
Recent work has also incorporated ODEs into machine learning models.
\citet{chen2018neural} introduce neural ODEs, employing general nonlinear ODEs
parameterized by neural networks in the context of normalizing flows and time
series modeling.
Neural ODEs have shown promising results in modeling irregularly sampled time
series~\citep{kidger2020neural}, especially when combined with
RNNs~\citep{rubanova2019latent}.
Though neural ODEs are expressive~\citep{dupont2019augmented, zhang2019approximation},
due to their complex parameterization, they often suffer from slow
training~\citep{quaglino2020snode, finlay2020train, massaroli2020stable} because
of their need for more complicated ODE solvers.
On the other hand, HiPPO ODEs are linear and are fast to solve with classical
discretization techniques in linear systems, such as Euler method, Bilinear
method, and Zero-Order Hold (ZOH)~\citep{iserles2009first}.

\section{Technical Preliminaries}
\label{sec:background}

We collect here some technical background that will be used in presenting the
general HiPPO framework and in deriving specific HiPPO update rules.

\subsection{Orthogonal Polynomials}
\label{sec:op}
Orthogonal polynomials are a standard tool for working with function spaces~\cite{chihara,szego}.
Every measure $\mu$ induces a unique (up to a scalar) sequence of \emph{orthogonal polynomials} (OPs) $P_0(x), P_1(x), \dots$ satisfying $\mathrm{deg}(P_i) = i$ and
$ \langle P_i, P_j \rangle_{\mu} \defeq \int P_i(x) P_j(x) \d \mu(x) = 0 $ for all $i \neq j$.
This is the sequence found by orthogonalizing the monomial basis $\{x^i\}$ with Gram-Schmidt with respect to $\langle \cdot, \cdot \rangle_\mu$.
The fact that OPs form an orthogonal basis is useful because the optimal polynomial $g$ of degree $\deg(g) < N$
that approximates a function $f$ is then given by
\begin{align*}%
  \sum_{i=0}^{N-1} c_i P_i(x)/\|P_i\|_\mu^2 \qquad \text{where } c_i = \langle f, P_i \rangle_\mu = \int f(x) P_i(x) \d \mu(x).
\end{align*}
Classical OPs families comprise Jacobi (which include Legendre and Chebyshev
polynomials as special cases), Laguerre, and Hermite polynomials.
The Fourier basis can also be interpreted as OPs on the unit circle in the
complex plane.

\subsubsection{Properties of Legendre Polynomials}
\label{sec:legendre-properties}

\paragraph{Legendre polynomials}

Under the usual definition of the canonical Legendre polynomial $P_n$, they are
orthogonal with respect to the measure $\omega^{\mathrm{leg}} = \mathbf{1}_{[-1,1]}$:
\begin{equation}
  \label{eq:legendre}
  \frac{2n+1}{2} \int_{-1}^1 P_n(x)P_m(x) \dd x = \delta_{nm}
\end{equation}

Also, they satisfy
\begin{align*}%
  P_n(1) &= 1 \\
  P_n(-1) &= (-1)^{n}
  .
\end{align*}

\paragraph{Shifted and Scaled Legendre polynomials}
We will also consider scaling the Legendre polynomials to be orthogonal on the
interval $[0, t]$.
A change of variables on \eqref{eq:legendre} yields
\begin{align*}
  (2n+1) \int_0^t P_n\left( \frac{2x}{t}-1 \right) P_m \left( \frac{2x}{t} - 1 \right) \frac{1}{t} \dd x
  &=
  (2n+1) \int P_n\left( \frac{2x}{t}-1 \right) P_m \left( \frac{2x}{t} - 1 \right) \omega^{\mathrm{leg}}\left( \frac{2x}{t}-1 \right) \frac{1}{t} \dd x
  \\
  &=
  \frac{2n+1}{2} \int P_n(x) P_m(x) \omega^{\mathrm{leg}}(x) \dd x
  \\
  &= \delta_{nm}
  .
\end{align*}
Therefore, with respect to the measure $\omega_t = \mathbf{1}_{[0,t]}/t$ (which is a probability measure for all $t$),
the normalized orthogonal polynomials are
\begin{align*}
  (2n+1)^{1/2} P_n\left( \frac{2x}{t}-1 \right)
  .
\end{align*}

Similarly, the basis
\begin{align*}
  (2n+1)^{1/2} P_n\left( 2\frac{x-t}{\theta} + 1 \right)
\end{align*}
is orthonormal for the uniform measure $\frac{1}{\theta}\mathbb{I}_{[t-\theta, t]}$.

In general, the orthonormal basis for any uniform measure
consists of $(2n+1)^{\frac{1}{2}}$ times the corresponding linearly shifted version of $P_n$.

\paragraph{Derivatives of Legendre polynomials}

We note the following recurrence relations on Legendre polynomials (\citep[Chapter~12]{arfken2005mathematical}):
\begin{align*}
  (2n+1)P_n &= P_{n+1}' - P_{n-1}' \\
  P_{n+1}' &= (n+1)P_n + xP_n'
\end{align*}
The first equation yields
\begin{equation}%
  P_{n+1}' = (2n+1)P_n + (2n-3)P_{n-2} + \dots,
\end{equation}
where the sum stops at $P_0$ or $P_1$.

These equations directly imply
\begin{align}
  P_{n}' &= (2n-1)P_{n-1} + (2n-5)P_{n-3} + \dots
  \label{eq:legendre-d}
\end{align}
and
\begin{align}
  (x+1)P_n'(x) &= P_{n+1}' + P_n' - (n+1)P_n
  \nonumber \\
  &= nP_n + (2n-1)P_{n-1} + (2n-3)P_{n-2} + \dots
  .
  \label{eq:legendre-xd}
\end{align}
These will be used in the derivations of the HiPPO-LegT and HiPPO-LegS updates, respectively.

\subsubsection{Properties of Laguerre Polynomials}
\label{sec:laguerre-properties}

The standard Laguerre polynomials $L_n(x)$ are defined to be orthogonal with
respect to the weight function $e^{-x}$ supported on $[0, \infty)$, while the
generalized Laguerre polynomials (also called associated Laguerre polynomials)
$L_n^{(\alpha)}$ are defined to be orthogonal with respect to the weight function
$x^{\alpha} e^{-x}$ also supported on $[0, \infty)$:
\begin{equation}
  \label{eq:laguerre}
  \int_{0}^{\infty} x^\alpha e^{-x} L_n^{(\alpha)}(x) L_m^{(\alpha)}(x) \d x = \frac{(n+\alpha)!}{n!} \delta_{n, m}.
\end{equation}
Also, they satisfy
\begin{equation}%
  \label{eq:laguerre-endpoint}
  L_n^{(\alpha)}(0) = \binom{n + \alpha}{n} = \frac{\Gamma(n+\alpha+1)}{\Gamma(n+1)\Gamma(\alpha+1)}.
\end{equation}

The standard Laguerre polynomials correspond to the case of $\alpha = 0$ of
generalized Laguerre polynomials.

\paragraph{Derivatives of generalized Laguerre polynomials}

We note the following recurrence relations on generalized Laguerre polynomials (\citep[Chapter~13.2]{arfken2005mathematical}):
\begin{align*}
  \frac{\d}{\d x} L_n^{(\alpha)}(x) &= -L_{n-1}^{(\alpha+1)} (x) \\
  L_{n}^{(\alpha+1)}(x) &= \sum_{i=0}^{n} L_i^{(\alpha)}(x).
\end{align*}
These equations imply
\begin{equation*}
  \ddt L_n^{(\alpha)}(x) = - L_{0}^{(\alpha)}(x) - L_{1}^{(\alpha)} (x) - \dots - L_{n-1}^{(\alpha)}(x).
\end{equation*}

\subsubsection{Properties of Chebyshev polynomials}
\label{sec:chebyshev-properties}

Let $T_n$ be the classical Chebyshev polynomials (of the first kind), defined to
be orthogonal with respect to the weight function $(1-x^2)^{1/2}$ supported on
$[-1, 1]$,
and let $p_n$ be the normalized version of $T_n$ (i.e, with norm 1):
\begin{align*}
  \omega^{\mathrm{cheb}} &= (1-x^2)^{-1/2} \mathbb{I}_{(-1, 1)},
  \\
  p_n(x) &= \sqrt{\frac{2}{\pi}} T_n(x) \qquad \text{for } n \geq 1,
  \\
  p_0(x) &= \frac{1}{\sqrt\pi}.
\end{align*}
Note that $\omega^{\mathrm{cheb}}$ is not normalized (it integrates to $\pi$).

\paragraph{Derivatives of Chebyshev polynomials}
The chebyshev polynomials satisfy
\begin{align*}
  2T_n(x) = \frac{1}{n+1} \frac{d}{d x} T_{n+1}(x) - \frac{1}{n-1} \frac{d}{dx} T_{n-1}(x) \qquad n = 2, 3, \dots
  .
\end{align*}
By telescoping this series, we obtain
\begin{equation}
  \label{eq:chebyshev-derivative}
  \frac{1}{n} T_n' =
  \begin{cases}
    2(T_{n-1} + T_{n-3} + \dots + T_2) + T_0  & n \mbox{ odd}
    \\
    2(T_{n-1} + T_{n-3} + \dots + T_1) & n \mbox{ even}
  \end{cases}
  .
\end{equation}

\paragraph{Translated Chebyshev polynomials}
We will also consider shifting and scaling the Chebyshev polynomials to be orthogonal on the
interval $[t-\theta, t]$ for fixed length $\theta$.

The normalized (probability) measure is
\begin{align*}%
  \omega(t, x) &= \frac{2}{\theta\pi} \omega^{\mathrm{cheb}}\left( \frac{2(x-t)}{\theta} + 1 \right)
  = \frac{1}{\theta\pi} \left( \frac{x-t}{\theta} + 1 \right)^{-1/2} \left( -\frac{x-t}{\theta} \right)^{-1/2}
  \mathbb{I}_{(t-\theta, t)}
  .
\end{align*}
The orthonormal polynomial basis is
\begin{align*}%
  p_n(t, x) &= \sqrt\pi p_n\left( \frac{2(x-t)}{\theta} + 1 \right)
  .
\end{align*}
In terms of the original Chebyshev polynomials, these are
\begin{align*}%
  p_n(t, x) &= \sqrt{2} T_n\left( \frac{2(x-t)}{\theta} + 1 \right) \qquad \text{for } n \geq 1,
  \\
  p_0^{(t)} &= T_0\left( \frac{2(x-t)}{\theta} + 1 \right).
\end{align*}

\subsection{Leibniz Integral Rule}
As part of our standard strategy for deriving HiPPO update rules (\cref{sec:hippo-framework-details}),
we will differentiate through integrals with changing limits.
For example, we may wish to differentiate with respect to $t$ the expression
$\int f(t, x) \mu(t, x) \dd x = \int_0^t f(t, x) \frac{1}{t} \d x$
when analyzing the scaled Legendre (LegS) measure.

Differentiating through such integrals can be formalized by the Leibniz integral rule,
the basic version of which states that
\[
  \frac{\partial}{\partial t} \int_{\alpha(t)}^{\beta(t)} f(x, t) \dd x
  = \int_{\alpha(t)}^{\beta(t)} \frac{\partial}{\partial t} f(x, t) \dd x
  - \alpha'(t) f(\alpha(t), t) + \beta'(t) f(\beta(t), t).
\]

We elide over the formalisms in our derivations (\cref{sec:derivations}) and
instead use the following trick.
We replace integrand limits with an indicator function;
and using the Dirac delta function $\delta$ when differentiating (i.e., using the formalism
of distributional derivatives).
For example, the above formula can be derived succinctly with this trick:
\begin{align*}%
  \frac{\partial}{\partial t} \int_{\alpha(t)}^{\beta(t)} f(x, t) \dd x
  &= \frac{\partial}{\partial t} \int f(x, t) \mathbb{I}_{[\alpha(t), \beta(t)]}(x) \dd x
  \\
  &= \int \frac{\partial}{\partial t} f(x, t) \mathbb{I}_{[\alpha(t), \beta(t)]}(x) \dd x
  + \int f(x, t) \frac{\partial}{\partial t} \mathbb{I}_{[\alpha(t), \beta(t)]}(x) \dd x
  \\
  &= \int \frac{\partial}{\partial t} f(x, t) \mathbb{I}_{[\alpha(t), \beta(t)]}(x) \dd x
  + \int f(x, t) (\beta'(t) \delta_{\beta(t)}(x) - \alpha'(t) \delta_{\alpha(t)})(x) \dd x
  \\
  &= \int_{\alpha(t)}^{\beta(t)} \frac{\partial}{\partial t} f(x, t) \dd x
  - \alpha'(t) f(\alpha(t), t) + \beta'(t) f(\beta(t), t)
  .
\end{align*}

\subsection{ODE Discretization}
\label{sec:discretization-full}

In our framework, time series inputs will be modeled with a continuous function and then discretized.
Here we provide some background on ODE discretization methods,
including a new discretization that applies to a specific type of ODE that our new method encounters.

The general formulation of an ODE is $\frac{d}{dt} c(t) = f(t, c(t))$.
We will also focus on the linear time-invariant ODE of the form $\frac{d}{dt} c(t) = A c(t) + Bf(t)$
for some input function $f(t)$, as a special case.
The general methodology for discretizing the ODE, for step size $\Delta t$, is to
rewrite the ODE as
\begin{equation}
  \label{eq:ode-rhs}
  c(t + \Delta t) - c(t) = \int_t^{t+\Delta t} f(s, c(s)) \dd s,
\end{equation}
then approximate the RHS integral.

Many ODE discretization methods corresponds to different ways to
approximate the RHS integral:

\paragraph{Euler (aka forward Euler).}
To approximate the RHS of equation~\eqref{eq:ode-rhs}, keep the left endpoint
$\Delta t f(t, c(t))$.
For the linear ODE, we get:
\begin{equation*}
  c(t + \Delta t) = (I + \Delta t A) c(t)+ \Delta t B f(t).
\end{equation*}

\paragraph{Backward Euler.}
To approximate the RHS of equation~\eqref{eq:ode-rhs}, keep the right endpoint
$\Delta t f(t+\Delta t, c(t+\Delta t))$.
For the linear ODE, we get the linear equation and the update:
\begin{align*}
  c(t + \Delta t) - \Delta t A c(t + \Delta t) &= c(t)+ \Delta t B f(t) \\
  c(t + \Delta t) &= (I - \Delta t A)^{-1} c(t) + \Delta t (I - \Delta t A)^{-1} B f(t).
\end{align*}

\paragraph{Bilinear (aka Trapezoid rule, aka Tustin's method).}
To approximate the RHS of equation~\eqref{eq:ode-rhs}, average the endpoints
$\Delta t \frac{f(t, c(t))+f(t+\Delta t, c(t+\Delta t))}{2}$.
For the linear ODE, again we get a linear equation and the update:
\begin{align*}
  c(t + \Delta t) - \frac{\Delta t}{2} A c(t + \Delta t) &= (I + \Delta t/2 A) c(t) + \Delta t B f(t) \\
  c(t + \Delta t) &= (I - \Delta t/2 A)^{-1}(I + \Delta t/2 A) c(t) + \Delta t (I - \Delta t/2 A)^{-1} B f(t).
\end{align*}

\paragraph{Generalized Bilinear Transformation (GBT).}
This method~\citep{zhang2007performance} approximates the RHS of
equation~\eqref{eq:ode-rhs} by taking a weighted average of the endpoints
$\Delta t [(1 - \alpha) f(t, c(t)) + \alpha f(t+\Delta t, c(t+\Delta t))]$, for some parameter $\alpha \in [0, 1]$.
For the linear ODE, again we get a linear equation and the update:
\begin{align}
  c(t + \Delta t) - \Delta t \alpha A c(t + \Delta t) &= (I + \Delta t (1 - \alpha) A) c(t) + \Delta t B f(t) \nonumber \\
  c(t + \Delta t) &= (I - \Delta t \alpha A)^{-1}(I + \Delta t (1 - \alpha) A) c(t) + \Delta t (I - \Delta t \alpha A)^{-1} B f(t).
  \label{eq:gbt}
\end{align}
GBT generalizes the three methods mentioned above: forward Euler corresponds to
$\alpha = 0$, backward Euler to $\alpha = 1$, and bilinear to $\alpha = 1/2$.

We also note another method called \emph{Zero-order Hold}
(ZOH)~\citep{decarlo1989linear} that specializes to linear ODEs.
The RHS of equation~\eqref{eq:ode-rhs} is calculated in closed-form assuming
constant input $f$ between $t$ and $t + \Delta t$.
This yields the update
$c(t + \Delta t) = e^{\Delta t A} c(t) + \left( \int_{\tau=0}^{\Delta t} e^{\tau A} \d \tau \right) B f(t)$.
If $A$ is invertible, this can be simplified as
$c(t + \Delta t) = e^{\Delta t A} c(t) + A^{-1} (e^{\Delta t A} - I) B f(t)$.

\paragraph{HiPPO-LegS invariance to discretization step size.}
In the case of HiPPO-LegS, we have a linear ODE of the form
$\frac{d}{dt} c(t) = \frac{1}{t} A c(t) + \frac{1}{t} Bf(t)$.
Adapting the GBT discretization (which generalizes forward/backward Euler and
bilinear) to this linear ODE, we obtain:
\begin{align*}
  c(t + \Delta t) - \Delta t \alpha \frac{1}{t + \Delta t} A c(t + \Delta t) &= \left(I + \Delta t (1 - \alpha) \frac{1}{t} A\right) c(t) + \Delta t \frac{1}{t} B f(t) \\
  c(t + \Delta t) &= \left(I - \frac{\Delta t}{t + \Delta t}\alpha A\right)^{-1}\left(I + \frac{\Delta t}{t} (1 - \alpha) A\right) c(t)
               + \frac{\Delta t}{t} \left(I - \frac{\Delta t}{t + \Delta t} \alpha A\right)^{-1} B f(t).
\end{align*}
We highlight that this system is invariant to the discretization step size
$\Delta t$.
Indeed, if $c^{(k)} \defeq c(k \Delta t)$ and $f_k \defeq f(k \Delta t)$ then we have the recurrence
\begin{equation*}
  c^{(k+1)} = \left(I - \frac{1}{k+1}\alpha A\right)^{-1}\left(I + \frac{1}{k} (1 - \alpha) A\right) c^{(k)}
  + \frac{1}{k} \left(I - \frac{1}{k+1} \alpha A\right)^{-1} B f_k,
\end{equation*}
which does not depend on $\Delta t$.

\begin{figure}[ht]
  \centering
  \includegraphics[width=0.5\linewidth]{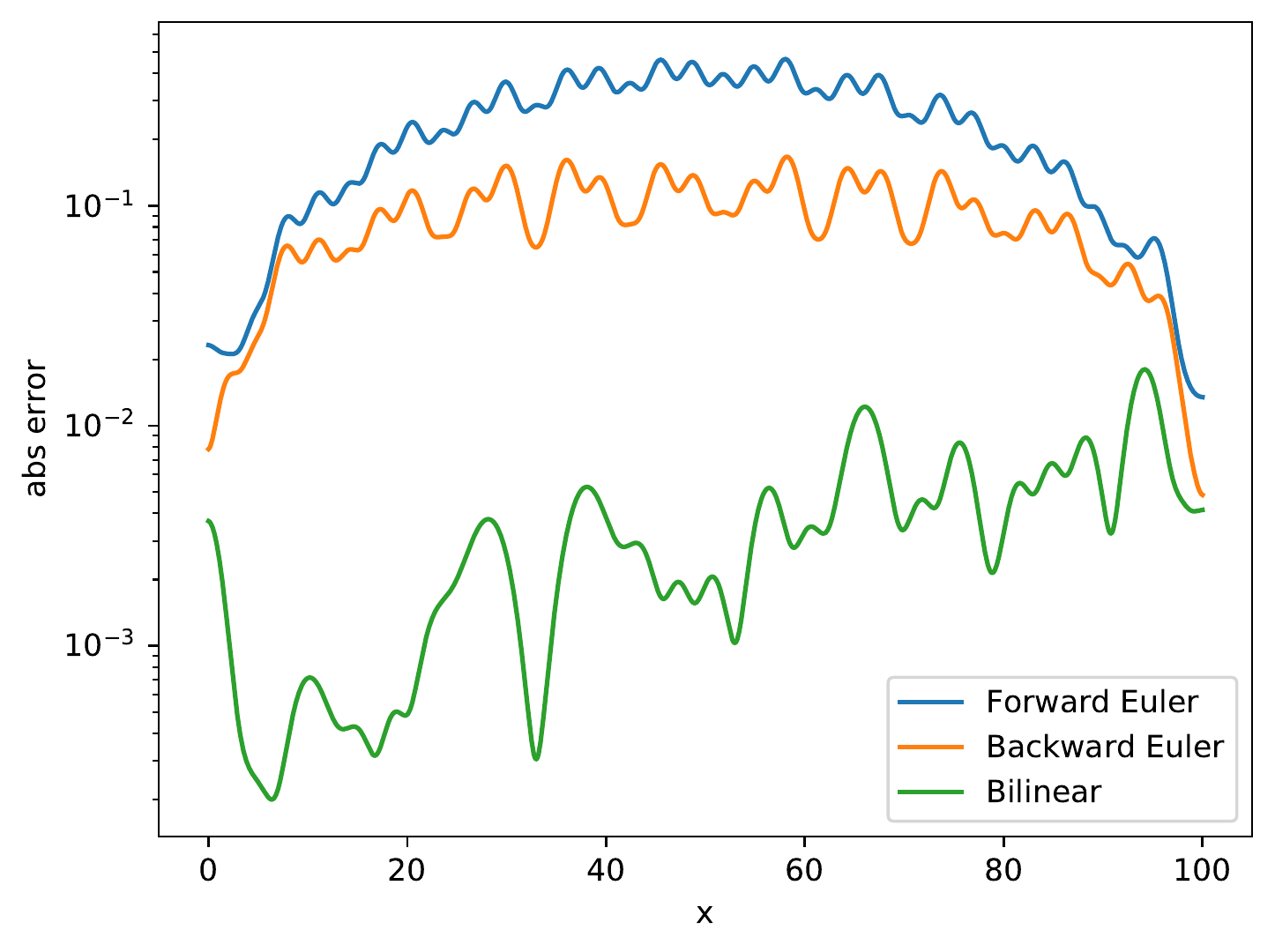}
  \caption{Absolute error for different discretization methods. Forward and
    backward Euler are generally not very accurate, while bilinear
  yields more accurate approximation.}
  \label{fig:discretization_error}
\end{figure}

\paragraph{Ablation: comparison between different discretization methods}
To understand the impact of approximation error in discretization, in
\cref{fig:discretization_error}, we show the absolute error for the HiPPO-LegS
updates in function approximation (\cref{sec:exp-hippo-ablations})
for different discretization methods: forward Euler, backward Euler,
and bilinear.
The bilinear method generally provide sufficiently accurate approximation.
We will use bilinear as the discretization method for the LegS updates
for the experiments.

\section{General HiPPO Framework}
\label{sec:hippo-framework-details}

We present the general HiPPO framework, as described in \cref{sec:framework}, in
more details.
We also generalize it to include bases other than polynomials.

Given a time-varying measure family $\mu^{(t)}$ supported on $(-\infty, t]$,
a sequence of basis functions $\mathcal{G} = \mathrm{span} \{g_n^{(t)}\}_{n \in [N]}$,
and a continuous function $f \colon \mathbb{R}_{\geq 0} \to \mathbb{R}$,
HiPPO defines an operator that maps $f$ to the optimal projection coefficients
$c: \R_{\geq 0} \to \R^N$, such that
\begin{equation*}
  g^{(t)} \defeq \mathrm{argmin}_{g \in \mathcal{G}} \norm{f_{\leq t} - g}_{\mu^{(t)}},
  \qquad \text{and} \qquad
  g^{(t)} = \sum_{n=0}^{N-1} c_n(t) g_n^{(t)}.
\end{equation*}
The first step refers to the $\proj_t$ operator and the second the $\coef_t$
operator in \cref{def:hippo}.

We focus on the case where the coefficients $c(t)$ has the form of a linear
ODE satisfying $\frac{d}{dt} c(t) = A(t) c(t) + B(t) f(t)$ for some
$A(t) \in \R^{N \times N}$, $B(t) \in \R^{N \times 1}$.

We first describe the parameters of the $\hippo$ operator (a measure and basis) in more detail in \cref{sec:hippo-measure}.
We define the projection $\proj_t$ and coefficient $\coef_t$ operators in \cref{sec:hippo-calculation}.
Then we give a general strategy to calculate these coefficients $c(t)$, by deriving a differential equation that governs the coefficient dynamics (\cref{sec:hippo-dynamics}).
Finally we discuss how to turn the continuous $\hippo$ operator into a discrete
one that can be applied to sequence data (\cref{sec:framework-discretization}).

\subsection{Measure and Basis}
\label{sec:hippo-measure}

We describe and motivate the ingredients of HiPPO in more detail here.
Recall that the high level goal is online function approximation;
this requires both a set of valid approximations and a notion of approximation quality.

\textbf{Approximation Measures}
At every $t$, the approximation quality is defined with respect to a measure $\mu^{(t)}$ supported on $(-\infty, t]$.
We seek some polynomial $g^{(t)}$ of degree at most $N-1$ that minimizes the
error $\|f_{x \leq t} - g^{(t)}\|_{L_2(\mu^{(t)})}$.
Intuitively, this measure $\mu^{(t)}$ governs how much to weigh every time in the past.
For simplicity, we assume that the measures $\mu^{(t)}$ are sufficiently smooth across their domain as well as in time;
in particular, they have densities $\omega(t, x) := \frac{\dd \mu^{(t)}}{\dd \lambda}(x)$
with respect to the Lebesgue measure $\dd \lambda(x) \defeq \dd x$
such that $\omega$ is $C^1$ almost everywhere.
Thus integrating against $\dd \mu^{(t)}(x)$ can be rewritten as integrating against
$\omega(t, x) \dd x$.

We also assume for simplicity that the measures $\mu^{(t)}$ are normalized to be probability measures;
arbitrary scaling does not affect the optimal projection.

\textbf{Orthogonal polynomial basis}
Let $\{P_n\}_{n \in \N}$ denote a sequence of orthogonal polynomials with respect to some base measure $\mu$.
Similarly define $\{ P_n^{(t)} \}_{n \in \mathbb{N}}$ to be a sequence of
orthogonal polynomials with respect to the time-varying measure $\mu^{(t)}$.
Let $p_n^{(t)}$ be the normalized version of $P_n^{(t)}$ (i.e., have norm 1),
and define
\begin{equation}
  \label{eq:framework-p}
  p_n(t, x) = p_n^{(t)}(x).
\end{equation}
Note that the $P_n^{(t)}$ are not required to be normalized, while the
$p_n^{(t)}$ are.

\textbf{Tilted measure and basis}
Our goal is simply to store a compressed representation of functions, which can use any basis, not necessarily OPs.
For any scaling function
\begin{equation}
  \label{eq:framework-chi}
  \chi(t, x) = \chi^{(t)}(x)
  ,
\end{equation}
the functions $p_n(x) \chi(x)$ are orthogonal with respect to the density $\omega / \chi^2$ at every time $t$.
Thus, we can choose this alternative basis and measure to perform the projections.

To formalize this tilting with $\chi$, define $\nu^{(t)}$ to be the normalized measure with density proportional to $\omega^{(t)} / (\chi^{(t)})^2$.

We will calculate the normalized measure and the orthonormal basis for it.
Let
\begin{equation}
  \label{eq:framework-zeta}
  \zeta(t) = \int \frac{\omega}{\chi^2} = \int \frac{\omega^{(t)}(x)}{(\chi^{(t)}(x))^2} \dd x
\end{equation}
be the normalization constant,
so that $\nu^{(t)}$ has density $\frac{\omega^{(t)}}{\zeta(t) (\chi^{(t)})^2}$.
If $\chi(t, x) = 1$ (no tilting), this constant is $\zeta(t) = 1$.
In general, we assume that $\zeta$ is constant for all $t$; if not, it can
be folded into $\chi$ directly.

Next, note that (dropping the dependence on $x$ inside the integral for shorthand)
\begin{align*}
  \left\| \zeta(t)^{\frac{1}{2}} p_n^{(t)} \chi^{(t)} \right\|^2_{\nu^{(t)}}
  &= \int \left( \zeta(t)^{\frac{1}{2}} p_n^{(t)} \chi^{(t)} \right)^2 \frac{\omega^{(t)}}{\zeta(t) (\chi^{(t)})^2}
  \\
  &= \int (p_n^{(t)})^2 \omega^{(t)}
  \\
  &= \left\| p_n^{(t)} \right\|^2_{\mu^{(t)}} = 1.
\end{align*}

Thus we define the orthogonal basis for $\nu^{(t)}$
\begin{align}
  \label{eq:framework-g}
  g_n^{(t)} &= \lambda_n \zeta(t)^{\frac{1}{2}} p_n^{(t)} \chi^{(t)}, \quad n \in \mathbb{N}
  .
\end{align}
We let each element of the basis be scaled by a $\lambda_n$ scalar, for reasons discussed soon,
since arbitrary scaling does not change orthogonality:
\begin{align*}
  \langle g_n^{(t)},\ g_m^{(t)} \rangle_{\nu^{(t)}} &= \lambda_n^2 \delta_{n, m}
\end{align*}
Note that when $\lambda_n = \pm 1$, the basis $\{g_n^{(t)}\}$ is an orthonormal basis with respect to the measure $\nu^{(t)}$,
at every time $t$.
Notationally, let $g_n(t, x) := g_n^{(t)}(x)$ as usual.

We will only use this tilting in the case of Laguerre (\cref{sec:derivation-lagt} and Chebyshev (\cref{sec:derivation-chebyshev}).

Note that in the case $\chi = 1$ (i.e., no tilting),
we also have $\zeta = 1$ and $g_n = \lambda_n p_n$ (for all $t, x$).

\subsection{The Projection and Coefficients}
\label{sec:hippo-calculation}

Given a choice of measures and basis functions, we next see how the coefficients
$c(t)$ can be computed.

\paragraph{Input: Function}
We are given a $C^1$-smooth function $f: [0, \infty) \to \R$ which is seen \emph{online}, for which we wish to maintain a compressed representation of its history $f(x)_{\le t} = f(x)_{x \le t}$ at every time $t$.

\paragraph{Output: Approximation Coefficients}
The function $f$ can be approximated by storing its coefficients with respect to the basis $\{g_n\}_{n < N}$.
For example, in the case of no tilting $\chi = 1$, this encodes the optimal polynomial approximation of $f$ of degree less than $N$.
In particular, at time $t$ we wish to represent $f_{\leq t}$ as a linear combination of polynomials $g_n^{(t)}$.
Since the $g_n^{(t)}$ are orthogonal with respect to the Hilbert
space defined by $\langle \cdot, \cdot \rangle_{\nu^{(t)}}$, it suffices to calculate coefficients

\begin{equation}
  \label{eq:coefficient}
  \begin{aligned}
    c_n(t) &= \langle f_{\leq t}, g_n^{(t)} \rangle_{\nu^{(t)}}
    \\
    &= \int f g_n^{(t)} \frac{\omega^{(t)}}{\zeta(t) (\chi^{(t)})^2}
    \\
    &= \zeta(t)^{-\frac{1}{2}} \lambda_n \int f p_n^{(t)} \frac{\omega^{(t)}}{\chi^{(t)}}
    .
  \end{aligned}
\end{equation}

\paragraph{Reconstruction}
At any time $t$, $f_{\leq t}$ can be explicitly reconstructed as
\begin{equation}
  \label{eq:reconstruction}
  \begin{aligned}
    f_{\leq t} \approx g^{(t)}
    &\defeq \sum_{n=0}^{N-1} \langle f_{\leq t}, g_n^{(t)} \rangle_{\nu^{(t)}} \frac{g_n^{(t)}}{\| g_n^{(t)} \|_{\nu^{(t)}}^2}
    \\
    &= \sum_{n=0}^{N-1} \lambda_n^{-2} c_n(t) g_n^{(t)}
    \\
    &= \sum_{n=0}^{N-1} \lambda_n^{-1} \zeta^{\frac{1}{2}} c_n(t) p_n^{(t)} \chi^{(t)}
    .
  \end{aligned}
\end{equation}

Equation~\eqref{eq:reconstruction} is the $\proj_t$ operator;
given the measure and basis parameters, it defines the optimal approximation of $f_{\leq t}$.

The $\coef_t$ operator simply extracts the vector of coefficients $c(t) = (c_n(t))_{n \in [N]}$.

\subsection{Coefficient Dynamics: the $\hippo$ Operator}
\label{sec:hippo-dynamics}

For the purposes of end-to-end models consuming an input function $f(t)$, the coefficients $c(t)$ are enough to encode information about the history of $f$ and allow online predictions.
Therefore, defining $c(t)$ to be the vector of $c_n(t)$ from equation~\eqref{eq:coefficient},
our focus will be on how to calculate the function $c : \R_{\ge 0} \to \mathbb{R}^N$ from the input function $f : \mathbb{R}_{\ge 0} \to \mathbb{R}$.

In our framework, we will compute these coefficients over time by viewing them as a dynamical system.
Differentiating \eqref{eq:coefficient},
\begin{equation}
  \label{eq:coefficient-dynamics}
  \begin{aligned}
    \ddt c_n(t)
    &= \zeta(t)^{-\frac{1}{2}} \lambda_n \int f(x) \left(\ppt p_n(t, x)\right) \frac{\omega}{\chi}(t, x) \dd x
    \\
    &\quad + \int f(x) \left( \zeta^{-\frac{1}{2}} \lambda_n p_n(t, x) \right) \left(\ppt \frac{\omega}{\chi}(t, x)\right) \dd x
    .
  \end{aligned}
\end{equation}
Here we have made use of the assumption that $\zeta$ is constant for all $t$.

Let $c(t) \in \R^{N-1}$ denote the vector of all coefficients $(c_n(t))_{0 \le n < N}$.

The key idea is that if $\ppt P_n$ and $\ppt \frac{\omega}{\chi}$ have closed forms that can be related back to the polynomials $P_k$, then an ordinary differential equation can be written for $c(t)$.
This allows these coefficients $c(t)$ and hence the optimal polynomial approximation to be computed online.
Since $\ddt P_n^{(t)}$ is a polynomial (in $x$) of degree $n-1$, it can be written as linear
combinations of $P_0, \dots, P_{n-1}$, so the first term in
\cref{eq:coefficient-dynamics} is a linear combination of $c_0, \dots, c_{n-1}$.
For many weight functions $\omega$, we can find scaling function $\chi$ such that $\ppt \frac{\omega}{\chi}$ can also be written in terms of $\frac{\omega}{\chi}$
itself, and thus in those cases the second term of \cref{eq:coefficient-dynamics} is also
a linear combination of $c_0, \dots, c_{N-1}$ and the input $f$.
Thus this often yields a closed-form linear ODE for $c(t)$.

\paragraph{Normalized dynamics}

Our purpose of defining the free parameters $\lambda_n$ was threefold.
\begin{enumerate}%
  \item First, note that the orthonormal basis is not unique, up to a $\pm 1$ factor per element.
  \item Second, choosing $\lambda_n$ can help simplify the derivations.
  \item Third, although choosing $\lambda_n = \pm 1$ will be our default, since projecting onto an orthonormal basis is most sensible, the LMU~\citep{voelker2019legendre} used a different scaling. \cref{sec:derivation-legt} will recover the LMU by choosing different $\lambda_n$ for the LegT measure.
\end{enumerate}

Suppose that equation~\eqref{eq:coefficient-dynamics} reduced to dynamics of the form
\begin{align*}
  \ddt c(t) &= -A(t) c(t) + B(t) f(t).
\end{align*}

Then, letting $\Lambda = \diag_{n \in [N]} \{\lambda_n\}$,
\begin{align*}
  \ddt \Lambda^{-1} c(t) &= -\Lambda^{-1} A(t) \Lambda \Lambda^{-1} c(t) + \Lambda^{-1} B(t) f(t).
\end{align*}

Therefore, if we reparameterize the coefficients ($\Lambda^{-1} c(t) \to c(t)$) then the
\emph{normalized} coefficients projected onto the orthonormal basis satisfy dynamics
and associated reconstruction
\begin{align}
  \label{eq:coefficient-dynamics-normalized}
  \ddt c(t) &= -(\Lambda^{-1} A(t) \Lambda) c(t) + (\Lambda^{-1} B(t)) f(t)
  \\
  \label{eq:reconstruction-normalized}
  f_{\leq t} \approx g^{(t)}
  &= \sum_{n=0}^{N-1} \zeta^{\frac{1}{2}} c_n(t) p_n^{(t)} \chi^{(t)}
\end{align}
These are the $\hippo$ and $\proj_t$ operators.

\subsection{Discretization}
\label{sec:framework-discretization}

As defined here, $\hippo$ is a map on continuous functions.
However, as $\hippo$ defines a closed-form ODE of the coefficient dynamics,
standard ODE discretization methods (\cref{sec:discretization-full}) can be applied to turn this into discrete memory updates.
Thus we overload these operators, i.e.\ $\hippo$ either defines an ODE
of the form
\begin{align*}
  \frac{d}{dt} c(t) = A(t) c(t) + B(t) f(t)
\end{align*}
or a recurrence
\begin{align*}
  c_{t} = A_t c_{t-1} + B_t f_t,
\end{align*}
whichever is clear from context.

\cref{subsec:function_approx} validates the framework by applying \eqref{eq:coefficient-dynamics} and \eqref{eq:reconstruction} to approximate a synthetic function.

\section{Derivations of HiPPO Projection Operators}
\label{sec:derivations}

We derive the memory updates associated with the translated Legendre (LegT)
and translated Laguerre (LagT) measures as presented in
\cref{subsec:high_order_projection}, along with the scaling Legendre (LegS)
measure (\cref{sec:theory_legs}).
To show the generality of the framework, we also derive memory updates with
Fourier basis (recovering the Fourier Recurrent Unit~\citep{zhang2018learning})
and with Chebyshev basis.

The majority of the work has already been accomplished by setting up the
projection framework, and the proof simply requires following the technical
outline laid out in \cref{sec:hippo-framework-details}.
In particular, the definition of the coefficients \eqref{eq:coefficient} and reconstruction \eqref{eq:reconstruction} does not change,
and we only consider how to calculate the coefficients dynamics \eqref{eq:coefficient-dynamics}.

For each case, we follow the general steps:
\begin{description}%
  \item[Measure and Basis]
    define the measure $\mu^{(t)}$ or weight $\omega(t, x)$ and basis functions $p_n(t, x)$,
  \item[Derivatives]
    compute the derivatives of the measure and basis functions,
  \item[Coefficient Dynamics]
    plug them into the coefficient dynamics
    (equation~\eqref{eq:coefficient-dynamics}) to derive the ODE that describes how
    to compute the coefficients $c(t)$,
  \item[Reconstruction]
    provide the complete formula to reconstruct an approximation to the function $f_{\le t}$,
    which is the optimal projection under this measure and basis.
\end{description}

The derivations in \cref{sec:derivation-legt,sec:derivation-lagt}
prove \cref{thm:legt-lagt}, and the derivations in \cref{sec:derivation-legs}
prove \cref{thm:legs}.
\cref{sec:derivation-fourier,sec:derivation-chebyshev} show additional results for Fourier-based bases.

\begin{figure}
  \centering
  \includegraphics[width=\linewidth]{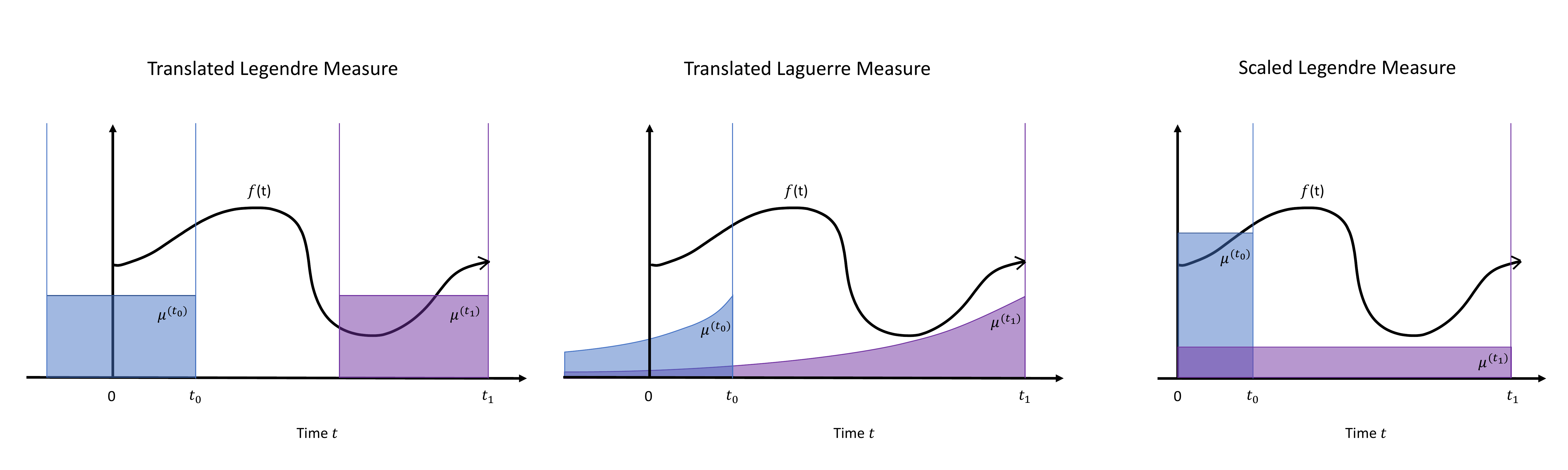}
  \caption{
    \textbf{Illustration of HiPPO measures.}
    At time $t_0$, the history of a function $f(x)_{x \le t_0}$ is summarized by polynomial approximation with respect to the measure $\mu^{(t_0)}$ (blue), and similarly for time $t_1$ (purple).
    (Left) The Translated Legendre measure (\textbf{LegT}) assigns weight in the window $[t-\theta, t]$.
    For small $t$, $\mu^{(t)}$ is supported on a region $x < 0$ where $f$ is not defined.
    When $t$ is large, the measure is not supported near $0$, causing the projection of $f$ to forget the beginning of the function.
    (Middle) The Translated Laguerre (\textbf{LagT}) measure decays the past exponentially.
    It does not forget, but also assigns weight on $x < 0$.
    (Right) The Scaled Legendre measure (\textbf{LegS}) weights the entire history $[0, t]$ uniformly.
  }
  \label{fig:measures}
\end{figure}

Figure~\ref{fig:measures} illustrates the overall framework when we use Legendre
and Laguerre polynomials as the basis,
contrasting our main families of time-varying measures $\mu^{(t)}$.

\IfFileExists{./derivations/legt.tex}{

\subsection{Derivation for Translated Legendre (HiPPO-LegT)}
\label{sec:derivation-legt}

This measure fixes a window length $\theta$ and slides it across time.

\paragraph{Measure and Basis}
We use a uniform weight function supported on the interval $[t - \theta, t]$ and pick
Legendre polynomials $P_n(x)$, translated from $[-1, 1]$ to $[t - \theta, t]$, as basis
functions:
\begin{align*}
  \omega(t, x) &= \frac{1}{\theta} \mathbb{I}_{[t-\theta, t]}
  \\
  p_n(t, x) &= (2n+1)^{1/2} P_n\left( \frac{2(x-t)}{\theta} + 1 \right)
  \\
  g_n(t, x) &= \lambda_n p_n(t, x)
  .
\end{align*}
Here, we have used no tilting so $\chi = 1$ and $\zeta = 1$ (equations~\eqref{eq:framework-chi} and~\eqref{eq:framework-zeta}).
We leave $\lambda_n$ unspecified for now.

At the endpoints, these basis functions satisfy
\begin{align*}
  g_n(t, t) &= \lambda_n (2n+1)^{\frac{1}{2}}
  \\
  g_n(t, t-\theta) &= \lambda_n (-1)^{n} (2n+1)^{\frac{1}{2}}
  .
\end{align*}

\paragraph{Derivatives}
The derivative of the measure is
\begin{align*}
  \ppt \omega(t, x) &= \frac{1}{\theta} \delta_t - \frac{1}{\theta} \delta_{t-\theta}
  .
\end{align*}
The derivative of Legendre polynomials can be expressed as linear combinations of
other Legendre polynomials (cf.\ \cref{sec:legendre-properties}).
\begin{align*}
  \ppt g_n(t, x) &= \lambda_n (2n+1)^{\frac{1}{2}} \cdot \frac{-2}{\theta} P_n'\left( \frac{2(x-t)}{\theta} + 1 \right)
  \\
  &= \lambda_n (2n+1)^{\frac{1}{2}} \frac{-2}{\theta} \left[ (2n-1)P_{n-1}\left( \frac{2(x-t)}{\theta} + 1 \right) + (2n-5) P_{n-3}\left( \frac{2(x-t)}{\theta} + 1 \right) + \dots \right]
  \\
  &= - \lambda_n (2n+1)^{\frac{1}{2}} \frac{2}{\theta} \left[ \lambda_{n-1}^{-1} (2n-1)^{\frac{1}{2}} g_{n-1}(t, x) + \lambda_{n-3}^{-1} (2n-3)^{\frac{1}{2}} g_{n-3}(t, x) + \dots \right]
  .
\end{align*}
We have used equation~\eqref{eq:legendre-d} here.

\paragraph{Sliding Approximation}
As a special case for the LegT measure, we need to consider an approximation due to the nature of the sliding window measure.

When analyzing $\ddt c(t)$ in the next section, we will need to use the value $f(t-\theta)$.
However, at time $t$ this input is no longer available.
Instead, we need to rely on our compressed representation of the function:
by the reconstruction equation \eqref{eq:reconstruction}, if the approximation is succeeding so far,
we should have
\begin{align*}
  f_{\leq t}(x) &\approx \sum_{k=0}^{N-1} \lambda_k^{-1} c_k(t) (2k+1)^{\frac{1}{2}} P_k \left( \frac{2(x-t)}{\theta} + 1 \right)
  \\
  f(t-\theta) &\approx \sum_{k=0}^{N-1} \lambda_k^{-1} c_k(t) (2k+1)^{\frac{1}{2}} (-1)^{k}
  .
\end{align*}

\paragraph{Coefficient Dynamics}
We are ready to derive the coefficient dynamics.

Plugging the derivatives of this measure and basis into equation~\eqref{eq:coefficient-dynamics} gives
\begin{align*}
  \frac{d}{dt} c_n(t)
  &= \int f(x) \left(\ppt g_n(t, x)\right) \omega(t, x) \dd x
  \\
  &\qquad + \int f(x) g_n(t, x) \left(\ppt \omega(t, x)\right) \dd x
  \\
  &= -\lambda_n (2n+1)^{\frac{1}{2}} \frac{2}{\theta} \left[ \lambda_{n-1}^{-1} (2n-1)^{\frac{1}{2}} c_{n-1}(t) + \lambda_{n-3}^{-1} (2n-5)^{\frac{1}{2}} c_{n-3}(t) + \dots \right]
  \\
  &\qquad + \frac{1}{\theta} f(t) g_n(t, t) - \frac{1}{\theta} f(t-\theta) g_n(t, t-\theta)
  \\
  &\approx -\frac{\lambda_n}{\theta} (2n+1)^{\frac{1}{2}} \cdot 2 \left[ (2n-1)^{\frac{1}{2}} \frac{c_{n-1}(t)}{\lambda_{n-1}} + (2n-5)^{\frac{1}{2}} \frac{c_{n-3}(t)}{\lambda_{n-3}} + \dots \right]
  \\
  &\qquad + (2n+1)^{\frac{1}{2}} \frac{\lambda_n}{\theta} f(t)
  - (2n+1)^{\frac{1}{2}} \frac{\lambda_n}{\theta} (-1)^n \sum_{k=0}^{N-1} (2k+1)^{\frac{1}{2}} \frac{c_k(t)}{\lambda_k} (-1)^{k}
  \\
  &= - \frac{\lambda_n}{\theta} (2n+1)^{\frac{1}{2}} \cdot 2 \left[ (2n-1)^{\frac{1}{2}} \frac{c_{n-1}(t)}{\lambda_{n-1}} + (2n-5)^{\frac{1}{2}} \frac{c_{n-3}(t)}{\lambda_{n-3}} + \dots \right]
  \\
  &\qquad - (2n+1)^{\frac{1}{2}} \frac{\lambda_n}{\theta} \sum_{k=0}^{N-1}  (-1)^{n-k} (2k+1)^{\frac{1}{2}} \frac{c_k(t)}{\lambda_{k}}
  + (2n+1)^{\frac{1}{2}} \frac{\lambda_n}{\theta} f(t)
  \\
  &= - \frac{\lambda_n}{\theta} (2n+1)^{\frac{1}{2}} \sum_{k=0}^{N-1}  M_{nk} (2k+1)^{\frac{1}{2}} \frac{c_k(t)}{\lambda_{k}}
  + (2n+1)^{\frac{1}{2}} \frac{\lambda_n}{\theta} f(t)
  ,
\end{align*}
where
\begin{align*}
  M_{nk} = 
  \begin{cases}
    1 & \mbox{if } k \le n \\
    (-1)^{n-k} & \mbox{if } k \ge n
  \end{cases}.
\end{align*}

Now we consider two instantiations for $\lambda_n$.
The first one is the more natural $\lambda_n = 1$, which turns $g_n$ into an orthonormal basis.
We then get
\begin{align*}
  \ddt c(t) &= -\frac{1}{\theta} A c(t) + \frac{1}{\theta} B f(t)
  \\
  A_{nk} &=
  (2n+1)^{\frac{1}{2}}(2k+1)^{\frac{1}{2}}
  \begin{cases}
    1 & \mbox{if } k \le n \\
    (-1)^{n-k} & \mbox{if } k \ge n
  \end{cases}
  \\
  B_n &= (2n+1)^{\frac{1}{2}}
  .
\end{align*}

The second case takes $\lambda_n = (2n+1)^{\frac{1}{2}} (-1)^{n}$.
This yields
\begin{align*}
  \ddt c(t) &= -\frac{1}{\theta} A c(t) + \frac{1}{\theta} B f(t)
  \\
  A_{nk} &=
  (2n+1)
  \begin{cases}
    (-1)^{n-k} & \mbox{if } k \le n \\
    1          & \mbox{if } k \ge n
  \end{cases}
  \\
  B_n &= (2n+1) (-1)^{n}
  .
\end{align*}
This is exactly the LMU update equation.

\paragraph{Reconstruction}

By equation~\eqref{eq:reconstruction}, at every time $t$ we have
\begin{align*}
  f(x) \approx g^{(t)}(x) = \sum_n \lambda_n^{-1} c_n(t) (2n+1)^{\frac{1}{2}} P_n\left( \frac{2(x-t)}{\theta} + 1 \right).
\end{align*}

\subsection{Derivation for Translated Laguerre (HiPPO-LagT)}
\label{sec:derivation-lagt}

We consider measures based on the generalized \emph{Laguerre} polynomials.
For a fixed $\alpha \in \mathbb{R}$, these polynomials $L^{(\alpha)}(t-x)$ are
orthogonal with respect to the measure $x^{\alpha} e^{-x}$ on $[0, \infty)$
(cf.\ \cref{sec:laguerre-properties}).
This derivation will involve tilting the measure governed by another parameter $\beta$.

The result in \cref{thm:legt-lagt} for HiPPO-LagT is for the case $\alpha = 0, \beta = 1$,
corresponding to the basic Laguerre polynomials and no tilting.

\paragraph{Measure and Basis}
We flip and translate the generalized Laguerre weight function and polynomials from $[0, \infty)$ to $(-\infty, t]$.
The normalization is found using equation~\eqref{eq:laguerre}.
\begin{align*}
  \omega(t, x) &=
  \begin{cases}
    (t-x)^\alpha e^{x-t} & \mbox{if } x \le t \\
    0 & \mbox{if } x > t
  \end{cases}
  \quad
  \\
  &= (t-x)^\alpha e^{-(t-x)} \mathbb{I}_{(-\infty, t]}
  \\
  p_n(t, x)
  &= \frac{\Gamma(n+1)^{\frac{1}{2}}}{\Gamma(n+\alpha+1)^{\frac{1}{2}}} L^{(\alpha)}_n(t-x)
\end{align*}

\paragraph{Tilted Measure}

We choose the following tilting $\chi$
\begin{align*}%
  \chi(t, x) &= (t-x)^\alpha \exp\left( -\frac{1-\beta}{2} (t-x) \right) \mathbb{I}_{\Rn{t}} \\
\end{align*}
for some fixed $\beta \in \mathbb{R}$.
The normalization is (constant across all $t$)
\begin{align*}%
  \zeta &= \int \frac{\omega}{\chi^2} = \int (t-x)^{-\alpha} e^{-\beta(t-x)} \mathbb{I}_{\Rn{t}} \dd x
  \\
  &= \Gamma(1-\alpha) \beta^{\alpha-1}
  ,
\end{align*}
so the tilted measure has density
\begin{align*}%
  \zeta(t)^{-1} \frac{\omega^{(t)}}{(\chi^{(t)})^2} &= \Gamma(1-\alpha)^{-1} \beta^{1-\alpha} (t-x)^{-\alpha} \exp\left( -\beta(t-x) \right) \mathbb{I}_{\Rn{t}}
  .
\end{align*}

We choose
\begin{align*}
  \lambda_n = \frac{\Gamma(n+\alpha+1)^{\frac{1}{2}}}{\Gamma(n+1)^{\frac{1}{2}}}
\end{align*}
to be the norm of the generalized Laguerre polynomial $L_n^{(\alpha)}$,
so that $\lambda_n p_n^{(t)} = L_n^{(\alpha)}(t-x)$,
and (following equation~\eqref{eq:framework-g}) the basis for $\nu^{(t)}$ is
\begin{equation}%
  \begin{aligned}
    g_n^{(t)} &= \lambda_n \zeta^{\frac{1}{2}} p_n^{(t)} \chi^{(t)}
    \\
    &= \zeta^{\frac{1}{2}} \chi^{(t)} L_n^{(\alpha)}(t-x)
  \end{aligned}
\end{equation}

\paragraph{Derivatives}

We first calculate the density ratio
\begin{align*}%
  \frac{\omega}{\chi}(t, x) &= \exp \left( -\frac{1+\beta}{2} (t-x) \right) \mathbb{I}_{\Rn{t}}
  .
\end{align*}
and its derivative
\begin{align*}
  \ppt \frac{\omega}{\chi}(t, x)
  &= -\left( \frac{1+\beta}{2} \right)\frac{\omega}{\chi}(t, x) + \exp\left( -\left( \frac{1+\beta}{2} \right)(t-x) \right) \delta_t
  .
\end{align*}

The derivative of Laguerre polynomials can be expressed as linear combinations of
other Laguerre polynomials (cf.\ \cref{sec:laguerre-properties}).
\begin{align*}
  \ppt \lambda_n p_n(t, x)
  &= \ppt L^{(\alpha)}_n(t-x)
  \\
  &= -L^{(\alpha)}_0(t-x) - \dots - L^{(\alpha)}_{n-1}(t-x)
  \\
  &= -\lambda_0 p_0(t, x) - \dots - \lambda_{n-1} p_{n-1}(t, x)
\end{align*}

\paragraph{Coefficient Dynamics}
Plugging these derivatives into equation~\eqref{eq:coefficient-dynamics} (obtained from differentiating the coefficient equation~\eqref{eq:coefficient}), where we suppress the dependence on $x$ for convenience:
\begin{align*}
  \ddt c_n(t)
  &= \zeta^{-\frac{1}{2}} \int f \cdot \left(\ppt \lambda_n p_n^{(t)}\right) \frac{\omega^{(t)}}{\chi^{(t)}}
  \\&\quad + \int f \cdot \left( \zeta^{-\frac{1}{2}} \lambda_n p_n^{(t)} \right) \left( \ppt \frac{\omega^{(t)}}{\chi^{(t)}} \right)
  \\
  &= -\sum_{k=0}^{n-1} \int f \cdot \left( \zeta^{-\frac{1}{2}} \lambda_k p_k^{(t)} \chi^{(t) }\right) \frac{\omega^{(t)}}{(\chi^{(t)})^2}
  \\&\quad - \left( \frac{1+\beta}{2} \right) \int f \cdot \left( \zeta^{-\frac{1}{2}} \lambda_n p_n^{(t)} \right) \frac{\omega^{(t)}}{\chi^{(t)}} + f(t) \cdot \zeta^{-\frac{1}{2}} L_n^{(\alpha)}(0)
  \\
  &= -\sum_{k=0}^{n-1} c_k(t) - \left( \frac{1+\beta}{2} \right) c_n(t) + \Gamma(1-\alpha)^{-\frac{1}{2}} \beta^{\frac{1-\alpha}{2}} \binom{n+\alpha}{n} f(t).
\end{align*}

We then get
\begin{equation}
  \begin{aligned}
    \ddt c(t)
    &= -A c(t) + B f(t)
    \\
    A &=
    \begin{bmatrix}
      \frac{1+\beta}{2} & 0 & \dots & 0 \\
      1 & \frac{1+\beta}{2} & \dots & 0 \\
      \vdots & & \ddots & \\
      1 & 1 & \dots & \frac{1+\beta}{2}
    \end{bmatrix}
    \\
    B &= \zeta^{-\frac{1}{2}} \cdot \begin{bmatrix} \binom{\alpha}{0} \\ \vdots \\ \binom{N-1+\alpha}{N-1} \end{bmatrix}
  \end{aligned}
\end{equation}

\paragraph{Reconstruction}
By equation~\eqref{eq:reconstruction}, at every time $t$, for $x \leq t$,
\begin{align*}
  f(x) \approx g^{(t)}(x)
  &= \sum_{n=0}^{N-1} \lambda_n^{-1} \zeta^{\frac{1}{2}} c_n(t) p_n^{(t)} \chi^{(t)}
  \\
  &= \sum_n \frac{n!}{(n+\alpha)!} \zeta^{\frac{1}{2}} c_n(t) \cdot L^{(\alpha)}_n(t-x) \cdot (t-x)^\alpha e^{\left( \frac{\beta-1}{2} \right)(t-x)}.
\end{align*}

\paragraph{Normalized Dynamics}
Finally, following equations~\eqref{eq:coefficient-dynamics-normalized} and \eqref{eq:reconstruction-normalized} to convert these to dynamics on the orthonormal basis of the normalized (probability) measure $\nu^{(t)}$ leads to the following $\hippo$ operator
\begin{equation}
  \begin{aligned}
    \ddt c(t)
    &= -A c(t) + B f(t)
    \\
    A &=
    -\Lambda^{-1}
    \begin{bmatrix}
      \frac{1+\beta}{2} & 0 & \dots & 0 \\
      1 & \frac{1+\beta}{2} & \dots & 0 \\
      \vdots & & \ddots & \\
      1 & 1 & \dots & \frac{1+\beta}{2}
    \end{bmatrix}
    \Lambda
    \\
    B &= \Gamma(1-\alpha)^{-\frac{1}{2}} \beta^{\frac{1-\alpha}{2}} \cdot \Lambda^{-1}
    \begin{bmatrix} \binom{\alpha}{0} \\ \vdots \\ \binom{N-1+\alpha}{N-1} \end{bmatrix}
    \\
    \Lambda &= \diag_{n \in [N]} \left\{ \frac{\Gamma(n+\alpha+1)^{\frac{1}{2}}}{\Gamma(n+1)^{\frac{1}{2}}} \right\}
  \end{aligned}
\end{equation}
and correspondingly a $\proj_t$ operator:
\begin{equation}
  f(x) \approx g^{(t)}(x)
  = \Gamma(1-\alpha)^{\frac{1}{2}} \beta^{-\frac{1-\alpha}{2}}
  \sum_n c_n(t) \cdot \frac{\Gamma(n+1)^{\frac{1}{2}}}{\Gamma(n+\alpha+1)^{\frac{1}{2}}} \cdot L^{(\alpha)}_n(t-x) \cdot (t-x)^\alpha e^{\left( \frac{\beta-1}{2} \right)(t-x)}.
\end{equation}

\subsection{Derivation for Scaled Legendre (HiPPO-LegS)}
\label{sec:derivation-legs}

As discussed in \cref{sec:theory_legs},
the scaled Legendre is our only method that uses a measure with varying width.

\paragraph{Measure and Basis}
We instantiate the framework in the case
\begin{align}
  \omega(t, x) &= \frac{1}{t} \mathbb{I}_{[0, t]}
  \label{eq:scale-legendre-weight} \\
  g_n(t, x) &= p_n(t, x) = (2n+1)^{\frac{1}{2}} P_n\left( \frac{2x}{t} - 1 \right)
  \label{eq:scale-legendre-p}
\end{align}

Here, $P_n$ are the basic Legendre polynomials (\cref{sec:legendre-properties}).
We use no tilting, i.e.\ $\chi(t, x) = 1$, $\zeta(t) = 1$, and $\lambda_n=1$ so that
the functions $g_n(t, x)$ are an orthonormal basis.

\paragraph{Derivatives}

We first differentiate the measure and basis:
\begin{align*}
  \ppt \omega(t, \cdot)   &= -t^{-2}\mathbb{I}_{[0,t]} + t^{-1}\delta_t = t^{-1} (-\omega(t) + \delta_t)
  \nonumber \\
  \ppt g_n(t, x)
  &= -(2n+1)^{\frac{1}{2}} 2xt^{-2} P_n'\left( \frac{2x}{t} - 1 \right)
  \nonumber \\
  &= -(2n+1)^{\frac{1}{2}} t^{-1} \left(\frac{2x}{t} - 1 + 1\right) P_n'\left( \frac{2x}{t} - 1 \right)
  .
\end{align*}
Now define $z = \frac{2x}{t} - 1$ for shorthand and apply the properties of derivatives of Legendre polynomials (equation~\eqref{eq:legendre-xd}).
\begin{align*}
  \ppt g_n(t, x)
  &= -(2n+1)^{\frac{1}{2}} t^{-1} (z+1) P_n'\left( z \right)
  \\
  &= - (2n+1)^{\frac{1}{2}} t^{-1} \left[n P_n\left( z \right) + (2n-1)P_{n-1}(z) + (2n-3)P_{n-2}(z) + \dots \right]
  \nonumber \\
  &= -t^{-1} (2n+1)^{\frac{1}{2}} \left[ n (2n+1)^{-\frac{1}{2}} g_n(t, x) + (2n-1)^{\frac{1}{2}} g_{n-1}(t, x) + (2n-3)^{\frac{1}{2}}g_{n-2}(t, x) + \dots \right]
  \nonumber
\end{align*}

\paragraph{Coefficient Dynamics}

Plugging these into~\eqref{eq:coefficient-dynamics}, we obtain
\begin{align*}
  \frac{d}{dt} c_n(t)
  &= \int f(x) \left( \ppt g_n(t, x) \right) \omega(t, x) \dd x
  + \int f(x) g_n(t, x) \left( \ppt \omega(t, x) \right) \dd x
  \\
  &= -t^{-1} (2n+1)^{\frac{1}{2}}\left[ n(2n+1)^{-\frac{1}{2}}c_n(t) + (2n-1)^{\frac{1}{2}}c_{n-1}(t) + (2n-3)^{\frac{1}{2}}c_{n-2}(t) + \dots \right]
  \\
  &\qquad - t^{-1} c_n(t) + t^{-1} f(t)g_n(t,t)
  \\
  &= -t^{-1} (2n+1)^{\frac{1}{2}} \left[ (n+1)(2n+1)^{-\frac{1}{2}}c_n(t) + (2n-1)^{\frac{1}{2}} c_{n-1}(t) + (2n-3)^{\frac{1}{2}}c_{n-2}(t) + \dots \right]
  \\
  &\qquad + t^{-1} (2n+1)^{\frac{1}{2}} f(t) 
\end{align*}
where we have used $g_n(t, t) = (2n+1)^{\frac{1}{2}} P_n(1) = (2n+1)^{\frac{1}{2}}$.
Vectorizing this yields equation~\eqref{eq:scaled-legendre-dynamics}:
\begin{align}
  \ddt c(t) &= -\frac{1}{t} A c(t) + \frac{1}{t} B f(t)
  \label{eq:scaled-legendre-dynamics-details}
  \\
  A_{nk}
  &=
  \begin{cases}
    (2n+1)^{1/2}(2k+1)^{1/2} & \mbox{if } n > k \\
    n+1 & \mbox{if } n = k \\
    0 & \mbox{if } n < k
  \end{cases},
  \nonumber \\
  B_n &= (2n+1)^{\frac{1}{2}} \nonumber
\end{align}

Alternatively, we can write this as
\begin{equation}
  \ddt c(t) = -t^{-1} D
  \left[
    M D^{-1}
    c(t)
    +
    \mathbf{1}
    f(t)
  \right]
  ,
\end{equation}
where
$D := \diag \left[ (2n+1)^{\frac{1}{2}} \right]_{n=0}^{N-1}$,
$\mathbf{1}$ is the all ones vector, and the state matrix $M$ is
\begin{align*}%
  M
  =
  \begin{bmatrix}
    1 & 0 & 0 & 0 & \dots & 0 \\
    1 & 2 & 0 & 0 & \dots & 0 \\
    1 & 3 & 3 & 0 & \dots & 0 \\
    1 & 3 & 5 & 4 & \dots & 0 \\
    \vdots & \vdots & \vdots & \vdots & \ddots & \vdots \\
    1 & 3 & 5 & 7 & \dots & N \\
  \end{bmatrix},
  \quad \text{that is, } \quad
  M_{nk} =
  \begin{cases}
    2k+1 & \mbox{if } k < n \\
    k+1 & \mbox{if } k = n \\
    0 & \mbox{if } k > n
  \end{cases}
\end{align*}

Equation~\eqref{eq:scaled-legendre-dynamics-details} is a linear dynamical system, except dilated by a time-varying factor $t^{-1}$, which arises from the scaled measure.

\paragraph{Reconstruction}

By equation~\eqref{eq:reconstruction}, at every time $t$ we have
\begin{align*}
  f(x) \approx g^{(t)}(x)
  &= \sum_n c_n(t) g_n(t, x).
  \\
  &= \sum_n c_n(t) (2n+1)^{\frac{1}{2}} P_n\left( \frac{2x}{t} - 1 \right).
\end{align*}

\subsection{Derivation for Fourier Bases}
\label{sec:derivation-fourier}

In the remainder of \cref{sec:derivations}, we consider some additional bases which are analyzable under the HiPPO framework.
These use measures and bases related to various forms of the Fourier transform.

\subsubsection{Translated Fourier}

Similar to the LMU, the sliding Fourier measure also has a fixed window length
$\theta$ parameter and slides it across time.

\paragraph{Measure}
The Fourier basis $e^{2\pi i n x}$ (for $n = 0, \dots, N-1$) can be seen as an
orthogonal polynomials basis $z^n$ with respect to the uniform measure on the
unit circle $\{ z \colon \abs{z} = 1 \}$.
By a change of variable $z \to e^{2\pi i x}$ (and thus changing the domain from the unit
circle to $[0, 1]$), we obtain the usual Fourier basis $e^{2\pi i n x}$.
The complex inner product $\langle f, g \rangle$ is defined as
$\int_{0}^{1} f(x) \overline{g(x)} \d x$.
Note that the basis $e^{2\pi i n x}$ is orthonormal.

For each $t$, we will use a sliding measure uniform on $[t - \theta, t]$ and rescale
the basis as $e^{2\pi i n \frac{t - x}{\theta}}$ (so they are still orthonormal, i.e.,
have norm 1):
\begin{align*}
  \omega(t, x) &= \frac{1}{\theta} \mathbb{I}_{[t-\theta, t]}
  \\
  p_n(t, x) &= e^{2\pi in \frac{t-x}{\theta}}.
\end{align*}
We sue no tilting (i.e., $\chi(t, x) = 1$).

\paragraph{Derivatives}
\begin{align*}
  \ppt \omega(t, x) &= \frac{1}{\theta} \delta_t - \frac{1}{\theta} \delta_{t-\theta}
  \\
  \ppt p_n(t, x) &= \frac{2\pi in}{\theta} e^{2\pi in \frac{t-x}{\theta}} = \frac{2 \pi in}{\theta} p_n(t, x).
\end{align*}

\paragraph{Coefficient Updates}
Plugging into equation~\eqref{eq:coefficient-dynamics} yields
\begin{align*}
  \ddt c_n(t)
  &= \frac{2\pi in}{\theta} c_n(t) + \frac{1}{\theta} f(t) p_n(t, t) - \frac{1}{\theta} f(t-\theta) p_n(t, t-\theta) \\
  &= \frac{2\pi in}{\theta} c_n(t) + \frac{1}{\theta} f(t) - \frac{1}{\theta} f(t-\theta).
\end{align*}
Note that $p_n(t, t) = p_n(t, t-\theta) = 1$.
Additionally, we no longer have access to $f(t-\theta)$ at time $t$, but this is
implicitly represented in our compressed representation of the function:
$f = \sum_{k=0}^{N-1} c_k(t) p_k(t)$.
Thus we approximate $f(t-\theta)$ by $\sum_{k=0}^{N-1} c_k(t) p_k(t, t-\theta) = \sum_{k=0}^{N-1} c_k(t)$.
Finally, this yields
\begin{align*}
  \ddt c_n(t) &= \frac{2\pi in}{\theta} c_n(t) + \frac{1}{\theta} f(t) - \frac{1}{\theta} \sum_{k=0}^{N-1} c_k(t).
\end{align*}
Hence $\ddt c(t) = A c(t) + B f(t)$
where
\begin{align*}
  A_{nk} =
  \begin{cases}
    -1/\theta & \mbox{if } k \neq n \\
    (2\pi in-1)/\theta & \mbox{if } k = n
  \end{cases},
  \qquad
  B_n = \frac{1}{\theta}.
\end{align*}

\paragraph{Reconstruction}

At every time step $t$, we have
\begin{align*}
  f(x) \approx \sum_n c_n(t) p_n(t, x) = \sum_{n} c_n(t) e^{2\pi i \frac{t - x}{\theta}}.
\end{align*}

\subsubsection{Fourier Recurrent Unit}

Using the HiPPO framework, we can also derive the Fourier Recurrent Unit (FRU)~\citep{zhang2018learning}.
\paragraph{Measure}
For each $t$, we will use a sliding measure uniform on $[t - \theta, t]$ and the
basis $e^{2\pi i n \frac{x}{\theta}}$:
\begin{align*}
  \omega(t, x) &= \frac{1}{\theta} \mathbb{I}_{[t-\theta, t]}
  \\
  p_n(t, x) &= e^{2\pi in \frac{x}{\theta}}.
\end{align*}
In general the basis is not orthogonal with respect to the measure $\omega(t, x)$, but
orthogonality holds at the end where $t = \theta$.

\paragraph{Derivatives}
\begin{align*}
  \ppt \omega(t, x) &= \frac{1}{\theta} \delta_t - \frac{1}{\theta} \delta_{t-\theta}
  \\
  \ppt p_n(t, x) &= 0.
\end{align*}

\paragraph{Coefficient Updates}
Plugging into equation~\ref{eq:coefficient-dynamics} yields
\begin{align*}
  \ddt c_n(t)
  &= \frac{1}{\theta} f(t) p_n(t, t) - \frac{1}{\theta} f(t-\theta) p_n(t, t-\theta) \\
  &= \frac{1}{\theta} e^{2\pi i n \frac{t}{\theta}} f(t) - \frac{1}{\theta}  e^{2\pi i n \frac{t}{\theta}} f(t-\theta).
\end{align*}
We no longer have access to $f(t-\theta)$ at time $t$, but we can approximate by
ignoring this term (which can be justified by assuming that the function $f$ is
only defined on $[0, \theta]$ and thus $f(x)$ can be set to zero for $x < 0$).
Finally, this yields
\begin{align*}
  \ddt c_n(t) &= \frac{e^{2\pi i n \frac{t}{\theta}}}{\theta} f(t).
\end{align*}
Applying forward Euler discretization (with step size = 1), we obtain:
\begin{equation*}
  c_n(k+1) = c_{n}(k) + \frac{e^{2\pi i n \frac{t}{\theta}}}{\theta} f(t).
\end{equation*}
Taking the real parts yields the Fourier Recurrent Unit updates~\citep{zhang2018learning}.

Note that the recurrence is independent in each $n$, so we don't have the pick
$n = 0, 1, \dots, N - 1$.
We can thus pick random frequencies $n$ as done in~\citet{zhang2018learning}.

\subsection{Derivation for Translated Chebyshev}
\label{sec:derivation-chebyshev}

The final family of orthogonal polynomials we analyze under the HiPPO framework are the Chebyshev polynomials.
The Chebyshev polynomials can be seen as the purely real analog of the Fourier basis;
for example, a Chebyshev series is related to a Fourier cosine series through a change of basis~\citep{boyd2001chebyshev}.

\paragraph{Measure and Basis}

The basic Chebyshev measure is $\omega^{\mathrm{cheb}} = (1-x^2)^{-1/2}$ on $(-1, 1)$.
Following \cref{sec:chebyshev-properties}, we choose the following measure and orthonormal basis polynomials in terms of the Chebyshev polynomials of the first kind $T_n$.
\begin{align*}%
  \omega(t, x) &= \frac{2}{\theta\pi} \omega^{cheb}\left( \frac{2(x-t)}{\theta} + 1 \right) \mathbb{I}_{(t-\theta, t)}
  \\
  &= \frac{1}{\theta\pi} \left( \frac{x-t}{\theta} + 1 \right)^{-1/2} \left( -\frac{x-t}{\theta} \right)^{-1/2} \mathbb{I}_{(t-\theta, t)}
  \\
  p_n(t, x) &= \sqrt{2} T_n\left( \frac{2(x-t)}{\theta} + 1 \right) \qquad \text{for } n \geq 1,
  \\
  p_0(t, x) &= T_0\left( \frac{2(x-t)}{\theta} + 1 \right).
\end{align*}

Note that at the endpoints, these evaluate to
\begin{align*}%
  p_n(t, t) &=
  \begin{cases}
    \sqrt{2} T_n(1) = \sqrt{2} & n \ge 1
    \\
    T_n(1) = 1 & n = 0
  \end{cases}
  \\
  p_n(t, t-\theta) &=
  \begin{cases}%
    \sqrt{2} T_n(-1) = \sqrt{2} (-1)^{n} & n \ge 1
    \\
    T_n(-1) = 1 & n = 0
  \end{cases}
\end{align*}

\paragraph{Tilted Measure}

Now we choose
\begin{align*}
  \chi^{(t)} = 8^{-1/2} \theta \pi \omega^{(t)},
\end{align*}
So
\begin{align*}
  \frac{\omega}{\chi^2} = \frac{1}{\frac{\theta^2\pi^2}{8}\omega} = \frac{8}{\theta \pi} \left( \frac{x-t}{\theta} + 1 \right)^{1/2} \left( -\frac{x-t}{\theta} \right)^{1/2} \mathbb{I}_{(t-\theta, t)}
\end{align*}
which integrates to $1$.

We also choose $\lambda_n = 1$ for the canonical orthonormal basis, so
\begin{align*}
  g^{(t)} = p_n^{(t)} \chi^{(t)}
\end{align*}

\paragraph{Derivatives}

The derivative of the density is
\begin{align*}
  \ppt \frac{\omega}{\chi} = \ppt \frac{8^{1/2}}{\theta\pi}\mathbb{I}_{(t-\theta, t)} = \frac{8^{1/2}}{\theta\pi}(\delta_t - \delta_{t-\theta}).
\end{align*}

We consider differentiating the polynomials separately for $n = 0$, $n$ even, and $n$ odd,
using equation~\eqref{eq:chebyshev-derivative}.
Defined $z = \frac{2(x-t)}{\theta} + 1$ for convenience.
First, for $n$ even,
\begin{align*}%
  \ppt p_n(t, x) &= -\frac{2^{\frac{3}{2}}}{\theta} T_n'\left( \frac{2(x-t)}{\theta} + 1 \right)
  \\
  &= -\frac{2^{\frac{3}{2}}}{\theta} T_n'\left( z \right)
  \\
  &= -\frac{2^{\frac{3}{2}}}{\theta} \cdot 2n \left( T_{n-1}(z) + T_{n-3}(z) + \dots + T_1(z) \right)
  \\&
  = -\frac{4n}{\theta} \left( p_{n-1}(t, x) + p_{n-3}(t, x) + \dots + p_1(t, x) \right)
\end{align*}
For $n$ odd,
\begin{align*}%
  \ppt p_n(t, x) &= -\frac{2^{\frac{3}{2}}}{\theta} T_n'\left( \frac{2(x-t)}{\theta} + 1 \right)
  \\
  &= -\frac{2^{\frac{3}{2}}}{\theta} T_n'\left( z \right)
  \\
  &= -\frac{2^{\frac{3}{2}}}{\theta} \cdot 2n \left( T_{n-1}(z) + T_{n-3}(z) + \dots + T_1(z) + \frac{1}{2}T_0(z) \right)
  \\&
  = -\frac{4n}{\theta} \left( p_{n-1}(t, x) + p_{n-3}(t, x) + \dots + 2^{-\frac{1}{2}}p_0(t, x) \right)
\end{align*}
And
\begin{align*}
  \ppt p_0(t, x) &= 0.
\end{align*}

\paragraph{Coefficient Dynamics}

\begin{align*}%
  c_n(t) &= \int f(x) p_n(t, x) \frac{2^{3/2}}{\theta\pi} \mathbb{I}_{(t-\theta, t)} \dd x
  \\
  \ddt c_n(t) &= \int f(x) \ppt p_n(t, x) \frac{2^{3/2}}{\theta\pi} \mathbb{I}_{(t-\theta, t)} \dd x + \frac{2^{3/2}}{\theta\pi} f(t)p_n(t, t) - \frac{2^{3/2}}{\theta\pi} f(t-\theta)p_n(t, t-\theta)
  \\&
  = -\frac{4n}{\theta} (c_{n-1} + c_{n-3} + \dots)  + \frac{2^{3/2}}{\theta\pi} f(t) \begin{cases} \sqrt{2} & n \geq 1 \\ 1 & n = 0 \end{cases}
  ,
\end{align*}
where we take $f(t - \theta) = 0$ as we no longer have access to it (this holds
when $t < \theta$ as well).

In the usual way, we can write this as linear dynamics
\begin{align*}
  \ddt c(t) &= -\frac{1}{\theta} A c(t) + \frac{1}{\theta} B f(t)
  \\
  A &=
  4
  \begin{bmatrix}
    0 & & & & \dots \\
    2^{-\frac{1}{2}} & 0 & & & \\
    0 & 2 & 0 & & \dots \\
    2^{-\frac{1}{2}} \cdot 3 & 0 & 3 & 0 & \\
    & \ddots & & \ddots &
  \end{bmatrix}
  \\
  B &=
  \frac{2^{3/2}}{\pi}
  \begin{bmatrix}
    1 \\ \sqrt{2} \\ \sqrt{2} \\ \sqrt{2} \\ \vdots
  \end{bmatrix}
\end{align*}

\paragraph{Reconstruction}
In the interval $(t-\theta, t)$,
\begin{align*}
  f(x) \approx \sum_{n=0}^{N-1} c_n(t) p_n(t, x) \chi(t, x).
\end{align*}

  }{

}

\section{HiPPO-LegS Theoretical Properties}
\label{sec:hippo-theory}

\subsection{Timescale equivariance}

\begin{proof}[Proof of \cref{prop:timescale}]%
    Let $\tilde{f}(t) = f(\alpha t)$.
    Let $c = \proj f$ and $\tilde{c} = \proj \tilde{f}$.
    By the HiPPO equation~\eqref{eq:coefficient} update and the basis
    instantiation for LegS (equation \eqref{eq:scale-legendre-p}),
    \begin{align*}%
        \tilde{c}_n(t) &= \langle \tilde{f}, g_n^{(t)} \rangle_{\mu^{(t)}} \\
        &= \int \tilde{f}(t) (2n+1)^{\frac{1}{2}} P_n\left( 2\frac{x}{t} - 1 \right) \frac{1}{t} \mathbb{I}_{[0, 1]}\left(\frac{x}{t}\right) \dd x
        \\
        &= \int f(\alpha t) (2n+1)^{\frac{1}{2}} P_n\left( 2\frac{x}{t} - 1 \right) \frac{1}{t} \mathbb{I}_{[0, 1]}\left(\frac{x}{t}\right) \dd x
        \\
        &= \int f(\alpha t) (2n+1)^{\frac{1}{2}} P_n\left( 2\frac{x}{\alpha t} - 1 \right) \frac{1}{\alpha t} \mathbb{I}_{[0, 1]}\left(\frac{x}{\alpha t}\right) \dd x
        \\
        &= c_n(\alpha t).
    \end{align*}
    The second-to-last equality uses the change of variables $x \mapsto \frac{x}{\alpha}$.
\end{proof}

\subsection{Speed}

In this section we work out the fast update rules according to the forward
Euler, backward Euler, bilinear, or generalized bilinear transform
discretizations (cf.\ \cref{sec:discretization-full}).
Recall that we must be able to perform matrix-vector multiplication by $I + \delta A$ and $(I - \delta A)^{-1}$
where $\delta$ is some multiple of the step size $\Delta t$ (equation~\eqref{eq:gbt}).

It is easily seen that the LegS update rule involves a matrix $A$ of the following form (\cref{thm:legs}):
$A = D_1 (L + D_0) D_2$, where $L$ is the all $1$ lower triangular matrix and $D_0, D_1, D_2$ are diagonal.
Clearly, $I + \delta A$ is efficient (only requiring $O(N)$ operations),
as it only involves matrix-vector multiplication by diagonals $D_0, D_1, D_2$, or multiplication by $L$ which is the $\mathsf{cumsum}$ operation.

Now we consider multiplication by the inverse $(I+\delta A)^{-1}$ (the minus sign can be absorbed into $\delta$).
Write
\begin{align*}
  (I + \delta D_1 (L+D_0) D_2)^{-1}
  &= \left( D_1 (D_1^{-1} D_2^{-1} + \delta (L+D_0)) D_2 \right)^{-1} \\
  &= \delta^{-1} D_2^{-1} \left( \delta^{-1} D_1^{-1} D_2^{-1} + D_0 + L \right)^{-1} D_1^{-1}
\end{align*}
Since diagonal multiplication is efficient,
the crucial operation is inversion multiplication by a matrix of the form $L + D$.

Consider solving the equation $(L + D) x = y$.
This implies $x_0 + \dots + x_{k-1} = y_k - (1+d_k) x_k$.
The solution is
\begin{align*}
  x_0 &= \frac{y_0}{1+d_0} \\
  x_k &= \frac{y_k - x_0 - \dots - x_{k-1}}{1+d_k} \\
\end{align*}
Define $s_k = x_0 + \dots + x_k$.
Then
\begin{align*}
  s_k = s_{k-1} + x_k = s_{k-1} + \frac{y_k - s_{k-1}}{1 + d_k} = \frac{y_k + d_k s_{k-1}}{1 + d_k} = \frac{d_k}{1+d_k}s_{k-1} + \frac{y_k}{1+d_k}.
\end{align*}

Finally, consider how to calculate a recurrence of the following form efficiently.
\begin{align*}
  x_0 = \beta_0, x_k = \alpha_k x_{k-1} + \beta_{k}.
\end{align*}
This update rule can also be written
\begin{align*}
  \frac{x_k}{\alpha_k\dots\alpha_1} = \frac{x_{k-1}}{\alpha_{k-1}\dots\alpha_1} + \frac{\beta_k}{\alpha_k\dots\alpha_1}.
\end{align*}
Evidently $x$ can be computed in a vectorized way as
\begin{align*}
  x = \mathsf{cumsum}(\beta / \mathsf{cumprod}(\alpha)) \cdot \mathsf{cumprod}(\alpha)
  .
\end{align*}
This is an $O(N)$ computation.

\subsection{Gradient Norms}

We analyze the discrete time case under the Euler discretization (\cref{sec:discretization-full}),
where the HiPPO-LegS recurrent update is equation~\eqref{eq:legs-discrete}, restated here for convenience:
\begin{align*}
  c_{k+1} = \left( 1 - \frac{A}{k} \right) c_k + \frac{1}{k} B f_k.
\end{align*}
These gradient asymptotics hold under other discretizations.

We will show that
\begin{proposition}%
  For any times $k < \ell$, the gradient norm of the HiPPO-LegS operator
  for the output at time $\ell+1$ with respect to input at time $k$ is
  $\left\| \frac{\partial c_{\ell+1}}{\partial f_{k}} \right\| = \Theta\left( 1 / \ell \right)$.
\end{proposition}
\begin{proof}%
  We take $N$ to be a constant.

  Without loss of generality assume $k > 2$, as the gradient change for a single initial step is bounded.
  By unrolling the recurrence~\eqref{eq:legs-discrete}, the dependence of $c_{\ell+1}$ on $c_k$ and $f_k, \dots, f_{\ell}$ can be made explicit:
  \begin{align*}
    c_{\ell+1} &= \left( I - \frac{A}{\ell} \right) \dots \left( I - \frac{A}{k} \right) c_k
    \\
    &\quad + \left( I - \frac{A}{\ell} \right) \dots \left( I - \frac{A}{k+1} \right) \frac{B}{k} f_k
    \\
    &\quad + \left( I - \frac{A}{\ell} \right) \dots \left( I - \frac{A}{k+2} \right) \frac{B}{k+1} f_{k+1}
    \\
    &\quad \quad \vdots
    \\
    &\quad + \left( I - \frac{A}{\ell} \right) \frac{B}{\ell-1} f_{\ell-1}
    \\
    &\quad + \frac{B}{\ell} f_{\ell}
    .
  \end{align*}
  Therefore
  \begin{align*}
    \frac{\partial c_{\ell+1}}{\partial f_k} = \left( I - \frac{A}{\ell} \right) \dots \left( I - \frac{A}{k+1} \right) \frac{B}{k}.
  \end{align*}
  Notice that $A$ has distinct eigenvalues $1, 2, \dots, N$, since those are the elements of its diagonal and $A$ is triangular (\cref{thm:legs}).
  Thus the matrices $I - \frac{A}{\ell}, \dots, I - \frac{A}{k+1}$ are diagonalizable with a common change of basis.
  The gradient then has the form $P D P^{-1} B$ for some invertible matrix
  $P$ and some diagonal matrix $D$.
  Its norm is therefore bounded from below (up to constant) by the smallest
  singular value of $P$ and $\|P^{-1} B\|$, both of which are nonzero constants,
  and the largest diagonal entry of $D$.
  It thus suffices to bound this largest diagonal entry of $D$, which is the
  largest eigenvalue of this product,
  \begin{align*}
    \rho = \left(1 - \frac{1}{\ell}\right) \dots \left(1 - \frac{1}{k+1}\right) \frac{1}{k}.
  \end{align*}
  The problem reduces to showing that $\rho = \Theta(1/l)$.

  We will use the following facts about the function $\log \left( 1 - \frac{1}{x} \right)$.
  First, it is an increasing function, so
  \begin{align*}
    \log \left( 1 - \frac{1}{x} \right) \ge \int_{x-1}^x \log \left( 1 - \frac{1}{\lambda} \right) \dd \lambda.
  \end{align*}
  Second, its antiderivative is
  \begin{align*}
    \int \log \left( 1 - \frac{1}{x} \right) = \int \log(x-1) - \log(x) = (x-1)\log(x-1) - x\log(x) = x\log\left( 1-\frac{1}{x} \right) - \log(x-1)
    .
  \end{align*}
  
  Therefore, we have
  \begin{align*}
    \log \left(1 - \frac{1}{\ell}\right) \dots \left(1 - \frac{1}{k+1}\right)
    &= \sum_{i=k+1}^\ell \log (1 - \frac{1}{i})
    \\
    &\ge \sum_{i=k+1}^\ell \int_{i-1}^i \log \left(1 - \frac{1}{x}\right) \dd x
    \\
    &= \int_{k}^\ell \log \left(1 - \frac{1}{x}\right) \dd x
    \\
    &= \left[ (x-1)\log(x-1) - x\log(x) \right] \big|_{k}^{\ell}
    \\
    &= \ell \log \left( 1-\frac{1}{\ell} \right) - \log(\ell-1)
    \\& - \left( k \log \left( 1-\frac{1}{k} \right) - \log(k-1) \right)
    .
  \end{align*}
  Finally, note that $x \log \left( 1 - \frac{1}{x} \right)$ is an increasing function, and bounded from above since it is negative, so it is $\Theta(1)$ (this can also be seen from its Taylor expansion).
  Thus we have
  \begin{align*}
    \log \rho \ge \Theta(1) - \log(\ell-1) + \log(k-1) - \log(k),
  \end{align*}
  Furthermore, all inequalities are asymptotically tight, so that $\rho = \Theta(1 / \ell)$ as desired.
  
\end{proof}

\subsection{Function Approximation Error}

\begin{proof}[Proof of \cref{prop:approximation_error}]%
  Fix a time $t$.
  HiPPO-LegS uses the measure $\omega(t, x) = \frac{1}{t} \mathbb{I}_{[0, t]}$ and the
  polynomial basis $p_n(t, x) = (2n+1)^{\frac{1}{2}} P_n \left( \frac{2x}{t} - 1 \right)$.
  Let $c_n(t) = \langle f_{\leq t}, p_n^{(t)} \rangle_{\mu^{(t)}}$ for $n = 0, 1, \dots$.
  Then the projection $g^{(t)}$ is obtained by linear combinations of the basis
  functions, with $c_n(t)$ as coefficients:
  \begin{equation*}
    g^{(t)}= \sum_{n=0}^{N - 1} c_n(t) p_n^{(t)}.
  \end{equation*}
  Since $p_n^{(t)}$ forms an orthonormal basis of the Hilbert space defined by
  the inner product $\langle \cdot, \cdot \rangle_{\mu^{(t)}}$~\citep{chihara}, by Parseval's
  identity,
  \begin{equation*}
    \norm{f_{\leq t} - g^{(t)}}^2_{\mu^{(t)}} = \sum_{n=N}^{\infty} c_n^2(t).
  \end{equation*}

  To bound the error $\norm{f_{\leq t} - g^{(t)}}_{\mu^{(t)}}$, it suffices to
  bound the sum of the squares of the high-order coefficients $c_n(t)$ for
  $n = N, N+1, \dots$.
  We will bound each coefficient by integration by parts.

  We first simplify the expression for $c_n(t)$.
  For any $n \geq 1$, we have
  \begin{align*}
    c_n(t)
    &= \langle f_{\leq t}, p_n^{(t)} \rangle_{\mu^{(t)}} \\
    &= \frac{1}{t} (2n+1)^{\frac{1}{2}} \int_{0}^{t} f(x) P_n\left(\frac{2x}{t} - 1\right) \d x \\
    &= \frac{(2n+1)^{\frac{1}{2}}}{2} \int_{-1}^{1} f \left( \frac{1+x}{2} t \right) P_n(x) \d x  & \text{(change of variable $x \to \frac{1+x}{2} t$)}.
  \end{align*}

  As $P_n(x) = \frac{1}{2n+1} \frac{d}{dx} (P_{n+1}(x) - P_{n-1}(x))$ (cf.\
  \cref{sec:legendre-properties}), integration by parts yields:
  \begin{align*}
    c_n(t)
    &= \frac{(2n+1)^{\frac{1}{2}}}{2} \left. \left[ f \left( \frac{1+x}{2} t \right) \frac{1}{2n+1} (P_{n+1}(x) - P_{n-1}(x)) \right] \right|_{-1}^1 \\
    &\quad - \frac{(2n+1)^{\frac{1}{2}}}{2} \int_{-1}^{1} \frac{t}{2} f' \left( \frac{1+x}{2} t \right) \frac{1}{2n+1} (P_{n+1}(x) - P_{n-1}(x)) \d x.
  \end{align*}
  Notice that the boundary term is zero, since
  $P_{n+1}(1) = P_{n-1}(1) = 1$ and $P_{n+1}(-1) = P_{n-1}(-1) = \pm 1$ (either
  both 1 or both $-1$ depending on whether $n$ is odd or even).
  Hence:
  \begin{equation*}
    c_n(t) = -\frac{1}{4} \cdot \frac{1}{(2n+1)^{\frac{1}{2}}} \cdot t \int_{-1}^{1} f' \left( \frac{1+x}{2} t \right) (P_{n+1}(x) - P_{n-1}(x)) \d x.
  \end{equation*}

  Now suppose that $f$ is $L$-Lipschitz, which implies that $\abs{f'} \leq L$.
  Then
  \begin{align*}
    c_n^2(t)
    &\leq t^2 L^2 \frac{1}{16} \cdot \frac{1}{2n+1} \left[ \int_{-1}^{1} \abs{P_{n+1}(x) - P_{n-1}(x)} \d x \right]^2 \\
    &\leq t^2 L^2 \frac{1}{16} \cdot \frac{1}{2n+1} \cdot 2 \int_{-1}^{1} (P_{n+1}(x) - P_{n-1}(x))^2 \d x && \text{(Cauchy--Schwarz)} \\
    &= t^2 L^2 \frac{1}{8} \frac{1}{2n+1} \left[ \int_{-1}^{1} P_{n+1}^2(x) \d x + \int_{-1}^{1} P_{n-1}^2(x) \d x \right] && \text{($P_{n+1}$ and $P_{n-1}$ are orthogonal)} \\
    &= t^2 L^2 \frac{1}{8} \frac{1}{2n+1} \left[ \frac{2}{2n+3} + \frac{2}{2n-1} \right] \\
    &= O(1) t^2 L^2 \frac{1}{n^2}.
  \end{align*}
  Summing for all $n \geq N$ yields:
  \begin{align*}
    \norm{f_{\leq t} - g^{(t)}}^2_{\mu^{(t)}}
    &= \sum_{n=N}^{\infty} c_n^2(t) = O(1) t^2 L^2 \sum_{n=N}^{\infty} \frac{1}{n^2} = O(1) t^2 L^2 \frac{1}{N}.
  \end{align*}
  We then obtain that $\norm{f_{\leq t} - g^{(t)}}_{\mu^{(t)}} = O(tL/\sqrt{N})$ as claimed.

  Now supposed that $f$ has $k$ derivatives and the $k$-th derivative is bounded.
  The argument is similar to the one above where we integrate by parts $k$
  times.
  We sketch this argument here.

  Take $k$ to be a constant, and let $n \geq k$.
  Applying integration by parts $k$ times, noting that all the boundary
  terms are zero, gives:
  \begin{equation*}
    c_n(t) = O(1) (2n+1)^{\frac{1}{2}} t^k \int_{-1}^{1} f^{(k)} \left( \frac{1+x}{2} t \right) q_k(x) \d x,
  \end{equation*}
  where $q_k(x)$ is a polynomial such that $\frac{\d^k}{\d x^k} q_k(x) = P_n(x)$.
  Then since $f^{(k)}$ is bounded,
  $\abs{c_n(t)} = O(1) (2n+1)^{\frac{1}{2}} \int_{-1}^{1} \abs{q_k(x)} \d x$, and so
  \begin{equation*}
    c_n^2(t) = O(1) t^{2k} (2n+1) \left[ \int_{-1}^{1} \abs{q_k(x)} \d x \right]^2 = O(1) t^{2k} (2n+1) \int_{-1}^{1} q_k^2(x) \d x \quad \text{(Cauchy--Schwarz)}.
  \end{equation*}
  It remains to bound $\int_{-1}^{1} q_k^2(x) \d x$.
  Using the fact that
  $\frac{\d}{\d x} P_n(x) = \frac{1}{2n+1} (P_{n+1}(x) - P_{n-1}(x))$ repeatedly, we have:
  \begin{align*}
    q_1 &= \frac{1}{2n+1} (P_{n+1} - P_{n-1}) = \frac{1}{n+O(1)} \cdot \frac{1}{2} (P_{n+1} - P_{n-1}) \\
    q_2 &= \frac{1}{(n+O(1))^2} \frac{1}{2^2} (P_{n+2} - P_n - P_n + P_{n-2}) = \frac{1}{(n+O(1))^2} \frac{1}{2^2} (P_{n+2} - 2P_n + P_{n-2}) \\
    q_3 &= \frac{1}{(n+O(1))^3} \frac{1}{2^3} (P_{n+3} - P_{n+1} - 2P_{n+1} + 2P_{n-1} + P_{n-1} - P_{n-3}) = \frac{1}{(n+O(1))^3} \frac{1}{2^3} (P_{n+3} - 3P_{n+1} + 3P_{n-1} - P_{n-3}) \\
    \dots
  \end{align*}
  In general, when we expand out $\int_{-1}^{1} q_k^2(x) \d x$, since the $P_m$'s
  are orthogonal, we get $k+1$ terms of the form
  $\frac{1}{(n+O(1))^{2k}} \frac{1}{2^{2k}} \binom{k}{l}^2 \int_{-1}^{1} P_m^2(x) \d x$
  for $k$ different values of $m$ in the range $[n-k, n+k]$, and $l$ goes from 0 to $k$.
  For each $m$, $\int_{-1}^{1} P_m^2(x) \d x = \frac{1}{n+O(1)}$, and
  $\sum_{l=0}^{k} \binom{k}{l}^2 = \binom{2k}{k}$.
  Summing up all $k+1$ terms yields
  \begin{equation*}
    \int_{-1}^{1} q_k^2(x) \d x = \frac{1}{(n+O(1))^{2k+1}} \frac{1}{2^k} \binom{2k}{k}.
  \end{equation*}
  By Stirling's approximation, $\binom{2k}{k} = O(1) 4^k$, so
  $\int_{-1}^{1} q_k^2(x) \d x = \frac{O(1) 2^k}{(n+O(1))^{2k+1}}$.
  Noting that $k$ is a constant, plugging this into the bound for $c_n^2(t)$:
  \begin{equation*}
    c_n^2(t) = O(1) t^{2k} (2n+1) \frac{O(1)2^k}{(n+O(1))^{2k+1}} = O(1) t^{2k} \frac{1}{n^{2k}}.
  \end{equation*}
  Summing for all $n \geq N$ yields:
  \begin{align*}
    \norm{f_{\leq t} - g^{(t)}}^2_{\mu^{(t)}}
    &= \sum_{n=N}^{\infty} c_n^2(t) = O(1) t^{2k} \sum_{n=N}^{\infty} \frac{1}{n^{2k}} = O(1) t^{2k} \frac{1}{N^{2k-1}}.
  \end{align*}
  We then obtain that $\norm{f_{\leq t} - g^{(t)}}_{\mu^{(t)}} = O(t^k N^{-k+1/2})$ as claimed.
\end{proof}

\textbf{Remark.} The approximation error of Legendre polynomials
reduces to how fast the Legendre coefficients decay, subjected to the smoothness
assumption of the input function.
This result is analogous to the classical result in Fourier
analysis, where the $n$-th Fourier coefficients decay as $O(n^{-k})$ if the input
function has order-$k$ bounded derivatives~\citep{korner1989fourier}.
That result is also proved by integration by parts.

\section{Experiment Details and Additional Results}
\label{sec:experiment-details}

\subsection{Model Architecture Details}
\label{subsec:model_architectures}

Given inputs $x_t$ or features thereof $f(x_t)$ in any model, the HiPPO framework can be used to memorize the history of features $f_t$ through time.
As the discretized HiPPO dynamics form a linear recurrent update similar in style to RNNs (e.g., \cref{thm:legs}),
we focus on these models in our experiments.

Thus, given any RNN update function $h_{t} = \tau(h_{t-1}, x_t)$, we simply replace the previous hidden state with a projected version of its entire history.
\begin{figure}[!ht]
  \begin{minipage}{.5\linewidth}%
    \centering
    \includegraphics[width=\linewidth]{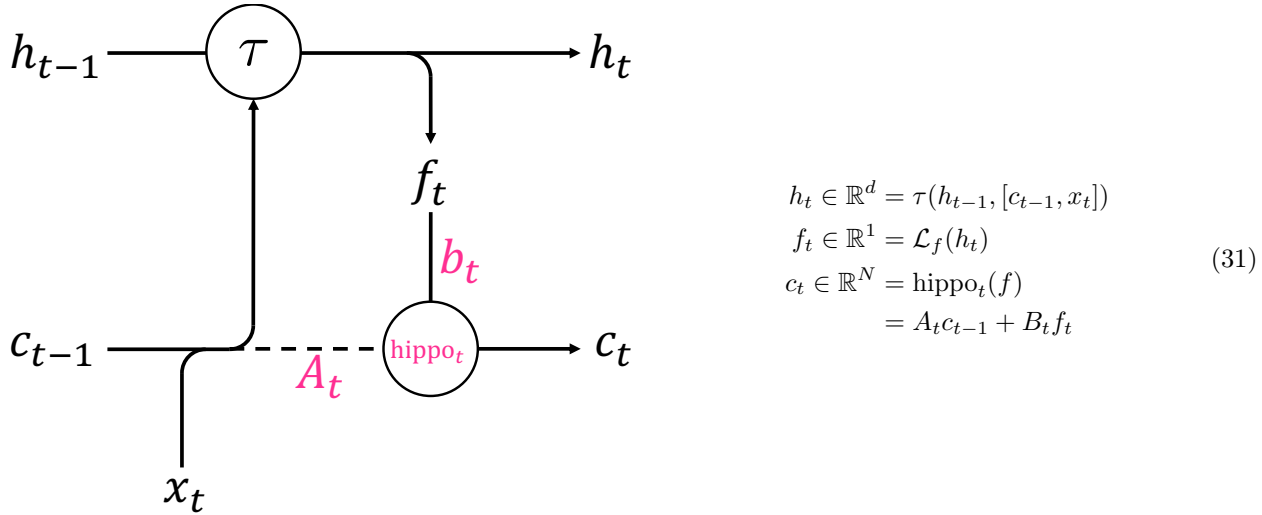}
  \end{minipage}
  \begin{minipage}{.5\linewidth}
    \begin{equation}%
      \label{eq:architecture}
      \begin{aligned}
        h_t \in \R^d &= \tau(h_{t-1}, [c_{t-1}, x_t]) \\
        f_t \in \R^1 &= \mathcal{L}_f(h_{t}) \\
        c_t \in \R^N &= \hippo_t(f) \\
        &= A_t c_{t-1} + B_t f_t
      \end{aligned}
    \end{equation}
  \end{minipage}
  \caption{The simple RNN model we use HiPPO with, and associated update equations. $\mathcal{L}_\square$ is a parametrized linear function, $\tau$ is any RNN update function, and $[\cdot]$ denotes concatenation.
    $\textcolor{magenta}{\hippo}$ is the HiPPO memory operator which orthogonalizes the history of the $f_t$ features up to time $t$. $A_t, B_t$ are fixed matrices depending on the chosen measure .
    $N$ and $d$ represent the approximation order and hidden state size, respectively.
  }
  \label{fig:cell-equation}
\end{figure}
Equations~\eqref{eq:architecture} lists the explicit update equations and Figure~\ref{fig:cell-equation} illustrates the model.
In our experiments, we choose a basic gated RNN update
\begin{align*}
  \tau(h, x) = (1-g(h, x)) \circ h + g(h, x) \circ \tanh(\mathcal{L}_\tau(h, x)), \qquad g(h, x) = \sigma(\mathcal{L}_g(h, x)).
\end{align*}

\paragraph{Methods and Baselines}

We consider the following instantiations of our framework HiPPO.

\textbf{HiPPO-}\textbf{LegT}, \textbf{LagT}, and \textbf{LegS}, use the translated Legendre, and tilted Laguerre, and scaled Legendre measure families with update dynamics \eqref{eq:translated-legendre-dynamics}, \eqref{eq:laguerre-dynamics}, and \eqref{eq:scaled-legendre-dynamics}.
As mentioned, LegT has an additional hyperparameter $\theta$, which should be set to the timescale of the data if known a priori.
We attempt to set it equal to its ideal value (the length of the sequences) in every task,
and also consider $\theta$ values that are too large and small
to illustrate the effect of this hyperparameter.

Our derivations in \cref{sec:derivation-legt,sec:derivation-lagt,sec:derivation-legs,sec:derivation-fourier,sec:derivation-chebyshev} show that there is a large variety of update equations that can arise from the HiPPO framework---for example, the tilted generalized Laguerre polynomials lead to an entire family governed by two free parameters (\cref{sec:derivation-lagt})---many of which lead to linear dynamics of the form $\ddt c(t) = -A c(t) + B f(t)$ for various $A, B$.
Given that many different update dynamics lead to such dynamical systems that give sensible results,
we additionally consider the \textbf{HiPPO-Rand} baseline that uses random $A$ and $B$ matrices (normalized appropriately) in its dynamics.

We additionally compare against the following standard RNN baselines.
The \textbf{RNN} is a vanilla RNN.
The \textbf{MGU} is a minimal gated architecture, equivalent to a \textbf{GRU} without the reset gate. The HiPPO architecture we use is simply the MGU with an additional $\hippo$ intermediate layer.
The \textbf{LSTM} is the most well-known and popular RNN architecture, which is a more sophisticated gated RNN.
The \textbf{expRNN}~\citep{lezcano2019cheap} is the state-of-the-art representative of the \emph{orthogonal RNN} family of models designed for long-term dependencies~\citep{arjovsky2016unitary}.
The \textbf{LMU} is the exact same model as in \citet{voelker2019legendre}; it is equivalent to HiPPO-LegT with a different RNN architecture.

All methods have the same hidden size in our experiments.
In particular, for simplicity and to reduce hyperparameters, HiPPO variants tie the memory size $N$ to the hidden state dimension $d$.
The hyperparameter $N$ and $d$ is also referred to as the number of hidden units.

\paragraph{Model and Architecture Comparisons}
The model~\eqref{eq:architecture} we use is a simple RNN that bears similarity to the classical LSTM and the original LMU cell.
In comparison to the LSTM, HiPPO can be seen as a variant where the memory $m_t$ plays the role of the LSTM's hidden state and $h_t$ plays the role of the LSTM's gated cell state, with equal dimensionalities.
HiPPO updates $m_t$ using the fixed $A$ transition matrix instead of a learned matrix, and also lacks ``input'' and ``output'' gates, so for a given hidden size, it requires about half the parameters.

The LMU is a version of the HiPPO-LegT cell with an additional hidden-to-hidden transition matrix and memory-to-memory transition vector instead of the gate $g$, leaving it with approximately the same number of trainable parameters.

\paragraph{Training Details}
Unless stated otherwise, all methods use the Adam optimizer~\citep{kingma2014adam} with learning rate frozen to $0.001$, which has been a robust default for RNN based models~\citep{voelker2019legendre, gu2020improving}.

All experiments use PyTorch 1.5 and are run on a Nvidia P100 GPU.

\subsection{Permuted MNIST}
\label{sec:pmnist-details}

\paragraph{Task}
The input to the sequential MNIST (sMNIST) task \cite{le2015simple} is an MNIST source image,
flattened in row-major order into a single sequence of length 784.
The goal of the model is to process the entire image sequentially before
outputting a classification label, requiring learning long-term dependencies.
A variant of this, the permuted MNIST (pMNIST) task, applies a fixed permutation
to every image, breaking locality and further straining a model's capacity for
long-term dependencies.

Models are trained using the cross-entropy loss.
We use the standard train-test split (60,000 examples for training and 10,000
for testing), and further split the training set with
10\% to be used as validation set.

\paragraph{Baselines and Ablations}

\cref{tab:pmnist} is duplicated here in \cref{tab:pmnist-val,tab:pmnist-test}, with more complete baselines and hyperparameter ablations.

\cref{tab:pmnist-val} consists of our implementations of various baselines related to our method, described in \cref{subsec:model_architectures}.
Each method was ran for 3 seeds, and the maximum average validation accuracy is reported.

All methods used the same hidden size of $512$; we found that this gave better performance than $256$, and further increasing it did not improve more.
All methods were trained for 50 epochs with a batch size of 100.

\paragraph{State of the Art}
\cref{tab:pmnist-test} directly shows the reported test accuracy of various methods on this data (Middle and Bottom).
\cref{tab:pmnist-test} (Top) reports the test accuracy of various instantations of our methods.
We additionally include our reproduction of the LMU,
which achieved better results than reported in~\citet{voelker2019legendre} (possibly due to a larger hidden size).
We note that \emph{all} of our HiPPO methods are competitive; each of them (HiPPO-LegT, HiPPO-LagT, HiPPO-LegS) achieves state-of-the-art among previous recurrent sequence models.
Note that differences between our HiPPO-LegT and LMU numbers in \cref{tab:pmnist-test} (Top) stem primarily from the architecture difference (\cref{subsec:model_architectures}).

\paragraph{Timescale Hyperparameters}
\cref{tab:pmnist-val} also shows ablations for the HiPPO-LegT and HiPPO-LagT timescale hyperparameters.
HiPPO-LagT sweeps the discretization step size $\Delta t$ (\cref{sec:discretization,sec:discretization-full}).
For LegT, we set $\Delta t = 1.0$ without loss of generality, as only the ratio of $\theta$ to $\Delta t$ matters.
These timescale hyperparameters are important for these methods.
Previous works have shown that the equivalent of $\Delta t$ in standard RNNs, i.e.\ the gates of LSTMs and GRUs (\cref{sec:discretization}), can also drastically affect their performance \cite{tallec2018can,gu2020improving}.
For example, the only difference between the URLSTM and LSTM in \cref{tab:pmnist-test} is a reparametrization of the gates.

\begin{table}[ht]%
  \caption{
    \textbf{Our methods and related baselines}. Permuted MNIST (pMNIST) validation scores. (Top): Our methods. (Bottom): Recurrent baselines.
  }
    \centering
    \begin{tabular}{@{}ll@{}}
      \toprule
      Method                     & Validation accuracy (\%) \\
      \midrule
      \textbf{HiPPO-LegS}        & \textbf{98.34}           \\
      HiPPO-LagT $\Delta t = 1.0$           & 98.15                    \\
      HiPPO-LegT $\theta = 200$  & 98.00                     \\
      HiPPO-LegT $\theta = 2000$ & 97.90                     \\
      HiPPO-LagT $\Delta t = 0.1$           & 96.44                    \\
      HiPPO-LegT $\theta = 20$   & 91.75                    \\
      HiPPO-LagT $\Delta t = 0.01$           & 90.71                    \\
      HiPPO-Rand                 & 69.93 \\
      \midrule
      LMU                        & 97.08                    \\
      ExpRNN                     & 94.67                    \\
      GRU                        & 93.04                    \\
      LSTM                       & 92.54                    \\
      MGU                        & 89.37                    \\
      RNN                        & 52.98                    \\
      \bottomrule
    \end{tabular}
    \label{tab:pmnist-val}
  \end{table}
\begin{table}[ht]
  \caption{ \textbf{Comparison to prior methods for pixel-by-pixel image classification.} Reported test accuracies from previous works on pixel-by-pixel image classification benchmarks. Top: Our methods. Middle: Recurrent baselines and variants. Bottom: Non-recurrent sequence models with global receptive field.
  }
    \centering
    \begin{tabular}{@{}lll@{}}
      \toprule
      Model                                     & Test accuracy (\%) \\
      \midrule
      \textbf{HiPPO-LegS}                       & \textbf{98.3}      \\
      HiPPO-Laguerre                            & 98.24              \\
      HiPPO-LegT                                & 98.03              \\
      LMU (ours)                                & 97.29              \\
      \midrule
      URLSTM + Zoneout \citep{krueger2016zoneout}          & 97.58              \\
      LMU~\citep{voelker2019legendre}           & 97.15              \\
      URLSTM~\citep{gu2020improving}            & 96.96              \\
      IndRNN~\citep{indrnn}                     & 96.0               \\
      Dilated RNN~\citep{chang2017dilated}      & 96.1               \\
      r-LSTM ~\citep{trinh2018learning}         & 95.2               \\
      LSTM~\citep{gu2020improving}              & 95.11              \\
      \midrule
      TrellisNet~\citep{trellisnet}             & 98.13              \\
      Temporal ConvNet~\citep{bai2018empirical} & 97.2               \\
      Transformer~\citep{trinh2018learning}     & 97.9               \\
      \bottomrule
    \end{tabular}
    \label{tab:pmnist-test}
\end{table}

\subsection{Copying}
\label{sec:copying-details}

\paragraph{Task}
In the Copying task~\citep{arjovsky2016unitary},
the input is a sequence of $L+20$ digits where the first 10 tokens $\left(a_0, a_1, \dots, a_9\right)$ are randomly chosen from $\left\{1,\dots,8 \right\}$, the middle N tokens are set to $0$, and the last ten tokens are $9$.
The goal of the recurrent model is to output $\left(a_0, \dots, a_9\right)$ in order on the last 10 time steps, whenever the cue token $9$ is presented.
Models are trained using the cross-entropy loss; the random guessing baseline has loss $\log(8) \approx 2.08$.
We use length $L=200$.
The training and testing examples are generated in the same way.

Our motivation of studying the Copying task is that standard models such as the LSTM struggle to solve it.
We note that the Copying task is much harder than other memory benchmarks such as the Adding task~\cite{arjovsky2016unitary}, and we do not consider those.

\begin{figure}[ht]
  \centering
  \includegraphics[width=.7\linewidth]{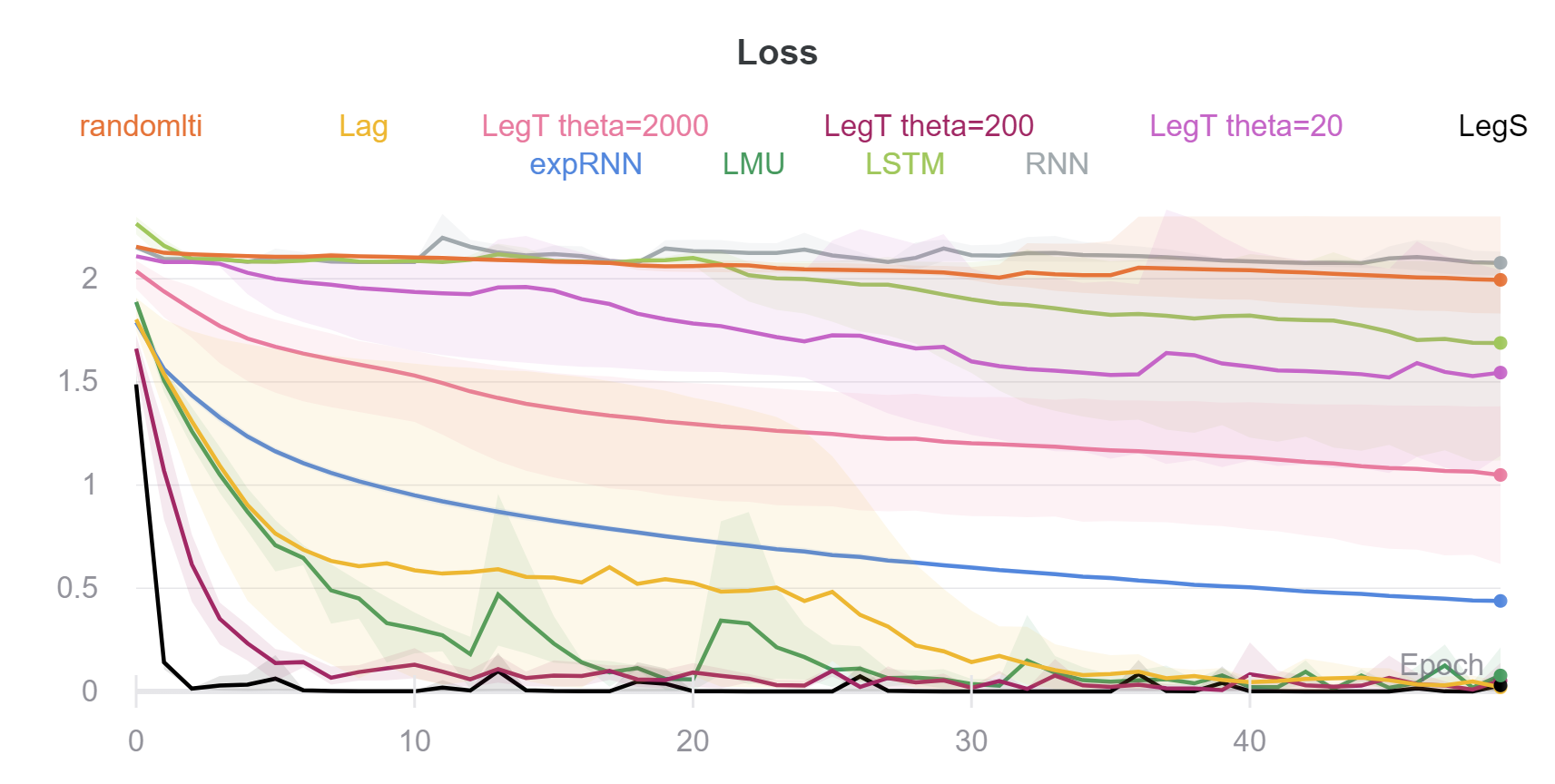}
  \caption{Loss on the Copying task.
    HiPPO methods are the only to fully solve the task.
    The hyperparameter-free LegS update is best, while methods with timescale parameters (e.g.\ LegT) do not solve the task if mis-specified.
  }
  \label{fig:copy200}
\end{figure}

\paragraph{Results}
The HiPPO-LegS method solves this task the fastest.
The LegT method also solves this task quickly, only if the parameter $\theta$ is initialized to the correct value of $200$.
Mis-specifying this timescale hyperparameter to $\theta=20$ or $\theta=2000$ drastically slows down the convergence of HiPPO-LegT.
The LMU (at optimal parameter $\theta=200$) solves this task at comparable speed;
like in \cref{sec:pmnist-details}, differences between HiPPO-LegT $(\theta=200)$ and LMU here arise from the minor architecture difference in \cref{subsec:model_architectures}.

The HiPPO-Rand baseline (denoted ``random LTI'' system here)
does much worse than the updates with the dynamics derived from our framework,
highlighting the importance of the precise dynamics (in contrast to just the architecture).

Standard methods such as the RNN and LSTM are also nearly stuck at baseline.

\subsection{Trajectory Classification}
\label{sec:trajectory-details}

\paragraph{Dataset}
The Character Trajectories dataset~\citep{bagnall2018uea} from the UCI machine learning repository~\citep{Dua:2019} consists of pen tip trajectories recorded from writing individual characters.
The trajectories were captured at 200Hz and data was normalized and smoothed.
Input is 3-dimensional ($x$ and $y$ positions, and pen tip force),
and there are 20 possible outputs (number of classes).
Models are trained using the cross-entropy loss.
The dataset contains 2858 time series.
The length of the sequences is variable, ranging up to $182$.
We use a train-val-test split of 70\%-15\%-15\%.

\paragraph{Methods}

RNN baselines include the LSTM~\citep{lstm}, GRU~\citep{chung2014empirical}, and LMU~\citep{voelker2019legendre}.
Our implementations of these used 256 hidden units each.

The GRU-D~\citep{che2018recurrent} is a method for handling missing values in time series that computes a decay between observations.
The ODE-RNN~\citep{rubanova2019latent} and Neural CDE (NCDE)~\citep{kidger2020neural} baselines are state-of-the-art neural ODE methods, also designed to handle irregularly-sampled time series.
Our GRU-D, ODE-RNN, and Neural CDE baselines used code from~\citet{kidger2020neural}, inheriting the hyperparameters for those methods.

All methods trained for 100 epochs.

\paragraph{Timescale mis-specification}

The goal of this experiment is to investigate the performance of models when the timescale is mis-specified between train and evaluation time, leading to distribution shift.
We considered the following two standard types of time series:
\begin{enumerate}%
    \item Sequences sampled at a fixed rate
    \item Irregularly-sampled time series (i.e., missing values) with timestamps
\end{enumerate}

Timescale shift is emulated in the corresponding ways, which can be interpreted as different sampling rates or trajectory speeds.
\begin{enumerate}%
    \item Either the train or evaluation sequences are downsampled by a factor of 2
    \item The train or evaluation timestamps are halved.\footnote{Instead of the train timestamps being halved, equivalently the evaluation timestamps can be doubled.}
\end{enumerate}
The first scenario in each corresponds to the original sequence being sampled at 100Hz instead of 200Hz; alternatively, it is equivalent to the writer drawing twice as fast.
Thus, these scenarios correspond to a train $\to$ evaluation timescale shift of 100Hz $\to$ 200Hz and 200Hz $\to$ 100Hz respectively.

Note that models are unable to obviously tell that there is timescale shift.
For example, in the first scenario, shorter or longer sequences can be attributed to the variability of sequence lengths in the original dataset.
In the second scenario, the timestamps have different distributions, but this can correspond to different rates of missing data, which the baselines for irregularly-sampled data are able to address.

\subsection{Online Function Approximation and Speed Benchmark}
\label{subsec:function_approx}

\paragraph{Task} The task is to reconstruct an input function (as a discrete
sequence) based on some hidden state produced after the model has traversed the
input function.
This is the same problem setup as in Section~\ref{subsec:hippo-setup};
the online approximation and reconstruction details are in \cref{sec:hippo-framework-details}.
The input function is randomly sampled from a continuous-time band-limited white
noise process, with length $10^6$.
The sampling step size is $\Delta t = 10^{-4}$, and the signal band limit is 1Hz.

\paragraph{Models} We compare HiPPO-LegS, LMU, and LSTM.
The HiPPO-LegS and LMU model only consists of the memory update and not the
additional RNN architecture.
The function is reconstructed from the coefficients using the formula in
\cref{sec:derivations}, so no training is required.
For LSTM, we use a linear decoder to reconstruct the function from the LSTM
hidden states and cell states, trained on a collection of 100 sequences.
All models use $N = 256$ hidden units.
The LSTM uses the $L2$ loss.
The HiPPO methods including LMU follow the fixed dynamics of \cref{thm:legt-lagt} and \cref{thm:legs}.

\paragraph{Speed benchmark}
We measure the inference time of HiPPO-LegS, LMU, and LSTM, in
single-threaded mode on a server Intel Xeon CPU E5-2690 v4 at 2.60GHz.

\subsection{Sentiment Classification on the IMDB Movie Review Dataset}
\label{sec:imdb}

\paragraph{Dataset} The IMDB movie review
dataset~\citep{maas-EtAl:2011:ACL-HLT2011} is a standard binary sentiment
classification task containing 25000 train and test sequences, with sequence
lengths ranging from hundreds to thousands of steps.
The task is to classify the sentiment of each movie review into either positive
or negative.
We use 10\% of the standard training set as validation set.

\paragraph{Methods}
RNN baselines include the LSTM~\citep{lstm}, vanilla RNN,
LMU~\citep{voelker2019legendre}, and expRNN~\citep{lezcano2019cheap}.
Our implementations of these used 256 hidden units each.

\paragraph{Result}
As shown in Table~\ref{tab:imdb_test_acc}, our HiPPO-RNNs have similar and consistent performance,
on par or better than LSTM.
Other long-range memory RNN approaches that constrains the expressivity of the
network (e.g.\ expRNN) performs worse on this more generic task.

\begin{table}
    \centering
    \begin{tabular}{@{}ll@{}}
      \toprule
      Model                  & Test accuracy (\%) \\
      \midrule
      HiPPO-LegS             & 87.8 $\pm$ 0.2 \\
      HiPPO-LagT         & \textbf{88.0} $\pm$ 0.2 \\
      HiPPO-LegT $\theta = 100$   & 87.4 $\pm$ 0.3 \\
      HiPPO-LegT $\theta = 1000$  & 87.7 $\pm$ 0.2 \\
      HiPPO-LegT $\theta = 10000$ & 87.9 $\pm$ 0.3 \\
      HiPPO-Rand             & 82.9 $\pm$ 0.3 \\ \midrule
      LMU $\theta = 1000$         & 87.7 $\pm$ 0.1 \\
      LSTM                   & 87.3 $\pm$ 0.4 \\
      expRNN                 & 84.3 $\pm$ 0.3 \\
      RNN                    & 67.4 $\pm$ 7.7 \\
      \bottomrule
    \end{tabular}
    \caption{IMDB test accuracy, averaged over 3 seeds. Top: Our methods.
      Bottom: Recurrent baselines.}
    \label{tab:imdb_test_acc}
\end{table}

\subsection{Mackey Glass prediction}
\label{sec:mackey}

The Mackey-Glass data~\citep{mackey1977oscillation}
is a time series prediction task for modeling chaotic dynamical systems.
We build on the implementation of~\citet{voelker2019legendre}.
The data is a sequence of one-dimensional observations,
and models are tasked with predicting 15 time steps into the future.
The models are 4-layer stacked recurrent neural networks, trained with the mean squared error (MSE) loss.
\citet{voelker2019legendre} additionally consider a hybrid model with alternating LSTM and LMU layers,
which improved on either by itself.
We did not try this approach with our method HiPPO-LegS such as combining it with the LSTM or other HiPPO methods,
but such ideas could further improve our performance.
As a baseline method, the identity function does not simulate the dynamics,
and simply guesses that the future time step is equal to the current input.

\begin{figure}[!ht]
  \begin{minipage}{.5\linewidth}%
    \centering
    \includegraphics[width=\linewidth]{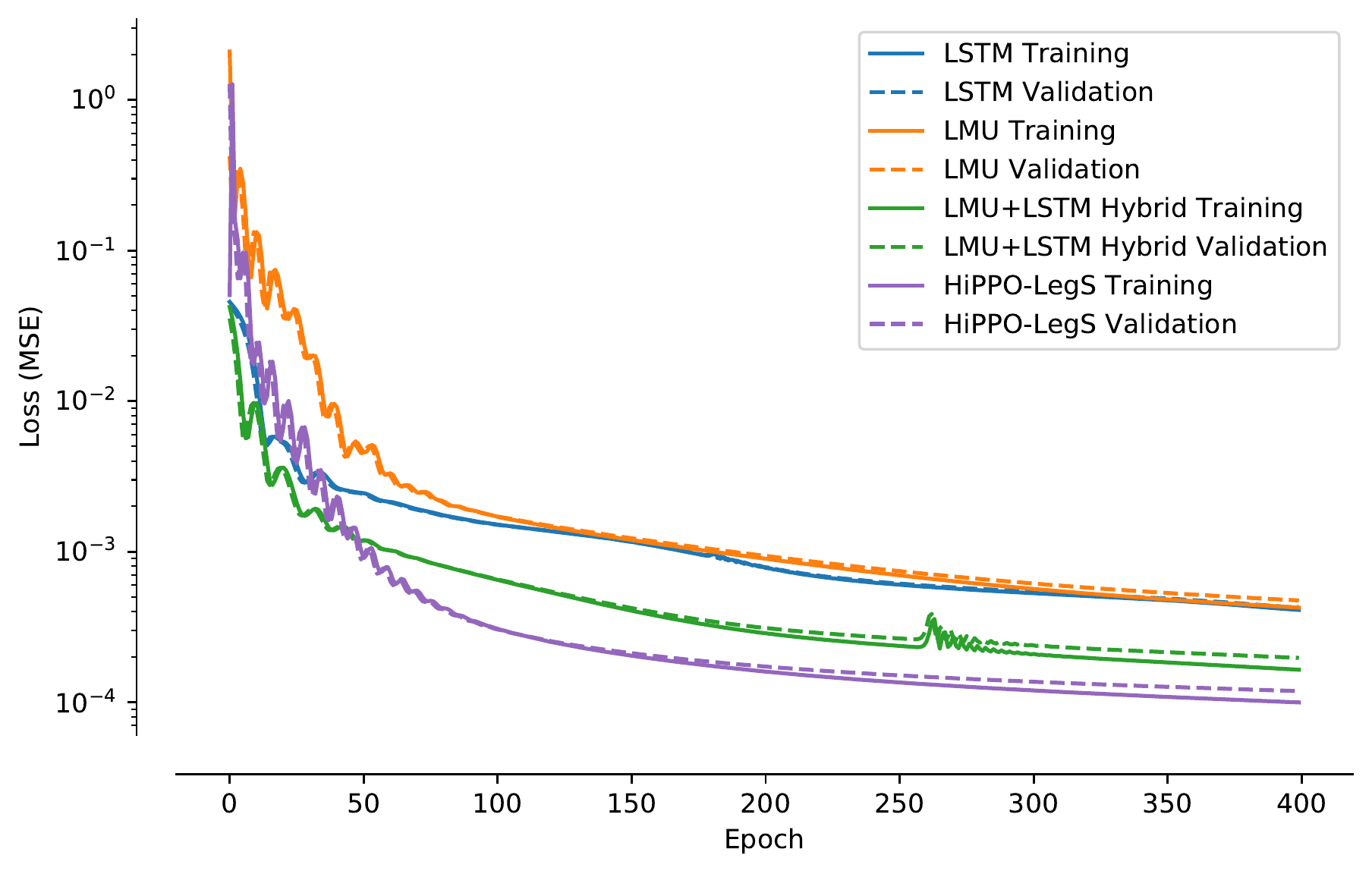}
  \end{minipage}
  \begin{minipage}{.5\linewidth}%
    \begin{tabular}{@{}lll@{}}
      \toprule
      Model    & Test MSE              & Test NRMSE \\
      \midrule
      Baseline & 0.1229                & 1.62274    \\
      LSTM     & 4.784e-4              & 0.10123    \\
      LMU      & 4.414e-4              & 0.09722    \\
      Hybrid LSTM+LMU  & 2.198e-4      & 0.06862    \\
      LegS     & 1.054e-4              & 0.04752    \\
      \bottomrule
    \end{tabular}
  \end{minipage}
  \caption{Mackey-Glass predictions}
  \label{fig:mackey}
\end{figure}

\cref{fig:mackey} plots the training and validation mean squared errors (MSE) of these methods.
The table reports final normalized root mean squared errors (NRMSE) $\sqrt{\frac{\mathbb{E}\left[ (Y-\hat{Y})^2 \right]}{\mathbb{E}[Y^2]}}$ between the targets $Y$ and predictions $\hat{Y}$.
HiPPO-LegS outperforms the LSTM, LMU, and the best hybrid LSTM+LMU
model from [68], reducing normalized MSE by over 30\%.

\subsection{Additional Analysis and Ablations of HiPPO}
\label{sec:exp-hippo-ablations}

\begin{figure}[ht]
  \centering
  \begin{subfigure}[t]{0.33\linewidth}
    \centering
    \includegraphics[width=\textwidth]{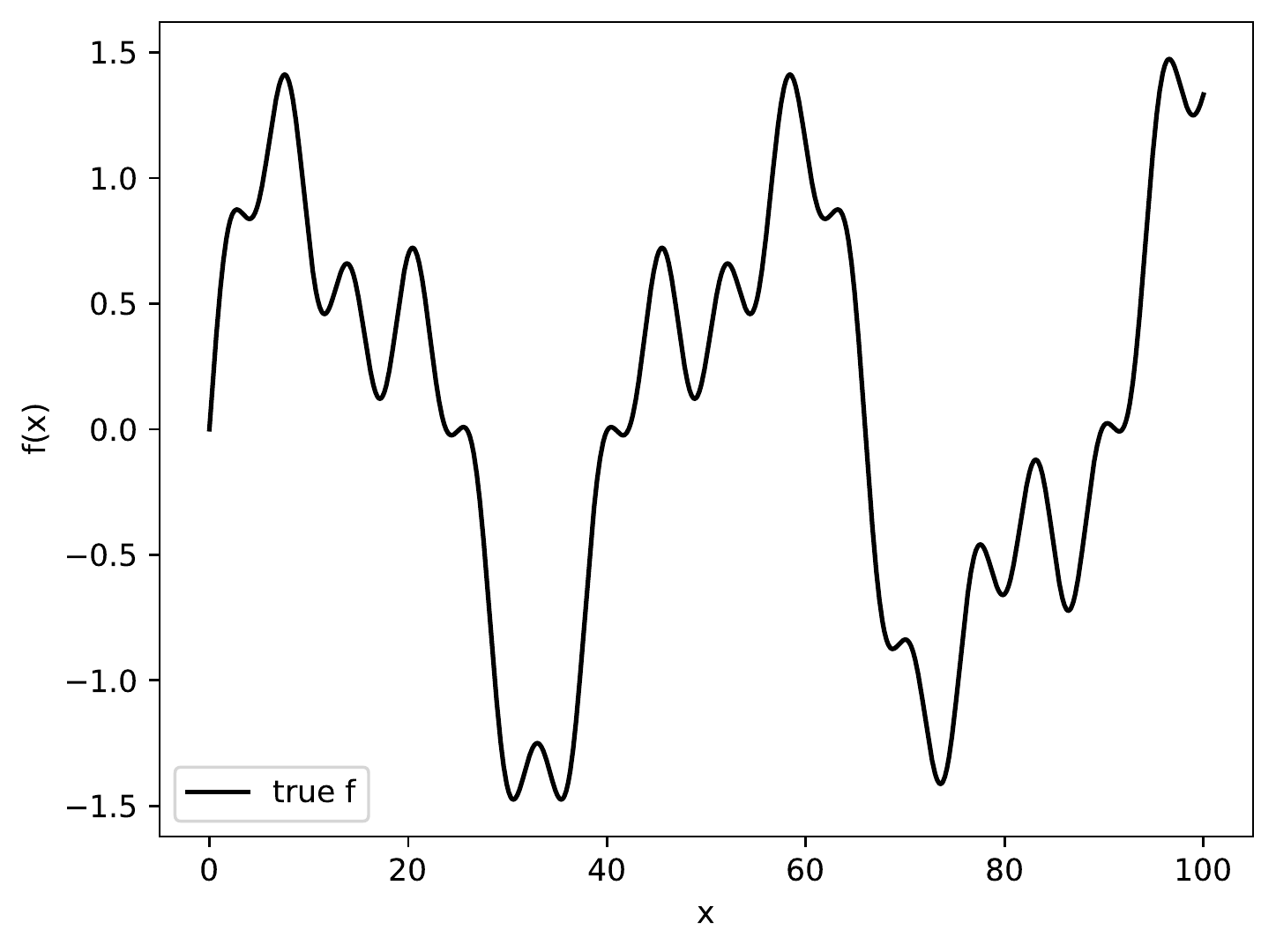}
    \caption{True function $f(x)$}
  \end{subfigure}%
  \begin{subfigure}[t]{0.33\linewidth}
    \centering
    \includegraphics[width=\textwidth]{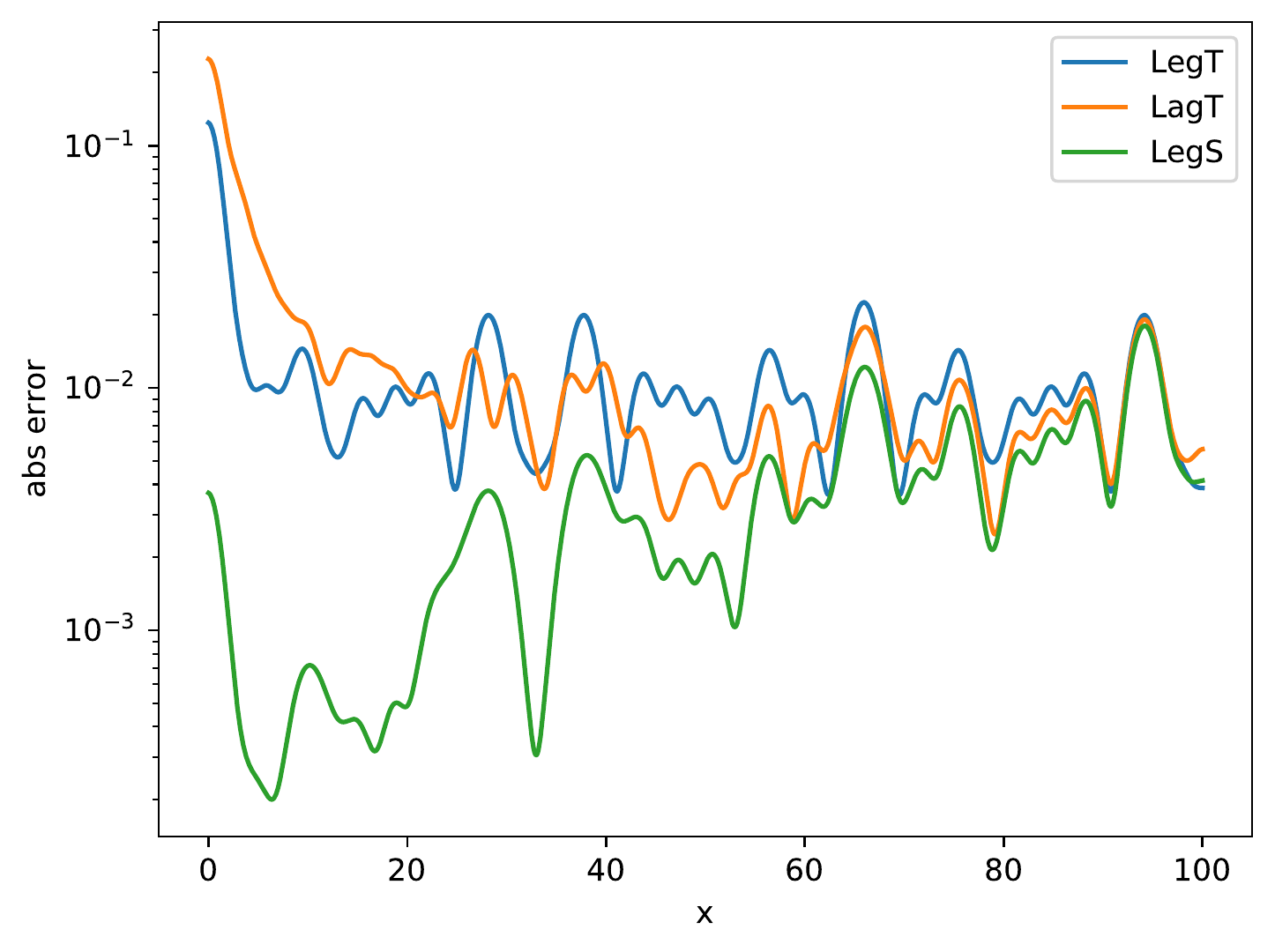}
    \caption{Absolute approx.\ error}
  \end{subfigure}%
  \begin{subfigure}[t]{0.33\linewidth}
    \centering
    \includegraphics[width=\textwidth]{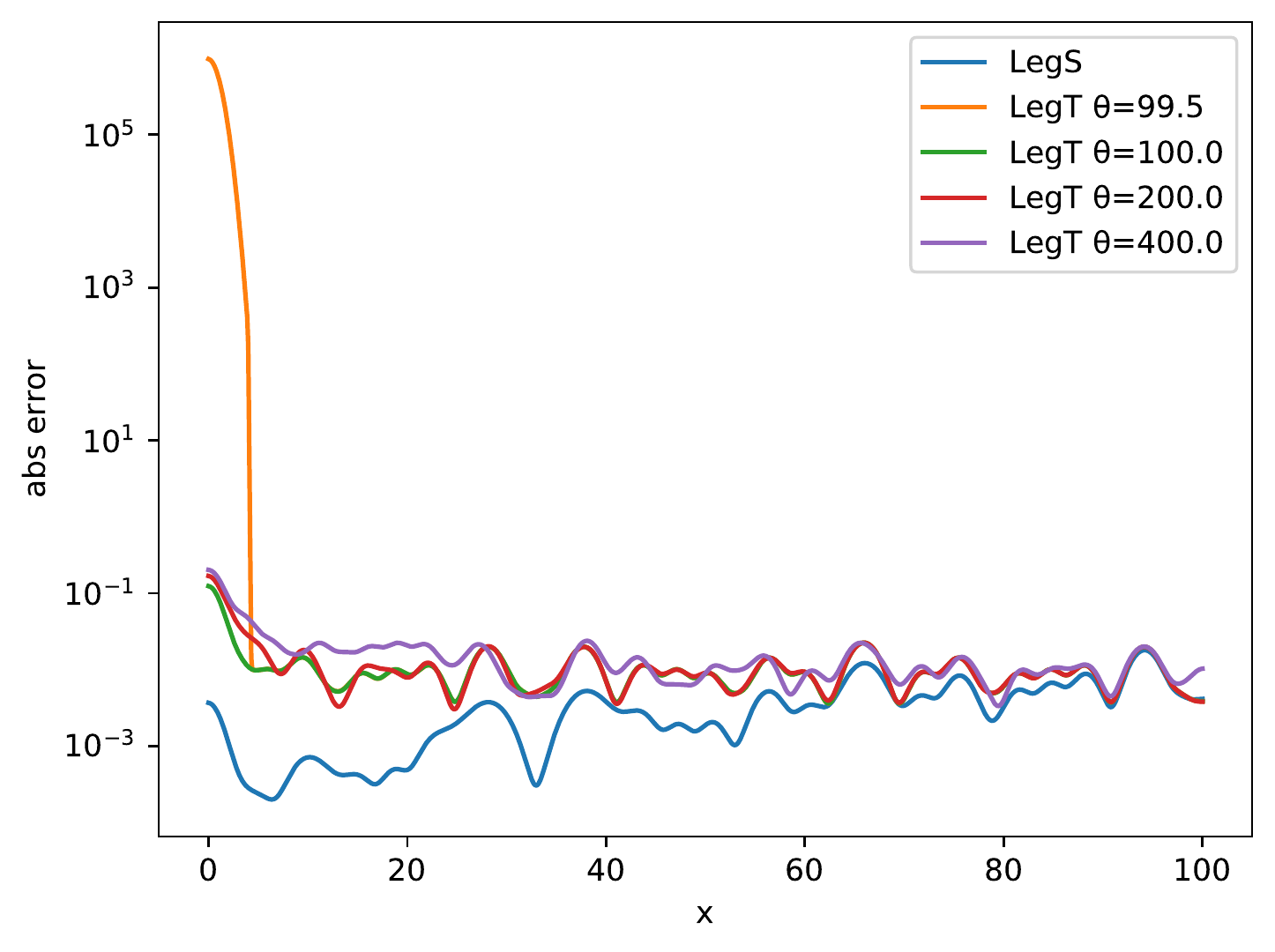}
    \caption{Error for different $\theta$'s in LegT}
  \end{subfigure}%
  \caption{Function approximation comparison between LegT, LagT, and LegS. LegS
  has the lowest approximation error. LegT error is sensitive to the choice of
  window length $\theta$, especially if $\theta$ is smaller than the length of the true function.}
  \label{fig:function_approx_full}
\end{figure}

To further analyze the tradeoffs of the memory updates derived from our framework,
in \cref{fig:function_approx_full} we plot a simple input function
$f(x) = 1/4 \sin x + 1/2 \sin (x/3) + \sin(x/7)$ to be approximated.
The function is subsampled on the range $x \in [0, 100]$, creating a sequence of length $1000$.
This function is simpler than the functions sampled from white noise
signals described in \cref{subsec:function_approx}.
Given this function, we use the same methodology as in \cref{subsec:function_approx} for processing the function online and then reconstructing it at the end.

In Figure~\ref{fig:function_approx_full}(a, b), we plot the true function $f$, and
its absolute approximation error based on LegT, LagT, and LegS.
LegS has the lowest approximation error, while LegT and LagT are similar and slightly worse than LegS.
Next, we analyze some qualitative behaviors.

\paragraph{LegT Window Length}

In Figure~\ref{fig:function_approx_full}(c), shows that the approximation error of
LegT is sensitive to the hyperparameter $\theta$, the length of the window.
Specifying $\theta$ to be even slightly too small (by $0.5\%$ relative to the total sequence length)
causes huge errors in approximation.
This is expected by the HiPPO framework, as the final measure $\mu^{(t)}$ is not supported everywhere, so the projection problem does not care that the reconstructed function is highly inaccurate near $x=0$.

\begin{figure}[ht]
  \centering
  \begin{subfigure}[t]{0.33\linewidth}
    \centering
    \includegraphics[width=\textwidth]{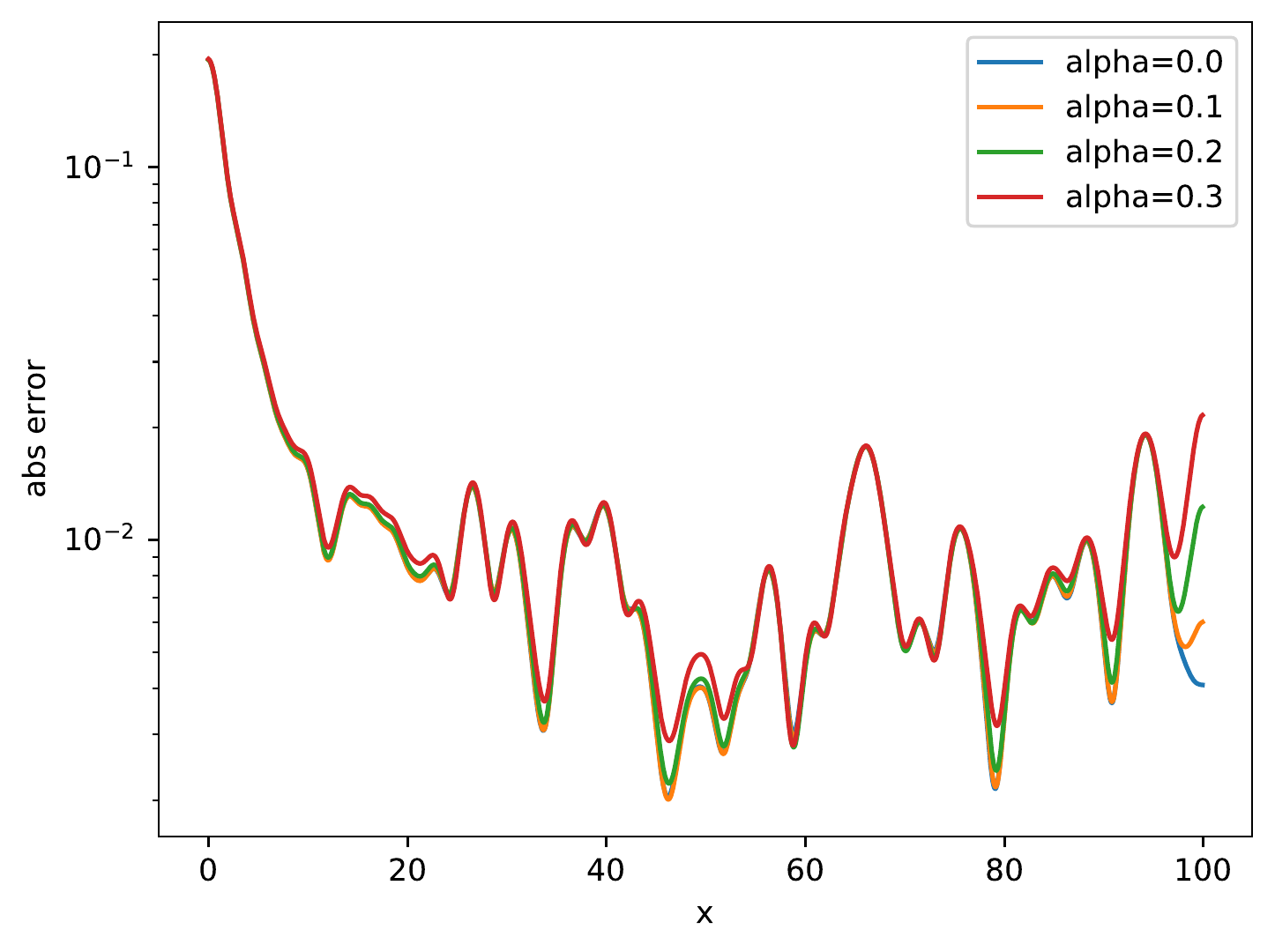}
    \caption{Generalized Laguerre family,\\ fixed $\beta=0.01$ and varying $\alpha$}
  \end{subfigure}
  \begin{subfigure}[t]{0.33\linewidth}
    \centering
    \includegraphics[width=\textwidth]{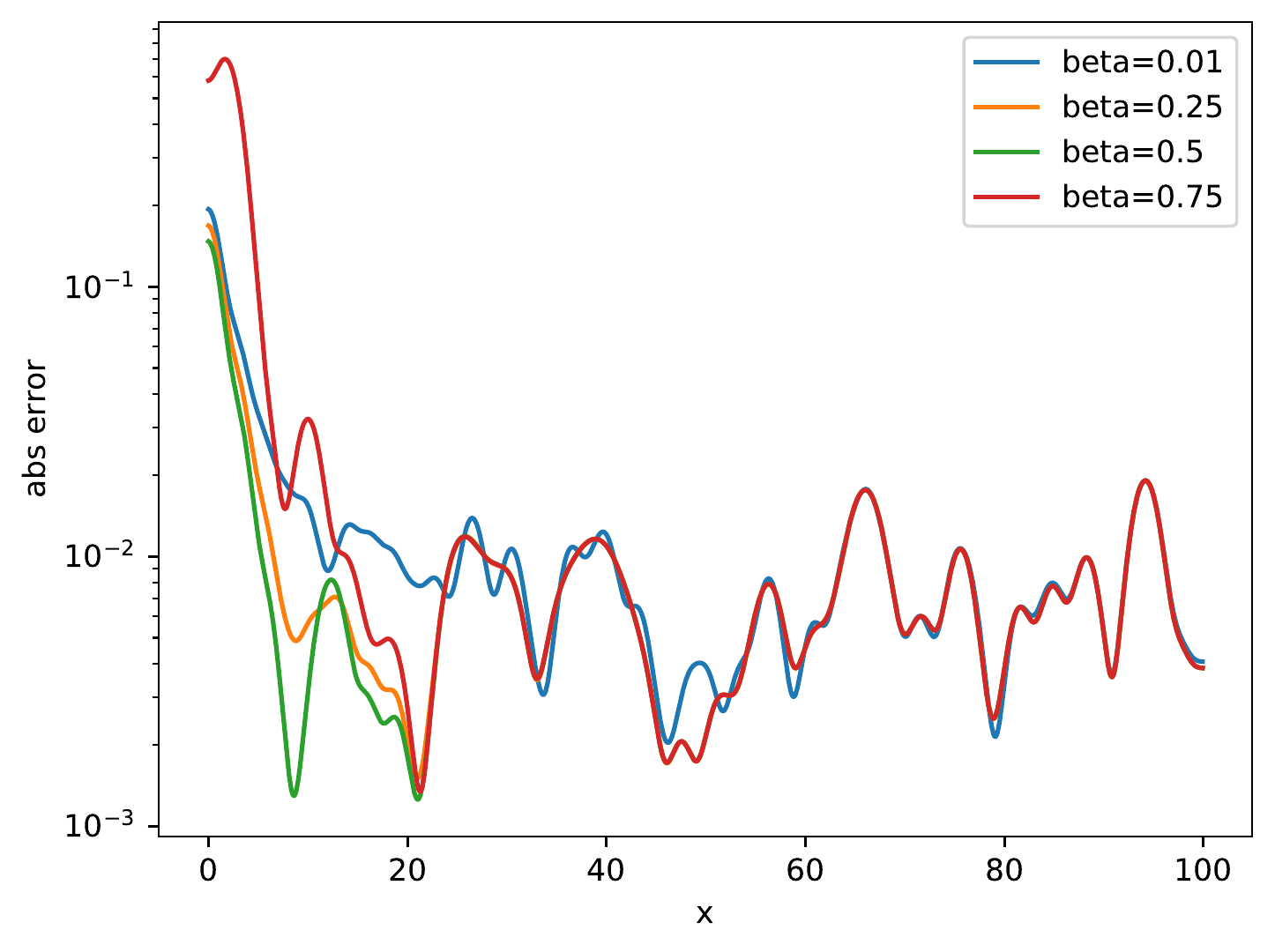}
    \caption{Generalized Laguerre family,\\ fixed $\alpha=0$ and small $\beta$}
  \end{subfigure}%
  \begin{subfigure}[t]{0.33\linewidth}
    \centering
    \includegraphics[width=\textwidth]{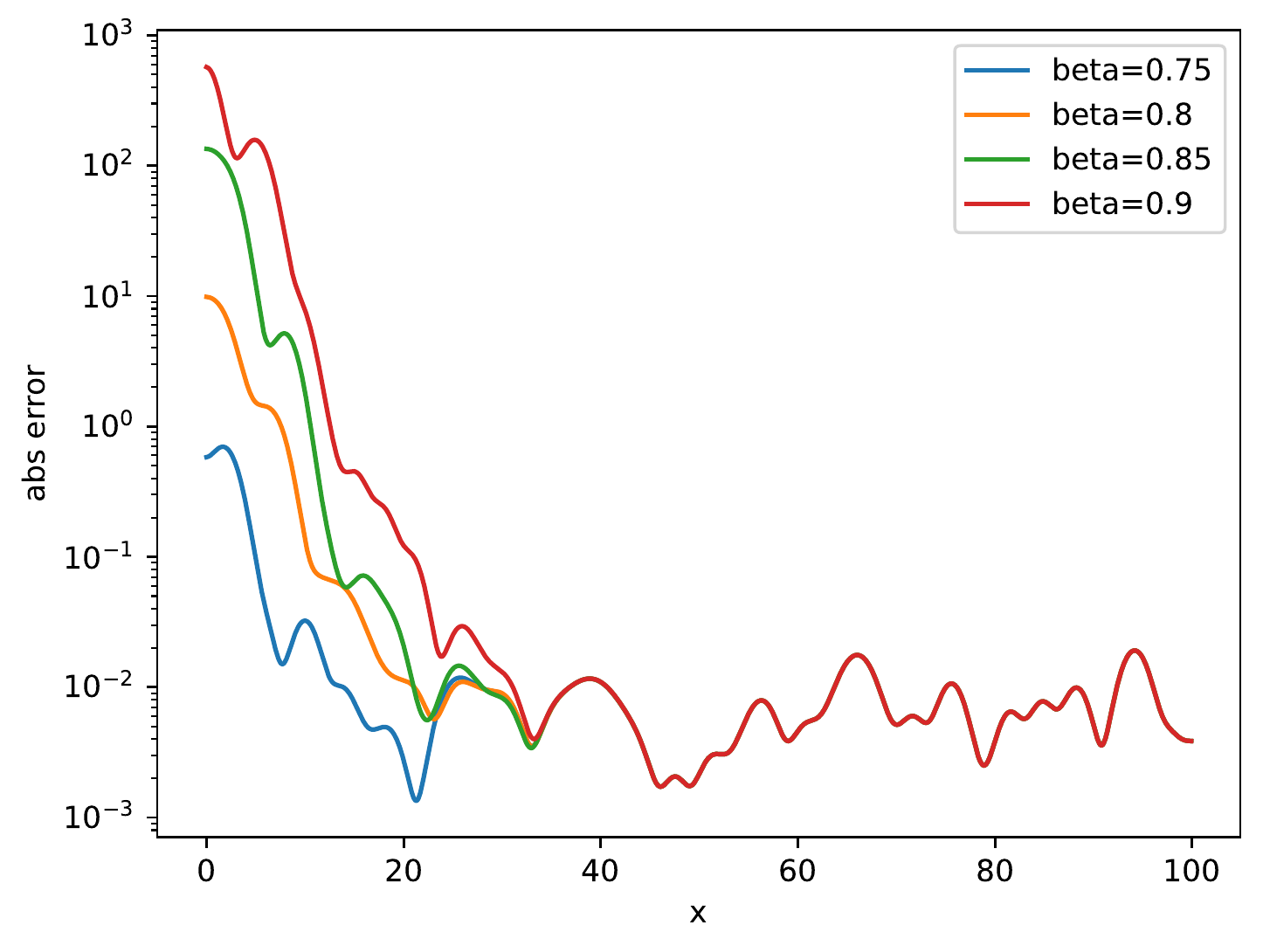}
    \caption{Generalized Laguerre family,\\ fixed $\alpha=0$ and large $\beta$}
  \end{subfigure}%
  \caption{
    Function approximation comparison between different instantiations of the generalized tilted Laguerre family (\cref{sec:derivation-lagt}).
  }
  \label{fig:function-approx-lagt}
\end{figure}

\paragraph{Generalized LagT Family}
Our LagT method actually comprises a family of related transforms,
governed by two parameters $\alpha, \beta$ specifying the original measure and the tilting (\cref{sec:derivation-lagt}).
\cref{fig:function-approx-lagt} shows the error as these parameters change.
\cref{fig:function-approx-lagt}(a) shows that small $\alpha$ generally performs better.
\cref{fig:function-approx-lagt}(b, c) show that the reconstruction is unstable for larger $\beta$, but small values of $\beta$ work well.
More detailed theoretical analysis explaining these tradeoffs would be an interesting question to analyze.

\paragraph{LegS vs. LegT}

In comparison to LegT, LegS does not need any hyperparameters governing the timescale.
However, suppose that the LegT $\theta$ window size was chosen perfectly to match the length of the sequence;
that is, $\theta = T$ where $T$ is the final time range.
Note that at the end of consuming the input function (time $t = T$),
the measures $\mu^{(t)}$ for LegS and LegT are \emph{both} equal to $\frac{1}{T}\mathbb{I}_{[0,T]}$ (\cref{subsec:high_order_projection,sec:theory_legs}).
Therefore, the approximation $\proj_T(f)$ is specifying the same function for both LegS and LegT at time $t=T$.
The sole difference is that LegT has an additional approximation term for $f(t-\theta)$ while calculating the update at every time $t$ (see \cref{sec:derivation-legt}), due to the nature of the sliding rather than scaling window.

\end{document}